\documentclass[man,12pt,floatsintext]{apa7}
\usepackage{amsmath,amssymb,amsfonts}
\usepackage{algorithmic}
\usepackage{graphicx}
\usepackage{textcomp}
\usepackage{tabu}
\usepackage{booktabs}
\usepackage{colortbl}
\usepackage{bm}
\usepackage{url}
\usepackage{xspace}
\usepackage{stfloats}
\usepackage{enumerate}
\usepackage{enumitem}
\usepackage{hyperref}
\usepackage[american]{babel}

\usepackage{csquotes} % One of the things you learn about LaTeX is at some level, it's like magic. The references weren't printing as they should without this line, and the guy who wrote the package included it, so here it is. Because LaTeX reasons.
\usepackage[style=apa,sortcites=true,sorting=nyt,backend=biber]{biblatex}
\DeclareLanguageMapping{american}{american-apa} % Gotta make sure we're patriotic up in here. Seriously, though, there can be local variants to how citations are handled, this sets it to the American idiosyncrasies 
\usepackage[T1]{fontenc} 
\usepackage{mathptmx} % This is the Times New Roman font, which was the norm back in my day. If you'd like to use a different font, the options are laid out here: https://www.overleaf.com/learn/latex/Font_typefaces
\graphicspath{{figures/}}
\newcommand{\argmax}{\mathop{\rm{arg~max}}\limits}
\newcommand{\argmin}{\mathop{\rm{arg~min}}\limits}
\newcommand{\Deg}{$^\circ$\xspace}
\newcommand{\rev}[1]{{\textcolor{black}{#1}}}
\title{Fixation-based Self-calibration for Eye Tracking in VR Headsets} % The big, long version of the title for the title page
\shorttitle{Behavior Research Method} % The short title for the header
\author{
    Ryusei Uramune,
    Sei Ikeda,
    Hiroki Ishizuka,
    Osamu Oshiro
}

\affiliation{{
    Osaka University,
    1-3 Machikaneyama,
    Toyonaka,
    Osaka,
    Japan,
    560-8531
    },
    \textit{Email}: ruramune@acm.org
}

\abstract{
This study proposes a novel self-calibration method for eye tracking in a virtual reality (VR) headset.
The proposed method is based on the assumptions that the user's viewpoint can freely move and that the points of regard (PoRs) from different viewpoints are distributed within a small area on an object surface during visual fixation.
In the method, fixations are first detected from the time-series data of uncalibrated gaze directions 
using an extension of the I-VDT (velocity and dispersion threshold identification) algorithm to a three-dimensional (3D) scene.
Then, the calibration parameters are optimized by minimizing the sum of a dispersion metrics of the PoRs.
The proposed method can potentially identify the optimal calibration parameters representing the user-dependent offset from the optical axis to the visual axis without explicit user calibration, image processing, or marker-substitute objects.
For the gaze data of 18 participants walking in two VR environments with many occlusions, the proposed method achieved an accuracy of 2.1\Deg, which was significantly lower than the average offset.
Our method is the first self-calibration method 
which is applicable to 3D environments without image-processing.
}

 % If you need to have keywords for your paper, delete the % at the start of this line

\begin{document}
\maketitle % This tells LaTeX to make the title page

\section{Introduction}
\label{sec:introduction}
%----------------------------------------------------------------------
% 視線計測とは，視線計測の応用例
%\label{secritical:introduction} %なぜか参照できない
%----------------------------------------------------------------------
% 視線計測は，心理分析\parencite{Smith_IEEE_PAMI2008,Senju_Science2009}などのオフライン用途だけでなく，
% foveated rendering~\parencite{Meng_ACM_TOG2018},
% presise ocular parallax~\parencite{Konrad_ACM_TOG2020,Krajancich_ACM_TOG2020},
% depth of field rendering~\parencite{ORIKASA_IEICE_TransInf2016},
% autofocus displays~\parencite{Padmanaban_SciAdv2019},
% object selection~\parencite{Mardanbegi_CHI2019}など様々なvirtual/augmented reality用途にも有効活用できることが示されている．
% 2022年現在においても，カメラ画像を用いる視線検出器が広く普及しており，
% FOVEに始まり，Vive Pro EyeやMagic Leap, Microsoft HoloLens2などの
% バーチャルリアリティ (VR) 用もしくは複合現実感 (MR) 用ヘッドマウントディスプレイ (HMD) では，視線計測器が標準搭載される傾向にある．
In virtual and augmented reality (VR/AR), eye tracking has been shown to be effective not only for offline applications, such as psychological analysis and usability evaluation,
but also for various online applications, including foveated rendering~\parencite{Meng_ACM_TOG2018}, precise ocular parallax displays~\parencite{Konrad_ACM_TOG2020,Krajancich_ACM_TOG2020}, depth of field rendering~\parencite{ORIKASA_IEICE_TransInf2016}, auto-focus displays~\parencite{Padmanaban_SciAdv2019}, and object selection~\parencite{Mardanbegi_CHI2019}.
As of 2023, camera-based eye-tracking is still widely used as a standard feature in VR/AR head-mounted displays (HMDs), as exemplified by FOVE, HTC Vive Pro Eye, Magic Leap, and Microsoft HoloLens2.
Most devices need to be calibrated before use to obtain highly accurate line of sight.
\rev{In the above applications, the accuracy of gaze measurement can be one of the main factors that determine the performance of the applications. For example, in foveated rendering, low accuracy in gaze measurement increases rendering cost by enlarging the area of detail to be rendered; in object selection, it becomes difficult to select small objects.}

%----------------------------------------------------------------------
% 既存の視線計測手法は明示的な較正動作が必要，自動較正が実現されると嬉しいこと
%----------------------------------------------------------------------
% 視線検出器の校正原理を理解するには、人の視覚システムの知識が必須である。
% 人間は，対象物を視細胞が最も密集する中心窩で鮮明に捉える．
% 中心窩と対象物を結ぶ線を視軸と呼ぶ．
% 一方，眼球の中心と瞳孔中心を通る直線を光軸と呼ぶ．
% 光軸と視軸は水平4から5deg，垂直約1.5deg\parencite{Hansen_IEEE_PAMI2010}のオフセットがあり，オフセットには最大3度の個人差ある\parencite{Carpenter_1988}．
% 眼画像から瞳孔中心や光軸を求めることはできるが，中心窩の位置や視軸を画像から直接求めることはできない．
The principle of eye tracking calibration is strongly based on the human visual system.
The human eyeball obtains a clear image by rotating to capture a target at the fovea,
\rev{a spot on the retina where the distribution density of cones, the photoreceptors mainly related to bright vision,} is the highest.
The line passing through the fovea and the center of the lens is called the visual axis.
In contrast, an eye tracker can directly observe \rev{some feature related to the optical (or geometric) axis}, which is the line passing through the center of the eyeball rotation and the center of the pupil.
The visual axis is typically offset from the optical axis by about 4\Deg to 7\Deg \parencite{Carpenter_1988}.
According to literature~\parencite{Abass_CER2018}, the offset has non-negligible individual differences, with a maximum difference of 19.57\Deg in the horizontal direction and 16.25\Deg in the vertical direction.
% Thus, only information about the optical axis can be directly obtained from eye observations;
% the offset angle is estimated in the calibration of eye trackers.
Therefore, in order to obtain accurate gaze directions by observing the eye balls, a calibration to estimate the offset cannot be avoided.

% そのため，一般に，視線計測器は使用者毎に較正する必要がある．
% 視線計測器の較正は通常，視線計測を始める前，場合によってはその後にも，基準となるマーカを使用者に提示し，使用者にマーカを注視する動作が課される．
% 次章でまとめるように，様々な視線計測手法が存在し，3種類に分類できる．
% それらの共通点は，いずれの較正手法においても，装置の取り付け位置由来と眼の個人差由来の偏差を含む何らかのパラメータを推定することである．
% 注視しているときの視線方向とマーカとの位置関係からそれらのパラメータを決定することで，視線計測器の出力を正しい視線への変換関数が定まる．
% また，較正後も視線計測器の位置が変化すると，視線計測の精度が低下し再較正が必要となる場合もある．
% 自動的に視線計測器を較正することができれば，視線計測後に一括解析する用途においては計測時間の短縮につながる．
% また，視線計測器を実時間利用する用途においては必要に応じて随時較正することが可能となる．
In traditional calibration methods, several reference markers are presented to the user, and the user is asked to gaze at each marker individually, both before and sometimes after gaze measurements.
\rev{Such calibration inhibits VR experiences.}
After the calibration, many state-of-the-art eye trackers maintain accuracy, even when the head is moved in relation to the device.
This ability depends not so much on the calibration method as on the gaze estimation model.
As summarized in the next section, various gaze estimation models exist and can be divided into four categories: two-dimensional (2D) regression models, three-dimensional (3D) eye model-based models, appearance-based models, and head-correlation models~\parencite{Hansen_IEEE_PAMI2010,Kar_IEEE_Access2017}.
The former three models are based on eye observations and require calibration.
The head-correlation models are challenging approach to estimate the visual axis from head motion~\parencite{Li_IEEE_CCV2013,Hu_IEEE_TVCG2019,Hu_IEEE_TVCG2020} without the use of eye trackers. 
Although they do not require the estimation of offsets, they are inaccurate.

%----------------------------------------------------------------------
% 従来の自動較正手法の問題点
%----------------------------------------------------------------------
% これまで，視線計測の自動較正に関する様々な手法が提案されてきた．
% 次章で詳述するとおり，自動較正手法は，learning-based, saliency-based, smooth pursuit-based, scene model-based methodsに分けることができる．
% learning-based methodは，頭部姿勢やシーン画像の入力と視線の出力の関係を大量に学習し
% そのため，視線計測器は必要ないという利点はある．
% しかしながら，平均的な視線しか推定することができず，個人差を補償することが原理的に難しい．
% それに対して，saliency-，smooth pursuit-，scene model-based methodsは，そのユーザの視線方向に基づいた較正が行われるため，その問題がない．
% saliency-based methodsは，多くの人の視線情報から学習されたsaliency map生成手法に頼っていおり，視線推定精度はsaliency mapの精度に依存する．
% 現状のsaliency mapの精度は十分高いとは言えず，学習するシーンに依存するという問題もある．
% Smooth pursuit-methodは，眼球の動きと視界内の同物体の動きの相関を利用する方法で，
% 容易に任意の三次元シーンに拡張することができる\parencite{Murauer_iWOAR2018,Tamura_ETRA2020}．
% しかし，同じ動きをする物体が多い場合や，大きな物体のどの部分を注視しているのかを特定することが難しい．
% scene model-based methodsはシーンの形状が既知であることを前提としている．
% この前提は，VRアプリケーションと親和性が高い．
% しかし，多くの手法は，対象シーンとして平面スクリーンか，微分可能な曲面シーンに特化されていた．
Self-calibration (also called automatic or implicit calibration) can be defined as a technology that estimates the individual offset angles from other information, such as scene images or time-series data of the optical axis.
Self-calibration can reduce \rev{total experimental time} in offline applications.
In online applications, the eye tracker can be calibrated at any time as needed.
Self-calibration methods can be divided into saliency-based~\parencite{Sugano_IEEE_PAMI2013,Chen_IEEE_TIP2015,Wang_ETRA2016,Shi_Elsevier_CAG2020,Liu_IEEE_Access2020}, smooth pursuit-based~\parencite{Murauer_iWOAR2018,Tamura_ETRA2020}, and scene model-based methods~\parencite{Nagamatsu_ETRA2010b,Model_ETRA2010,Maio_FG2011,Miki_ETRA2016,Wang_Elsevier_PR2018,Nagamatsu_PETMEI2013,Uramune_SIGMR202101},
as described in detail in the next section.
% In the learning-based method, input-output relations are learned from a large amount of data, using head posture and scene images as input and gaze as output.
% This method has the advantage of not requiring an eye tracker.
% However, only the average gaze can be estimated, and it is difficult in principle to compensate for individual differences.
% On the other hand, saliency-, smooth pursuit- and scene model-based methods do not have this disadvantage because they are estimate calibration parameters directly based on the user's gaze direction.
% Saliency-based methods rely on a saliency map generation~\parencite{Sugano_IEEE_PAMI2013,Chen_IEEE_TIP2015,Wang_ETRA2016,Shi_Elsevier_CAG2020,Liu_IEEE_Access2020} method learned from many people's gaze information, and the accuracy of gaze estimation depends on the accuracy of the saliency map.
% The accuracy of the current saliency map is not high enough, and it depends on the scene to be trained.
% The smooth pursuit-method uses the correlation between eye movement and movements of objects in the field of view, and can be easily extended to arbitrary 3D scenes~\parencite{Murauer_iWOAR2018,Tamura_ETRA2020}.
% However, it is difficult to identify when there are many objects with the same motion or which part of a large object is being gazed at.
% The scene model-based methods assume that the shape of the scene is known.
These approaches vary in terms of finding the visual axis cues.
% If visual axis cues are available for some frames or multiple frames, the offset can be computed.
% Once the offset is obtained, the optical axis can then be converted to the visual axis from the eye image alone.
Most methods compute the offset angles from the cues obtained by analyzing scene images or scene models over a set amount of time.
Once the offsets are obtained, the optical axis can be directly converted to the visual axis.
% Specifically, the saliency-based methods treat a saliency map of the scene image as a probability distribution of the visual axis and accumulate the maps of multiple frames to increase the probability of identifying the direction of the visual axis.
% The pursuit-based methods identify the direction of the visual axis by assuming that the visual axis has a high probability of being around the object that has similar motion in scene images to that of the optical axis.
% Although these methods have the advantage of easily extending to 3D scenes as well as flat monitors, 
% they are inaccurate because it is difficult to identify a single direction of the visual axis from the scene images.
In particular, scene model-based methods have been reported to achieve higher accuracy than the other methods. 
However, no cases of successful calibration using this type of method in 3D environments in which the user can freely move have been reported to date.
Moreover, the conventional scene model-based methods are limited to flat monitors~\parencite{Nagamatsu_ETRA2010b,Model_ETRA2010,Maio_FG2011,Miki_ETRA2016,Wang_Elsevier_PR2018}, objects at infinity~\parencite{Nagamatsu_PETMEI2013}, or differentiable curved surfaces~\parencite{Uramune_SIGMR202101}.

%----------------------------------------------------------------------
% 本論文の目的と手法
%----------------------------------------------------------------------
% 本論文では，次の特長を併せ持つ，視線計測器の自動較正手法を提案する．
% ・マーカを注視する能動的な較正動作は必要ない．
%   シーンを自由に見渡すだけの受動的な動作だけで良い．
% ・マーカやマーカの代わりとなる特定の物体が必要ない．
% ・シーンにはオクルージョンや様々な形の物体が含まれていても良い．
% ・計算量の高い顕著性マップ生成やオプティカルフロー推定，物体検出などの画像処理が必要ない．
% ・潜在的に眼球の個人差に由来する偏差を補正することができる．
% In this study, we propose a self-calibration method for an eye-tracking VR headset.
% The method has the following advantages:
% \begin{itemize}
% 	\setlength{\itemindent}{0pt}   %5. 最初のインデント
% 	\setlength{\labelsep}{5pt}     %6. item と文字の間
%     \setlength{\itemsep}{0mm} % 項目の隙間
%     \setlength{\parskip}{0mm} % 段落の隙間	
%     \item \rev{No active gazing at markers or fingers is required; instead, it is only necessary for a certain amount of passive movement such as freely looking or walking in a VR environment.}
%     % \item Only passive motion, such as freely looking or walking in a VR environment, rather than actively gazing at markers or fingers, is required.
%     \item No markers or specific objects to replace markers are required.
%     \item The scenes can contain multiple objects with various textures, shapes, and occlusions.
%     \item The method does not rely on computationally expensive image processing, such as saliency map generation, optical flow estimation, or object detection.
% \end{itemize}

In this study, we propose a self-calibration method for eye tracking in VR headsets.
The purpose of this paper is to show the following assertions:
\begin{enumerate}[label=\Roman*.]
	\setlength{\itemindent}{0pt}   %5. 最初のインデント
	\setlength{\labelsep}{5pt}     %6. item と文字の間
    \setlength{\itemsep}{0mm} % 項目の隙間
    \setlength{\parskip}{0mm} % 段落の隙間	
    \item The proposed method is the only self-calibration method applicable in 3D environments where scene images cannot be acquired (see~\hyperref[sec:relatedwork]{\it Related Work}).
    \item The proposed method takes advantage of the fact that many lines of sight during fixation intersect near a single point on the object surface even when the user's head is moving.
    The proposed method does not require scene images or specific objects such as markers (see \hyperref[sec:method]{\it Method}).
    \item The proposed method can calibrate the eye tracker with an accuracy of about 2 degrees in a situation where the user is walking and moving in a VR environment with many objects in view, including various textures and occlusions (see~\hyperref[sec:experiments]{\it Experiments}).
\end{enumerate}
% これらの特長の代わりに，提案法では既知情報として頭部位置とシーンモデルが仮定されている．
% 具体的には，視線計測器により各時刻に対応する使用者の視線と頭の位置姿勢を取得し，3次元シーン中で物体表面の形状が与えられるとき，使用者が自由にシーンを見渡している間に較正パラメータを推定する．
% このような前提において，物体の表面に分布する固視中の注視点群の3次元位置を指標として較正パラメータを推定する．
% 本論文では，先行研究と同様に，単純にオブジェクト座標で注視点が一定時間密集する状態を固視と定義する．
% 物体や頭部が動いている場合でも，追跡眼球運動によって，注視点がオブジェクト座標上では，密集する．
% 物体や頭部が動いている場合でも，眼球は追跡眼球運動によって，1点を追跡し続けるので，その意味で原理的には，smooth pursuit-based mehodsにも近い．
% 固視中の注視点は物体表面上に密集して分布するが，視線計測誤差が大きいときには，固視中の注視点群の分散が大きくなることを利用して較正関数のパラメータを最適化する．
The proposed method assumes the head position and scene model as known information and estimates the calibration parameters using the points of regard (PoRs) during fixational eye movements.
Therefore, this method can be classified as a scene model-based method.
The definition of fixation varies among researchers and communities~\parencite{Hessels_RSOS2018}.
In this study, as in previous studies~\parencite{Masse_IEEE_PAMI2018,Steil_ETRA2018}, we define fixation as a state in which the PoRs are densely packed in the object coordinates for a certain period of time.
According to this definition, even when the object being gazed at or the head is in motion, the eyes track a single point on the object by smooth pursuit eye movements during a fixation.
Thus, in principle, our method is also similar to smooth pursuit-based methods except that the object being tracked must be identified.

\begin{figure}[t]
\centering
\includegraphics[width=0.5\columnwidth]{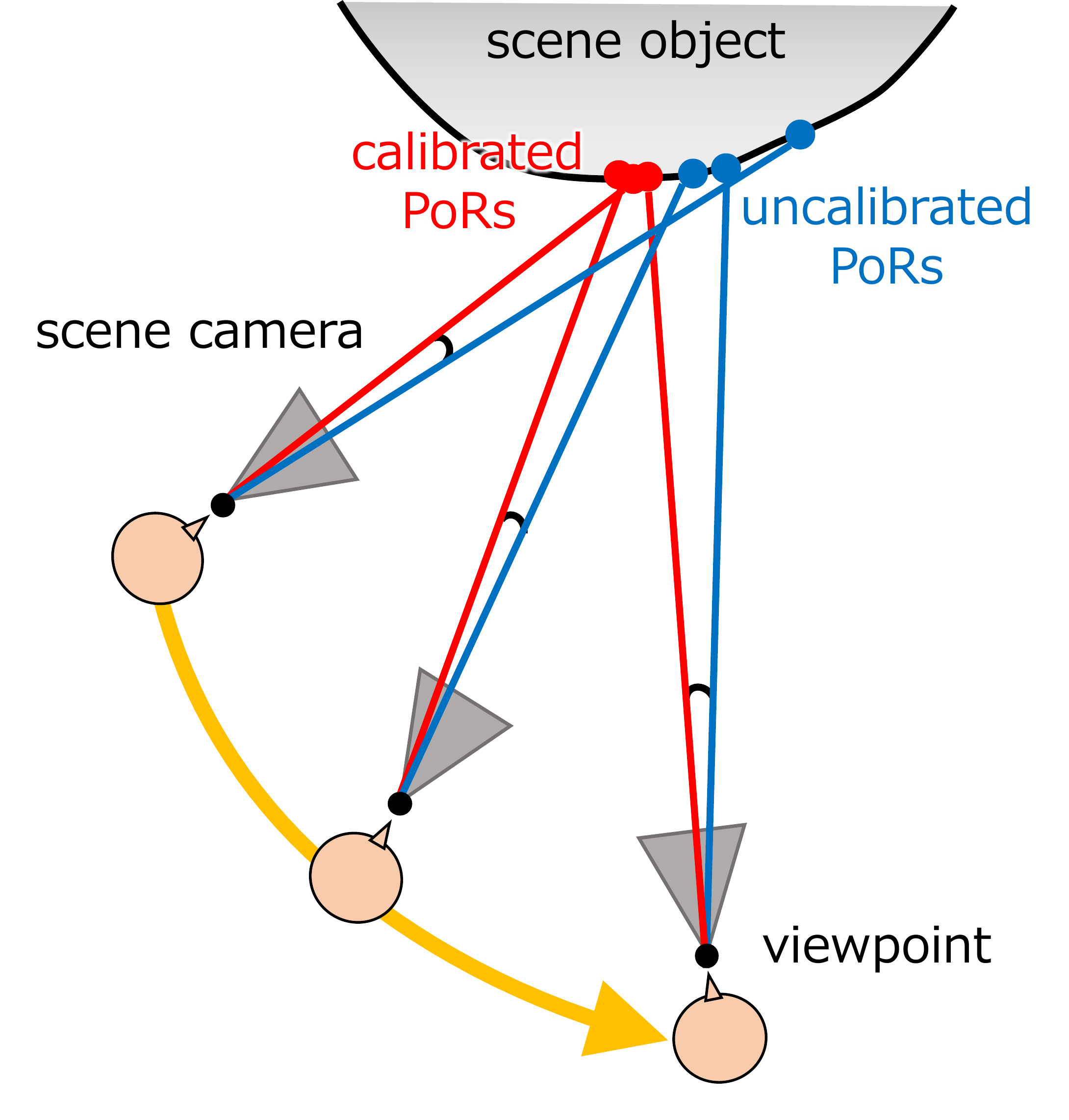}
\vspace{-3mm}
\caption{Principle of the proposed fixation-based calibration method.
% Fixation中の注視点は頭部が移動しても一箇所に集中するように眼球が運動する．
% 提案手法ではこの性質と多視点カメラ幾何学を用いて視線検出器を自動較正する．
During a fixation, human eyeballs move so that the PoRs are concentrated around a single point, even if the head moves.
We applied this characteristic and the corresponding multi-view geometry to the self-calibration of an eye tracker.}
\label{fig:principle}
\end{figure}

As shown in Fig.~\ref{fig:principle}, our method is based on the assumptions that the PoRs in each fixation are densely concentrated on an object surface if the eye tracker is accurately calibrated and that the dispersion of the PoRs in each fixation increases when the gaze measurement contains large errors.
\rev{
Evaluating the dispersion of PoRs is similar to evaluating whether the thinnest part of the line-of-sight bundle is near the object surface. 
The proposed method is inspired by multi-view geometry techniques such as multi-view stereo and camera self-calibration, where the user's view is represented by a pinhole camera model.
In such multi-view techniques, rays from multiple cameras to the same point intersect on the object surface as a clue to estimate some parameter.
It is well known that the rotation of the camera is generally irrelevant, and the baseline distance between different cameras affects the accuracy. The same is true for our method, where simply rotating the user's head at the same point is not sufficient for calibration; the head must be translating during the short period of time that fixation occurs.}

% 提案手法は，二つのステップから構成される．
% 最初に，時系列データから固視中の注視点群を検出する．
% 固視検出アルゴリズムには，既存のアルゴリズムと，3次元での固視検出に適するように拡張したアルゴリズムを使用した．
% 第二に，今回我々が新たに提案したコスト関数を最小化することにより，較正関数のパラメータを最適化する．
% コスト関数は，ユーザの視界をピンホールカメラとして扱い，再投影誤差をもとにデザインされている．
% このシーンカメラはユーザの視野を表すピンホールカメラであり，カメラ姿勢は頭部姿勢を表す．
The proposed method has two steps.
First, it detects fixations from the time-series data of uncalibrated gaze directions.
We extend an existing I-VDT (velocity and dispersion threshold identification)~\parencite{Komogortsev_BRM2013} algorithm suitable for fixation detection in 3D scenes.
Second, the method estimates calibration parameters by minimizing a novel cost function.
The cost function is designed based on the reprojection errors of the PoRs.
% The reprojection errors are defined on the image plane of a scene camera, which is a pinhole camera model representing the user's field of view (See Fig.~\ref{fig:principle}).
The reprojection errors are defined on the image plane of the pinhole camera representing the user's field of view, which is named a scene camera in this paper (see Fig.~\ref{fig:principle}).

%----------------------------------------------------------------------
% 本論文のコントリビューション
%----------------------------------------------------------------------
% 本論文のコントリビューションは次の通りまとめることができる．
% ・固視中の注視点の再投影誤差を用いた自動較正法を提案した．
%   頭部位置が自由に動き，オクルージョンを含む非平面のシーンに対応した同様の指標は初めてである．
% ・2つのリアリスティックなシーンに対して提案手法により自動較正が可能なことを示した．
% ・提案手法が固視検出アルゴリズムと検出時のパラメータに依存することを示した．
% 提案手法は，光軸で検出した場合2.1\Deg，視軸で検出した場合1.2\Degの精度を達成した．
% The contributions of this paper can be summarized as follows:
% \begin{itemize}
% 	\setlength{\itemindent}{0pt}   %5. 最初のインデント
% 	\setlength{\labelsep}{5pt}     %6. item と文字の間
%     \setlength{\itemsep}{0mm} % 項目の隙間
%     \setlength{\parskip}{0mm} % 段落の隙間
%     \item We proposed a novel self-calibration method which applicable to arbitrary VR scenes with occlusions. 
%     This is the first method that have all of the above-mentioned advantages.
%     \item We have shown that our self-calibration method can be applied to realistic VR scenes that contain multiple objects with various textures.
%     \item We showed that the proposed method depends on the fixation detection algorithm and the parameters at detection.
%     The proposed method achieved an accuracy of 2.1\Deg when detected in the optical axis and 1.2\Deg when detected in the visual axis.
% \end{itemize}

\section{Related Work}
\label{sec:relatedwork}
% 本章では，まず\ref{\label{sec:Calibration Models}}節で，
% 画像から視線を推定するための様々なキャリブレーションモデルが存在することを示し，それぞれのモデルで必要となる較正パラメータの種類について説明する．
% 次に\ref{sec:auto-calib}節で，そのようなパラメータを自動推定する手法を紹介し，提案手法との違いを明確化する．
% 最後に，\ref{sec:fixation detection algorithms}節で，既存の固視検出アルゴリズムとその問題点について述べる．
In \hyperref[sec:Calibration Models]{\it Gaze Estimation Models}, we first briefly summarize the various models for gaze estimation from sensing data and show the types of calibration parameters required for each model. 
Then, in \hyperref[sec:auto-calib]{\it Self-calibration Methods}, we explain the existing methods for automatically estimating such parameters and clarify the differences between these and the proposed method.
Finally, in \hyperref[sec:fixation detection algorithms]{\it Fixation Detection Algorithms for 3D Scenes}, we describe the existing fixation detection algorithms and their problems,
as the results of fixation detection significantly affect calibration accuracy (see \hyperref[sec:Dependence on Fixation Detection]{\it Dependence on Initial Parameters}).
% 固視検出アルゴリズムによる検出結果は，使用するアルゴリズムと閾値等のパラメータに強く依存する~\parencite{Shic_ETRA2008, Blignaut_Springer_APAP2009}．
% 先行研究~\parencite{Manor_NM2003,Poole_EHCI2006}は，アルゴリズムや閾値が変わると，固視の数，平均固視時間などが大きく変わることを報告した．
% 後段の分析における結論に影響を与えた例もある\parencite{Shic_ETRA2008}．
% 提案手法のような最適化問題では，固視検出結果の影響を強く受ける．
% そのため，\ref{sec:fixation detection algorithms}節で，3次元シーンでの検出に適したアルゴリズムの必要性を説明する．
% Fixations are strongly dependent on the algorithm used for detection and parameters such as threshold values~\parencite{Shic_ETRA2008, Blignaut_Springer_APAP2009}.
% Previous studies~\parencite{Manor_NM2003,Poole_EHCI2006} reported that the number of fixations and the average fixation time changed significantly when the algorithm and threshold changed.
% In some cases, this has affected the conclusions in later analyses\parencite{Shic_ETRA2008}.
% An optimization problem such as the proposed method is strongly affected by the result of fixation detection.
% Therefore, in the Section~\ref{sec:fixation detection algorithms}, we explain the need for algorithms suitable for detection in 3D scenes.

\subsection{Gaze Estimation Models}
\label{sec:Calibration Models}
% 既存の視線推定モデルは2次元回帰モデル，3次元眼球モデル，外観モデルの3つのタイプに分けられる．
% いずれのモデルにおいても様々なアプローチが提案されているが，各利用者に対して必ず一度個人差に由来する情報を取得する必要があることを示す．
As mentioned above, existing gaze estimation models can be divided into four types: 
2D regression models, 3D eye models, appearance models~\parencite{Hansen_IEEE_PAMI2010}, and head-correlation models. 
This section discusses the conventional gaze estimation models and how they are not capable of determining the visual axis with sufficient accuracy without the provision of individual offset angles.

\subsubsection{2D Regression Models}
%----------------------------------------------------------------------
% 2D回帰モデル
%----------------------------------------------------------------------
%多項式モデルの例．
% - \parencite{Stampe_BRM1993}~biquadratic polynomial function 
% - Morimoto et al.\parencite{Morimoto_Elseiver_IMAVIS2000}- one second order polynomial < そんな事書いてない
% - White et al.[156] \parencite{White_IEEE_SMC1993}- scaling and translational transformation
% - Merchant et al. [99] - * polynomial %<- 入手できず
% - Zhu [171] \parencite{Zhu_ICPR2006} - Support Vector Machines
% - Witzner et al. [45,46] Gaussian process interpolation - Hansen
% - Neural Network Zhu_Springer_MVA2004
%----------------------------------------------------------------------
% 2D回帰モデルは，眼画像における瞳孔中心やグリントなどの二次元座標からから2自由度の視線方向へ変換する写像として表される．
% 座標変換としては，任意の変換関数のテイラー近似として，多項式が使用されることが多い．
% 多項式の例として，
% scaling and translational transformation~\parencite{White_IEEE_SMC1993}, 
% biquadratic polynomial function~\parencite{Stampe_BRM1993}などが用いられており，
% 多項式の他にもSupport Vector Machines~\parencite{Zhu_ICPR2006}, 
% Gaussian process interpolation~\parencite{Hansen_WACV2002}を用いたモデルも存在する．
2D regression models map 2D coordinates, including the pupil center and glint positions (reflections of light sources), in an eye image to a two-degree-of-freedom (DoF) gaze direction.
As mapping functions, polynomials~\parencite{White_IEEE_SMC1993,Wang_ETRA2016,Stampe_BRM1993} are often used as Taylor approximations of arbitrary transformation functions~\parencite{Blignaut_2013}.
% Examples of polynomials include scaling and translational transformation~\parencite{White_IEEE_SMC1993}, a general linear transformation~\parencite{Wang_ETRA2016} and biquadratic polynomial function~\parencite{Stampe_BRM1993}.
In addition to polynomials, machine learning models such as Gaussian process interpolation~\parencite{Hansen_WACV2002}, neural networks~\parencite{Zhu_Springer_MVA2004}, and support vector machines~\parencite{Zhu_ICPR2006} are used.

%----------------------------------------------------------------------
% 2D回帰モデルの較正
%----------------------------------------------------------------------
% 2D回帰モデルの較正では，視野全体を偏りなくカバーするように配置されたマーカを注視する能動的な較正動作が必要となる．
% 較正における回帰係数は，多項式モデルの場合は多項式の係数であり，その他の学習器では各々のモデルパラメータである．
% 一旦較正すれば，後述する3次元モデルベース手法では別途必要となるeyeカメラの内部・外部パラメータを別途推定する必要がない．
% 逆に，2D regression modelsは，カメラに起因する射影歪みやレンズ歪を含むため，カメラ一台では，視線計測器と眼の位置関係の変化を扱うことができない．
% したがって，使用者が同一であっても視線検出器に位置ずれが生じた場合には，再較正が必要となる．
% また一般に，較正には視野角を覆うような十分な数のマーカが必要となる．
Calibration of 2D regression models requires numerous pairs of pupil center, approximately corresponding to the optical axis, and marker position, corresponding to the visual axis.
The regression coefficients in the calibration are the coefficients of the mapping function.
These coefficients contain information on the pose of the eyeball relative to the camera and the offset angles, which cannot be separated.
% Once calibrated, the intrinsic and extrinsic parameters of the eye camera do not need to be estimated separately from the calibration parameters, but are required in the 3D model-based method described below.
Therefore, in 2D regression models, recalibration is necessary when the user's head moves relative to the eye tracker, even if the same user is continuously using the same tracker.
In addition, calibration generally requires a sufficient number of markers to cover the entire field of view (FOV).
In our experiments, we used the calibration results from a polynomial-based 2D regression as a control condition.

\subsubsection{3D Eye Models}
%----------------------------------------------------------------------
% 3D眼球モデル
%----------------------------------------------------------------------
% 3次元眼球モデルベース手法は，画像特徴と光軸を推定するために利用される．
% それらの大半は，2つの球体の組み合わせを，そのうち少数は楕円体\parencite{Beymer_CVPR2003}や回転体\parencite{Nagamatsu_ETRA2010a}などのシンプルな図形で眼球の表面形状近似する．
% 3次元眼球モデルを用いた視線推定法では，理論的には，カメラ1台と2つの点光源があれば，頭部の動きを許容して光軸が推定できる~\parencite{Guestrin_IEEE_TBME2006}．
% しかし，視線方向に直接対応するのは光軸ではなく視軸であり，それは中心窩を通り，角膜曲率中心で光軸が交差する．
Calibration of 3D eye models determines the optical axis through feature detection on eye images.
Most of the eye model-based approaches approximate the surface shape of the eyeball using a combination of two spheres~\parencite{Wang_Elsevier_CVIU2005,Villanueva_WS_IJPRAI2007}; a few of them use an ellipsoid~\parencite{Beymer_CVPR2003} or a solid of revolution around the optical axis~\parencite{Nagamatsu_ETRA2010a}.
The optical axis can be determined using one camera and a few point light sources, allowing for head movement~\parencite{Guestrin_IEEE_TBME2006}.
% The direct correspondence to the direction of gaze is not the optical axis, but the visual axis, which passes through the fovea and intersects the optical axis at the center of corneal curvature.
Our method adopts two DoF offset angles as the calibration parameters of the sphere model.

%----------------------------------------------------------------------
% 3D眼球モデルの較正
%----------------------------------------------------------------------
% 3Dモデルのパラメータ\parencite{Hansen_IEEE_PAMI2010}[8] (これは記述してない)
% ・角膜曲率半径（cornea radii）
% ・視軸と光軸のオフセット
% ・房水の屈折率（refraction paraemters of aqueous humor）
% ・虹彩半径（iris radius）
%--------------------------------------------------------------
% モデルベース手法により光軸方向が得られたとしても，foveaは画像から直接観測不能であるため，少なくとも較正による光軸から視軸への回転変換が必要となる．
% 多くの3Dモデルでは，光軸と視軸のオフセットを1つ\parencite{Wang_Elsevier_CVIU2005,Nagamatsu_ETRA2008}もしくは2つ\parencite{Beymer_CVPR2003,Villanueva_WS_IJPRAI2007,Nagamatsu_ETRA2010a,Chen_IEEE_TIP2015}が個人パラメータであり，稀な例として3つのパラメータ（光軸と視軸のオフセット1自由度，角膜曲率半径，瞳孔中心と角膜曲率中心間距離）の推定が必要なモデル\parencite{Nagamatsu_ETRA2010a,Nagamatsu_IEICE_TransInf2021}も存在する．
% 以上のように，モデルにより推定パラメータは異なるが，これらの眼球固有のパラメータは個人差や年齢とともに変化するため，ユーザごとにそれらを推定する必要がある．
% Even if the optical axis is obtained by a model-based method, at a minimum a rotational transformation from the optical axis to the visual axis by user calibration is required since the fovea is not directly observable from the image.
For calibration, most 3D models use one~\parencite{Wang_Elsevier_CVIU2005,Nagamatsu_ETRA2008} or two~\parencite{Beymer_CVPR2003,Villanueva_WS_IJPRAI2007,Nagamatsu_ETRA2010a,Chen_IEEE_TIP2015} DoF offset angles between the optical and visual axes.
In rare cases, models that require the estimation of three parameters (one DoF offset between the optical and visual axes, radius of corneal curvature, and distance between the pupil center and the center of corneal curvature) also exist~\parencite{Nagamatsu_ETRA2010a,Nagamatsu_IEICE_TransInf2021}.
In summary, each model has different individual parameters, which must be estimated for each user calibration.
Our method adopts two DoF offset angles as the calibration parameters.

\subsubsection{Appearance Models}
%----------------------------------------------------------------------
% 外観モデルの較正
%----------------------------------------------------------------------
% アピアランスベース手法\parencite{Williams_IEEE_CVPR2006,Wood_IEEE_ICCV2015,Zhang_IEEE_PAMI2019}では，明示的な特徴抽出無しに眼画像を入力として注視点座標を直接出力するように機械学習される．
% 機械学習アルゴリズムとしては，three-layer neural network~\parencite{Baluja_NIPS1993,Stiefelhagen_WPUI1997,Xu_BMVC1998}, Locally Linear Embedding\parencite{Tan_WACV2002}，Gaussian processes\parencite{Williams_IEEE_CVPR2006}，and deep neural networks\parencite{Krafka_CVPR2016,Zhang_CVPR2015}が使用されている．
% 最新のCNNを用いた視線推定については\parencite{Akinyelu_IEEE_Access2020}にまとめれているが，詳細は論文の範囲を超える．
% 3次元CGで作成した大量のデータセットを用いて学習する手法\parencite{Wood_IEEE_ICCV2015}も提案されており，様々な外乱に対して頑健に視線を推定できる可能性がある\parencite{Kar_IEEE_Access2017}[85]．
In appearance models, an eye image is used without manually designed feature extraction and machine learning to directly output the gaze direction.
Several learning algorithms have been considered, including neural networks with three layers~\parencite{Baluja_NIPS1993,Stiefelhagen_WPUI1997,Xu_BMVC1998}, manifold learning~\parencite{Tan_WACV2002}, Gaussian processes~\parencite{Williams_CVPR2006}, and deep neural networks~\parencite{Zhang_CVPR2015,Krafka_CVPR2016}.
A complete description of the state-of-the-art convolutional neural network-based methods, summarized in \parencite{Akinyelu_IEEE_Access2020}, lies outside the scope of this study.

%----------------------------------------------------------------------
% \parencite{Hansen_IEEE_PAMI2010}内のアピアランスベース
% % 3層ニューラルネット
% [6]\parencite{Baluja_NIPS1993} % user-dependent 記述なし，offset table，2000点
% [133]\parencite{Stiefelhagen_WPUI1997} % user-dependent: 1.5-1.8 deg, multi-user: 1.9 deg
% [160]\parencite{Xu_BMVC1998} % user-dependent学習のみ 記述なし
% % manifold learning (locally linear embedding)
% [136]\parencite{Tan_WACV2002} % 3 subject, 0.38 deg, testでは自分の眼画像込み
% % Gaussian processes
% [157]\parencite{Williams_CVPR2006} % 16点較正で0.83 deg, user-dependentの話なし
%----------------------------------------------------------------------
% \parencite{Kar_IEEE_Access2017}内のアピアランスベース（CNN）
% [84] % 7方向に分類しただけ
% [85] % 学生のレポート？
% [86]\parencite{Krafka_CVPR2016} % CNNで多数の顔写真から視線推定　err 1.34cm/スマホの距離36.2cm\parencite{Bababekova_AAO_OVS2011}=2.11 deg
% [87]\parencite{Zhang_CVPR2015} % Suganoら．個人特化との区別あり．4degくらいの誤差
%----------------------------------------------------------------------
% 一般に，機械学習の性能は，トレーニングデータセットに依存する．
% これらの技術を較正不要な視線検出器として利用するためには，
% ユーザとは異なる人のデータセットで学習されていなければならない．
% しかし，眼球の外観からは中心窩の位置の個人差を予測することはできないため，
% 個人差の影響も含めて高精度に視線を推定するためには，ユーザ個人に特化したデータでトレーニングする\parencite{Stiefelhagen_WPUI1997,Zhang_CVPR2015}か，ユーザごとの補正量をマップ化したoffset table\parencite{Baluja_NIPS1993,Xu_BMVC1998}により精度補償する必要があり，これらには各ユーザに対して少なくとも1度は較正と同様な能動的な動作が必要となる．
% この区別を明確にしない文献が多く精度の単純比較が難しいが，一般に，3次元モデルベース手法や回帰ベース手法に比べて精度が低いと言われている\parencite{Kar_IEEE_Access2017}．
The performance of machine learning depends on the training dataset.
For these techniques to be used as calibration-free gaze estimators, they must be trained on a dataset that does not include the user's data.
However, individual differences in the position of the fovea cannot be predicted from the appearance of the eyeball.
For high-accuracy estimation of the individual visual axis, training with data specific to each individual user~\parencite{Stiefelhagen_WPUI1997,Zhang_CVPR2015} or compensating for accuracy using an offset table~\parencite{Baluja_NIPS1993,Xu_BMVC1998} that maps the amount of correction for each user is necessary.
Either of these requires at least one explicit operation similar to user calibration.
% Since several references do not accurately describe the use of such user-specific learning, it is difficult to make a simple comparison of accuracy. 
In general, appearance models are less accurate than 3D or regression models~\parencite{Kar_IEEE_Access2017}.

\subsubsection{Head-correlation Models}
% Learning-based methodsは，アイトラッカーを必要せずに視線を推定する．
% それらは，前提として，視線を計測しないので，キャリブレーションフリーである．
% Liらは，random regression treeによる視線推定モデルを提案した．
% このモデルは，頭部の動きと，手の位置をシーン画像から推定し，入力として用いる．
Head-correlation models are the challenging techniques that estimate the visual axis without observation of eyes~\parencite{Li_IEEE_CCV2013,Hu_IEEE_TVCG2019,Hu_IEEE_TVCG2020}.
% これらの手法の共通点は，頭部の姿勢と視線との連動制とシーン画像中の注目しやすい箇所を組み合わせて視軸方向を推定することである．
% シーンポイントを特定する手がかりとして，手先の位置やsaliency  mapが使用される．
% この手法に限っては，眼の画像や特徴を用いないため光軸から視軸へのオフセットを推定している訳ではない．
These models commonly estimate the direction of the visual axis by combining the head--eye coordination and the scene points that the user is likely to pay attention to.
The fingertip position~\parencite{Li_IEEE_CCV2013}, saliency map~\parencite{Hu_IEEE_TVCG2019,Hu_IEEE_TVCG2020}, and moving objects~\parencite{Hu_IEEE_TVCG2020} are used as cues to identify such scene points.
Calibration is not required for these models, as the visual axis is estimated frame by frame.
In principle, the offset can be determined by combining the visual axis estimated by these models and the optical axis obtained from the eye tracker.
That is, these models are applicable for self-calibration if the estimation accuracy is sufficiently high.
Unfortunately, the reported estimation errors exceed 7\Deg~\parencite{Li_IEEE_CCV2013,Hu_IEEE_TVCG2019,Hu_IEEE_TVCG2020}, which is an angle as large as or larger than the typical offset.

\subsection{Self-calibration Methods}
\label{sec:auto-calib}
% 自動較正とは，ユーザが較正を意識することや，較正のための特殊なパターンを表示することなく，前節で述べた何れかのモデルのパラメータを推定することである．
% 自動較正手法は，主に次の３つのアプローチが存在する：Saliencyベース手法，smooth pursuitベース手法，シーンモデルベース手法．
Self-calibration is a technique that estimates the offset angles without the placement of any special objects or without the user being aware of the process.
Table~\ref{table:auto-calib} summarizes the features of conventional self-calibration methods.
Three main approaches to self-calibration exist: saliency-based, smooth pursuit-based, and scene model-based methods.

\def\grayrow{\rowcolor[rgb]{0.95, 0.95, 0.95}}
\begin{table}[!ht]
\centering
\caption{Comparison of Self-calibration Methods}
\label{table:auto-calib}
\scriptsize
% \small
% \normalsize
\begin{tabular}{llllllll}
%\hline\hline
\toprule
Methods$^1$ & 
    Base approaches & 
    Target scene & 
    Head pose & 
    Scene image & 
    Accuracy$^2$\\% &
% \hline
% \grayrow
% Li et al. (2013)\parencite{Li_IEEE_CCV2013}& 
%     learning & 
%     arbitrary scene & 
%     free & 
%     necessary &
%     8.4\Deg\\
% Hu et al. (2019)\parencite{Hu_IEEE_TVCG2019}& 
%     learning & 
%     arbitrary scene & 
%     free & 
%     necessary &
%     8.5\Deg\\
% \grayrow
% Hu et al. (2020)\parencite{Hu_IEEE_TVCG2020}& 
%     learning & 
%     arbitrary scene & 
%     free & 
%     necessary &
%     7.1\Deg\\
\hline
\grayrow
\cite{Sugano_IEEE_PAMI2013} & 
    saliency & 
    flat monitor & 
    fixed & 
    necessary &
    3.5\Deg\\ 
\cite{Chen_IEEE_CVPR2011,Chen_IEEE_TIP2015}& 
    saliency & 
    flat monitor$^3$ & 
    fixed & 
    necessary &
    3.4\Deg\\
\grayrow
\cite{Wang_ETRA2016} & 
    saliency & 
    flat monitor$^3$ & 
    free& 
    necessary & 
    1.0\Deg/1.4\Deg\\
\cite{Alnajar_IJCV2017} & 
    saliency & 
    flat monitor & 
    free& 
    necessary & 
    4.2\Deg\\
\grayrow
\cite{Hiroe_CHI2019} & 
    saliency & 
    flat monitor$^4$ & 
    fixed& 
    necessary & 
    1.71\Deg\\
\cite{Liu_IEEE_Access2020}& 
    saliency & 
    arbitrary scene & 
    free& 
    necessary & 
    3.7\Deg/4.0\Deg \\
\grayrow
\cite{Shi_Elsevier_CAG2020}& 
    saliency & 
    arbitrary scene & 
    free& 
    necessary & 
    3\Deg--4\Deg \\
\hline
%----------------------------------------------------------------------
\cite{Murauer_iWOAR2018}& 
    smooth pursuit & 
    arbitrary scene$^5$ & 
    free & 
    necessary & 
    6.17\Deg \\
\grayrow
\cite{Tamura_ETRA2020} & 
    smooth pursuit & 
    arbitrary scene$^5$ & 
    free& 
    necessary & 
    ---\\
%----------------------------------------------------------------------
\hline
\cite{Nagamatsu_ETRA2010b} & 
    scene model&
    flat monitor &
    free & 
    unnecessary & 
    1.58\Deg \\
\grayrow
\cite{Model_ETRA2010,Model_IEEE_TBE2010} & 
    scene model&
    flat monitor &
    free & 
    unnecessary &
    1.3\Deg\\
\cite{Maio_FG2011} &
    scene model&
    flat monitor &
    free & 
    unnecessary &
    2.4\Deg--8.9\Deg\\
\grayrow
\cite{Nagamatsu_PETMEI2013} & 
    scene model&
    infinity &
    free & 
    unnecessary &
    0.2\Deg\\
\cite{Miki_ETRA2016} & 
    scene model&
    flat monitor &
    free & 
    unnecessary &
    0.5\Deg~$^6$\\
\grayrow
\cite{Wang_Elsevier_PR2018} & 
    scene model&
    flat monitor &
    free & 
    unnecessary &
    1.3\Deg\\
\cite{Uramune_SIGMR202101} & 
    scene model & 
    continuous surface & 
    free & 
    unnecessary & 
    1.29\Deg\\
\grayrow
\bf Proposed method (optical axis)$^7$ &
    \bf scene model &
    \bf arbitrary scene &
    \bf free &
    \bf unnecessary & 
    \bf 2.12\Deg\\
\bf Proposed method (visual axis)$^8$ &
    \bf scene model &
    \bf arbitrary scene &
    \bf free &
    \bf unnecessary & 
    \bf 1.20\Deg\\
\bottomrule
\end{tabular}\\\vspace{2mm}
\begin{minipage}{0.85\textwidth}
% 1. 現実もしくは仮想のマーカを必要とせず，ユーザが無意識のうちに較正パラメータが推定される手法が列挙されている．
% 2. 精度は，論文ごとに評価法が異なるため単純比較はできないため，大凡の傾向を把握するための情報である．
\footnotesize
1. The self-calibration methods that do not require real or virtual markers are listed.\\
2. The accuracy evaluation methods are not standardized. Only a rough comparison is meaningful.\\
3. The scene is limited to a static scene image with low entropy.\\
4. The face of a person must be displayed.\\
5. Smooth pursuit-based methods rely on a few objects that move in correlation with the eye.\\
6. Real gaze data were not used in the experiments. \\
7. The optical axis was used as the initial calibration parameter for fixation detection. \\
8. The visual axis was used as the initial calibration parameter for fixation detection.
\end{minipage}
\end{table} % 比較表

\subsubsection{Saliency-based Methods}
% Suganoらにより最初に顕著性マップを用いた自動較正手法が提案された\parencite{Sugano_IEEE_PAMI2013}．
% この手法では，シーン変化が眼画像に影響しないという仮定のもと，互いに類似した眼画像に対応するシーンの顕著性マップを生成し，複数のマップを積算することで，顕著性の極大位置をPoRとして抽出する．
% 眼画像から顕著性の極大位置へのノンパラメトリックな変換をガウス過程回帰により推定する．
% しかし，眼画像のシーン非依存性は応用を限定するし，顕著性マップ積算の収束の遅さは欠点である．
Saliency-based methods extract the global maximum position of the saliency in the FOV as the visual axis.
Because a saliency map generated from a single-frame scene image is multi-modal, most saliency-based methods try to improve accuracy by accumulating multiple frames of saliency maps.
However, the accuracy of most methods with simple accumulation is limited to about 3\Deg~\parencite{Sugano_IEEE_PAMI2013,Chen_IEEE_CVPR2011}.
% Chenらは，目画像から直接スクリーン上のPoR位置への変換の代わりに，光軸と視軸のオフセットをパラメータとして推定する手法を提案した\parencite{Chen_IEEE_TIP2015}
To further improve accuracy and efficiency, several methods limit the scenes or frames to be accumulated to only those with low entropy~\parencite{Chen_IEEE_TIP2015,Wang_ETRA2016}, with faces~\parencite{Hiroe_CHI2019}, or with gaze data of many other people looking at the same scene~\parencite{Alnajar_IJCV2017}.
% Alnajar et al. proposed using the gaze data of a large number of people who were not the user to improve the accuracy of the visual axis estimation from the saliency maps~\parencite{Alnajar_IJCV2017}.
% However, this method cannot be applied to self-calibration in general situations because it requires that all participants be shown the same scene as the user.
However, such approaches limit the applications or situations. 
In all of the studies presented above, experiments were conducted with the participants seated in front of a flat monitor.

For 3D scenes, two methods have been proposed to calibrate a head-mounted eye tracker while the user is freely walking in a real~\parencite{Liu_IEEE_Access2020} or VR~\parencite{Shi_Elsevier_CAG2020} environment.
To more accurately detect the maximum saliency, these methods adopt multi-frame accumulation considering the 3D environments.
Liu et al.~\parencite{Liu_IEEE_Access2020} limited the area of saliency to the object area, and Shi et al.~\parencite{Shi_Elsevier_CAG2020} projected the saliency map of one eye onto the other eye via the 3D shape of the scene. 
Despite these improvements, accuracy of about 3\Deg or worse has been reported~\parencite{Liu_IEEE_Access2020,Shi_Elsevier_CAG2020}.
In addition, acquiring scene images to compute the saliency map frequently incurs additional high computational costs in VR systems.

\subsubsection{Smooth Pursuit-based Methods}
% 個人キャリブレーションを必要しないPoR推定技術として，Smooth Pursuitに基づくアプローチが登場した．
% こうした手法は，視野カメラから得られる画像上での特徴点の動きと眼球の動きの類似性から注視対象を特定することができる．
% 注視対象が判明すれば，視線検出器の較正に使用することができる\parencite{Murauer_iWOAR2018,Tamura_ETRA2020}
Smooth pursuit-based methods have emerged as PoR estimation techniques without user calibration.
They can also be applied to self-calibration after PoRs with high accuracy are obtained~\parencite{Velloso_DIS2016}.
Such methods can identify the target of the user's gaze based on the similarity between the motion of the eye and the motion of the target in the scene images.
Once the target is determined, the offset can be calculated by considering one point on the target object as the direction of the visual axis~\parencite{Murauer_iWOAR2018,Tamura_ETRA2020}.

% こうした手法は，画像上で眼球と類似する動きをする部分を1点に特定できることを前提としている．
% そのため，対象となる動物体が小さく，同じ動きをする物体が少数である必要がある．
% それが原因かどうかは不明であるが，このアプローチの精度は，4から6度程度と報告されている\parencite{Murauer_iWOAR2018}.
% 提案法においても，頭部移動中のfixationにより，smooth pursuiteが生じたときのPoRを利用しており原理的には近い．
% しかし，提案法では物体を特定する必要はない．
Smooth pursuit-based methods assume that it is possible to identify a single image point that moves on a similar trajectory as the PoR.
This requires the target object to be small, and no other object must have the same movement. 
The accuracy of these methods has been reported to be around 4\Deg to 6\Deg~\parencite{Murauer_iWOAR2018}.
These methods are similar in principle to our proposed method, which utilizes the PoRs during smooth pursuits caused by fixation; however, the proposed method does not require the identification of the object at which the user is gazing.

\subsubsection{Scene Model-based Methods}
% Nagamatsuらは，モニタ平面上における両眼の光軸の２交点間の中点を両眼のPoRとして算出する手法を提案した\parencite{Nagamatsu_ETRA2010b}．
% これは光軸-視軸間オフセットの水平成分を推定することに相当し，垂直成分のoffsetを補正することはできない．
% そこで，Modelらは，予測された平面画面上の両眼のPoR間の距離を最小化する各眼水平・垂直2自由度のオフセット角を推定しようとした\parencite{Model_ETRA2010}．
% この手法は，解法の安定化や効率化などの改良が図られている\parencite{Miki_ETRA2016,Wang_ITME2016}．
% 他にも，視軸の交点の一致という制約\parencite{Nagamatsu_ETRA2010b}に加えて，PoRがスクリーン内かつオフセット角の範囲を制約に用いたり\parencite{Maio_FG2011}，これらに加えて異なる手法の結果が一致するという拘束も提案されている\parencite{Wang_Elsevier_PR2018}．
% 以上の手法は，スクリーンというシーンオブジェクトと視線との交点を利用していたという意味で，シーンの３Dモデルを前提とした手法である．
% しかしながら，これらの手法は，シーンが近接の平面ディスプレイ~\parencite{Nagamatsu_ETRA2010b,Model_ETRA2016}か無限遠~\parencite{Nagamatsu_PETMEI2013}と限られている．
% 提案手法では，シーンの3次元形状が既知であることを前提とするため，シーンモデルベース手法と分類できる．
% シーンの3次元モデルはVR用途であれば必然的に利用可能である．
% Uramuneらも，提案法と同様にシーンの三次元モデルを用いてPoRの分散を利用した自動較正法を提案しているが，不連続な奥行きを持つシーンに対応していなかった~\parencite{Uramune_SIGMR202101}．
In scene model-based methods, the optical axis direction and the scene shape are used to determine the visual axis direction.
Nagamatsu et al. proposed a method to calculate a single PoR as the midpoint between the two intersections of the optical axes of both eyes on a monitor plane~\parencite{Nagamatsu_ETRA2010b}.
This method is equivalent to estimating the horizontal component of the offset between the optical and visual axes but cannot correct for the vertical component. 
Model et al. generalized this method to estimate the two DoF offset angles for each eye, minimizing the distance between the two intersections of the optical axes of both eyes on the planar screen~\parencite{Model_ETRA2010}.
This generalized method has been improved to make the solution more stable~\parencite{Miki_ETRA2016} and efficient~\parencite{Wang_ITME2016}.
In addition to the constraint of coincidence of the intersections of the two eyes' visual axes~\parencite{Nagamatsu_ETRA2010b}, other constraints have been indicated. 
Examples include the constraint that the PoR is within the screen and the range of the offset angles~\parencite{Maio_FG2011}, and the constraint that the results of the different methods coincide~\parencite{Wang_Elsevier_PR2018}.
In the methods mentioned above, however, the scene is limited to a flat monitor~\parencite{Nagamatsu_ETRA2010b,Model_ETRA2010} or to infinity~\parencite{Nagamatsu_PETMEI2013}.

Our proposed method can be classified as a scene model-based method in the sense that it assumes the 3D shape of the scene, which is inevitably available for VR applications.
Our previous work has also proposed a self-calibration method using the variance of PoRs on the surface of a 3D scene model~\parencite{Uramune_SIGMR202101}. 
However, although this method is similar to the proposed method, it does not support scenes with discontinuous depths or occlusions.
From the foregoing discussion in \hyperref[sec:auto-calib]{\it Self-calibration Methods}, it is evident that Assertion I has been substantiated.

\subsection{Fixation Detection Algorithms for 3D Scenes}
\label{sec:fixation detection algorithms}
% 固視は，中心視野の2～5 degの範囲内で，少なくとも80～100 msの間視線が一定時間静止する眼球運動（状態）である~\parencite{Hansen_IEEE_PAMI2010,Manor_2003}．
% 固視とは別に，物体の像を中心窩まで移動させるために，高速で動く眼球運動をサッケードと呼ぶ．
% サッケードは100～700~\Deg/sの速度を持つ~\parencite{Kar_IEEE_Access2017}．
% ターゲットや頭部が動く場合でも，smooth pursuit眼球運動や前庭動眼反射により固視が維持される．
% 眼球運動の詳細な解説は，\parencite{Carpenter_1988}を参照されると良い．
In this section, we first review the fundamentals of fixation detection; next, we provide a overview of detection algorithms.
During fixation, the gaze remains stationary for at least 80--100~ms within 2\Deg -- 5\Deg of the central visual field~\parencite{Hansen_IEEE_PAMI2010,Manor_NM2003}.
When the target object or the head is moving, fixation is maintained by smooth pursuit eye movements and the vestibulo-ocular reflex.
Rapid eye movements between two successive fixations are called saccades, which have velocities of 100 -- 700~\Deg/s~\parencite{Kar_IEEE_Access2017}.
A detailed explanation of eye movements is beyond the scope of this paper.
Interested readers may refer to~\parencite{Carpenter_1988} for a detailed description.

% 固視検出の重要性
The results of fixation detection depend on the detection algorithm and parameters such as threshold values~\parencite{Shic_ETRA2008, Blignaut_Springer_APAP2009}.
Previous studies~\parencite{Manor_NM2003,Poole_EHCI2006,Hooge_BRM2022} reported that the number of fixations and the average fixation time are significantly influenced by changes in the algorithm and threshold values.
In some cases, these changes affect the conclusions of later analyses~\parencite{Shic_ETRA2008}.
The optimization results of the proposed method also depend on the results of fixation detection, as shown in \hyperref[sec:Dependence on Fixation Detection]{\it Dependence on Initial Parameters}.
Thus, here we explain the need for algorithms suitable for fixation detection in 3D scenes.

% 視線の時系列データから固視を検出する手法は，検出する座標系の違いから二つに大別される：シーンカメラ座標上とオブジェクト座標\parencite{Salvucci_ETRA2000}．
% シーンカメラ座標上で検出する手法は，固視中の視線の角速度がサッケードに比べて十分遅いことを利用する．
% その代表的なアルゴリズムであるI-VT (Velocity-Threshold Identification)\parencite{Salvucci_ETRA2000}は，視線の角速度が閾値以下のとき，固視中の視線と判定する．
% この手法は，オブジェクトの情報が不要なため適用範囲が広い．
% しかしながら，オブジェクト座標上で検出する手法よりもサッケードや追跡眼球運動の一部である視線を固視と誤検出する頻度が高い\parencite{Salvucci_ETRA2000}．
Fixation detection algorithms from time-series gaze direction data can be divided into two main categories based on the reference coordinate systems: scene camera coordinate system and object coordinate system~\parencite{Salvucci_ETRA2000}.
The scene camera coordinate system involves a coordinate system fixed to the scene camera-- that is, the user's head.
If the head is fixed, there is little difference between the two systems; otherwise, their properties significantly differ.
The algorithms using scene camera coordinates takes advantage of the fact that the angular velocity of the gaze during fixation is sufficiently slow compared with that of the saccades.
A typical algorithm is I-VT (velocity threshold identification)~\parencite{Salvucci_ETRA2000}, which identifies gaze directions as a fixation if the angular velocity of successive gaze directions in the scene camera coordinate system is below a certain threshold.
This algorithm has a wide range of applications because it does not require object information and is almost independent of the initial calibration parameters.
However, it tends to produce more misdetections than algorithms using object coordinates~\parencite{Salvucci_ETRA2000}. 
This may be even more likely when head rotation occurs.

% オブジェクト座標上で検出する手法は，固視中の注視点間の距離が近いことを利用する．
% その代表的なアルゴリズムであるI-DT (Dispersion-Threshold Identification)\parencite{Salvucci_ETRA2000}は，オブジェクト座標における注視点の座標が一定時間以上，一定の範囲内にあるとき，その間の視線が固視と判定される．
% この手法は，他の眼球運動との誤検出が少ないが，オブジェクトの情報が必要である．
The algorithms that use object coordinates are based
on the fact that the dispersion of PoRs during a fixation is small. 
A typical algorithm is I-DT (dispersion threshold identification)~\parencite{Salvucci_ETRA2000}, which identifies PoRs as a fixation if the PoRs in the object coordinate system are within a certain range for a certain period of time or longer.
This algorithm is less prone to false positives from other eye movements.
% 従来の固視検出アルゴリズムは，平面ディスプレイでの視線計測が前提であり，頭部移動に適したアルゴリズムはほとんど提案されていない．
% シーンカメラ座標上で検出するアルゴリズムは，オブジェクトの情報を必要としないため，頭部装着型の視線計測においても，固視を検出することができる．
% 眼球運動に頭部の動きが加わるため，VRシーンのような頭部の移動が大きい状況下では，単一の速度閾値による検出では正しく固視を検出できない場合がある．
Larsson et al. have tried to develop an intermediate algorithm that detects fixations by thresholding the angular velocity in the world coordinate system with the head pose without object shapes~\parencite{Larsson_NM2003}.
This method allows head movements and arbitrary scenes.
However, a single velocity threshold can incorrectly detect fixation in situations where head movement is large, such as in our VR environments.
% 他にも，速度閾値と分散閾値を組み合わせたI-VDTや速度閾値と視線パターンを組み合わせた速度閾値と視線パターンを組み合わせたI-VMPも提案されている．
I-VDT~\parencite{Komogortsev_BRM2013} simply combines a velocity threshold with a dispersion threshold to achieve the intermediate properties of I-DT and I-VT.
% Another algorithm, I-VDT (velocity and dispersion threshold identification)~\parencite{Komogortsev_BRM2013}, which combines velocity and variance thresholds, has been proposed.
% , and I-VMP (velocity and movement pattern identification)~\parencite{Komogortsev_BRM2013}, which combines a velocity threshold with a gaze pattern.
In principle, I-DT and I-VDT are expected to be insensitive to both rotation and translation of the head, but neither of them is applicable to arbitrary scenes, as the dispersion of PoRs is defined only on a flat monitor.

% 一方で，オブジェクト座標上で検出する手法は，平面ディスプレイでPoRsの分散を定義する場合がほとんどであり，頭部装着型の視線計測には適用できない．
% 近年，Steilらは，シーン画像を用いることで頭部移動に対応した手法を提案した．
% この手法は，注視点近傍のシーン画像パッチの類似度を用いて固視を検出する．
% しかしながら，この手法はシーン画像に同じパターンが含まれている場合誤検出する可能性がある．
% また，シーン画像を用いないためこの手法を提案手法に採用することはできない．
% 本論文では，頭部移動中の3Dシーンにおいて，シーン画像を用いずに，オブジェクト座標でロバストに固視を検出する手法を提案する．
% 提案する固視検出アルゴリズムは，既存のI-DTを3Dシーンに拡張したものであり，詳細は次章に述べる．
% As mentioned above, most conventional algorithm for fixation detection assume gaze measurements are performed in front of a flat display, and 
Only few methods for detecting fixation for scenes with large head movements and arbitrary shapes have been proposed~\parencite{Steil_ETRA2018, Jurado_Sensors2020}.
One is to use pattern matching with the similarity of scene image patches around the PoRs to detect fixations~\parencite{Steil_ETRA2018}.
The disadvantages of this method are misdetections in texture-less and repeated-pattern regions and bandwidth-consuming image capturing.
% Juradoらは，頭部移動を許容した3次元シーンにおける固視検出アルゴリズムを提案した．
% この手法は，シーンの3Dモデルを必要とし，固視の候補に対して，平均化された頭部位置を原点として，注視点の分散を評価する．
% 我々が拡張したI-DTは，この手法とは独立であるが，似たアルゴリズムである．
Another method is to define the dispersion on a 3D object~\parencite{Jurado_Sensors2020}.
This method is similar to the algorithm we employ. 
% Jurado et al. proposed a fixation detection algorithm for 3D scenes that allows head movement.
% The method requires a 3D model of the scene, and evaluates the dispersion of the PoRs for the fixation candidates using the averaged head position as the origin.
% Our extended I-DT has been independently developed, but it similar to this method.
Although this method has been recently published, various studies have clarified the properties of I-DT and I-VT.
Therefore, in this study, we further extend the I-VDT method, which combines both I-DT and I-VT, to a 3D environment.
% Based on the above, this study adopts one method for detecting fixation on object coordinates: an extension of the I-VDT algorithm to 3D scenes. 
% The algorithm we adopt is an extension of the existing I-DT algorithm to 3D scenes and 
The details of the algorithm and its relationship with I-DT and I-VT are explained in the next sections.

\section{Method}
\label{sec:method}
\subsection{Preliminary Information}
\label{sec:preliminary}
%----------------------------------------------------------------------
% 視線，投影関数，シーンカメラ
%----------------------------------------------------------------------
% ここでは，提案法の説明に必要な前提条件やいくつかの変数を定義を述べる．
% 提案法では，HMD上では両眼に異なる画像が提示されていても，視線の計算では，cyclopean eyeを表す1台の仮想のシーンカメラ（ピンホールカメラ）を用いる．
% シーンカメラの焦点距離を1とすると，シーンカメラ画像面上の座標$(u,v)$の同次座標表現により，視線方向を$\bm g=(u, v, 1)$と表すことができる．
% オブジェクト座標系で記述されたPoRを$\bm X$とすると，
% シーンカメラ画像上に投影することで，視線ベクトル$\bm g = \pi(\bm X)$が得られる．
% ただし，$\pi(\bm X)$はオブジェクト座標からシーンカメラ画像へ投影する写像であり，視線計測と同一レートで取得できるものとする．
% 逆に，視線方向$\bm g$の延長線が最初にオブジェクト表面に衝突する点を$X$とすると，
% シーンカメラの逆投影関数$\pi^{-1}$を用いて$\bm X = \pi^{-1}(\bm g)$と表現する．
We first present our assumptions and define the variables necessary to the explanation of our proposed method. 
In this method, a single virtual scene camera (pinhole camera) representing the user's cyclopean eye is used for all our gaze processing, although different stereoscopic images are presented to both eyes on the HMD. 
Assuming that the focal length of the scene camera is 1, 
an arbitrary gaze direction can be expressed as $\bm g=[u, v, 1]^T$ using the homogeneous representation of the 2D coordinate $[u, v]^T$ on the image plane. 
If the position of a PoR represented as an object coordinate $\bm X$, 
% the projection onto the scene camera image yields the gaze vector $\bm g = \pi(\bm X)$, where $\pi$ is a mapping from the object coordinate $\bm X$ to the image coordinate.
the projection onto the scene camera image yields the following equation:
\begin{eqnarray}
\bm g = \pi(\bm X),
\end{eqnarray}
where $\pi$ is the mapping from the object coordinate $\bm X$ to the image coordinate.
$\pi$ can be obtained at the same acquisition rate as the gaze measurement.
Conversely, if $\bm X$ is the point where the extension of the line of sight in the direction $\bm g$ first collides with the object surface, it is expressed as $\bm X = \pi^{-1}(\bm g)$ using the inverse projection $\pi^{-1}$.

%----------------------------------------------------------------------
% 入出力
%----------------------------------------------------------------------
% 本手法において，入力として使用する情報は，離散時間$i$に対応する視線方向$g_i$と頭部の位置姿勢（投影 or 逆投影関数$\pi_i$ or \pi^{-1}$に対応）とシーン3Dモデルのみである．
% 3Dモデルのテクスチャは表示するのみであり，提案手法の計算には使用しない．
% 最終出力として求めるパラメータは，前節で述べた3D Eye modelを使用する．
% 較正前後の視線をそれぞれ$\bm g, \bm{\tilde g}$とすると，較正による視線の較正関数$f$は続く式で表わされる．
The inputs in this method include a set of gaze directions $\{\bm{g}_i\}$, the position and orientation of the HMD (corresponding to the projection $\{\pi_i\}$ or inverse projection $\{\pi_i^{-1}\}$) at $M$ discrete time frames $\ (i=1,2,\cdots,M)$, and a 3D environment model.
The model's texture is displayed but not used for the computation of the proposed method.
In the general 3D eye model described in \hyperref[sec:Calibration Models]{\it Gaze Estimation Models}, the transformation from the optical axis $\bm{g}_{\rm{opt}}$ to the visual axis $\bm{g}_{\rm{vis}}$ is presented by the following equation:
\begin{eqnarray}
\label{eq:opt_to_vis}
\bm{g}_{\rm{vis}} \hspace{-2mm} &\propto& \hspace{-2mm} f(\bm g_{\rm{opt}}\,;\,{\bm \theta}_{\rm{offset}}),\\
\label{eq:opt_to_vis2}
&=& \hspace{-2mm}
    \begin{bmatrix}
    1 & \hspace{-1.5mm} 0 & \hspace{-1.5mm} 0 \\
    0 & \hspace{1.5mm} \cos \beta_0 & \hspace{-1.5mm} \sin \beta_0 \\
    0 & \hspace{-1.5mm} -\sin \beta_0 & \hspace{-1.5mm} \cos \beta_0
    \end{bmatrix}
    \begin{bmatrix}
    \hspace{3mm} \cos \alpha_0 & \hspace{-1.5mm} 0 & \hspace{-1.5mm} \sin \alpha_0 \\
    0 & \hspace{-1.5mm} 1 & \hspace{-1.5mm} 0 \\
    -\sin \alpha_0 & \hspace{-1.5mm} 0 & \hspace{-1.5mm} \cos \alpha_0
    \end{bmatrix}
    \bm{g}_{\rm{opt}}, \hspace{4mm}
\end{eqnarray}
where $\propto$ is the symbol for proportionality in the homogeneous representation, indicating that both sides are equal in Cartesian representation.
$\bm \theta_{\rm{offset}}=[\alpha_0, \beta_0]$ is the offset angle addressed in Section~1.
$\alpha_0$ and $\beta_0$ are the horizontal and vertical components of the offset angle, respectively.

In this model, the axis directly measured by the eye tracker is not necessarily the optical axis, as long as it satisfies Equation~(\ref{eq:opt_to_vis2}), i.e., the axis only needs to be fixed to the eyeball. Although strictly speaking, the axis must pass through the center of rotation of the eyeball, but if the scale of the environment is large enough, a relatively small discrepancy between the center of rotation and the axis can be ignored.

The calibration parameters to be obtained as the final output are the estimated offset angles in the 3D eye model.
% Let $\bm{g}$ be the calibrated gaze direction; 
The calibration function $f$ mapping $\bm g_{\rm{opt}}$ to an arbitrary direction $\bm{g}$ can be expressed as the following equation:
\begin{eqnarray}
\label{eq:calib_func}
\bm{g} &\propto& f(\bm g_{\rm{opt}}\,;\,{\bm \theta}),
% \label{eq:calib_func2}
% &=& \hspace{-2mm}
%     \begin{bmatrix}
%     1 & 0 & 0 \\
%     0 & \hspace{3mm} \cos \beta & \sin \beta \\
%     0 & -\sin \beta & \cos \beta
%     \end{bmatrix}
%     \begin{bmatrix}
%     \hspace{3mm} \cos \alpha & 0 & \sin \alpha \\
%     0 & 1 & 0 \\
%     -\sin \alpha & 0 & \cos \alpha
%     \end{bmatrix}
%     \bm{g}_{\rm{opt}}, \hspace{4mm}
\end{eqnarray}
% ただし，$\propto$は同一点を表す同次座標表現の記号であり，両辺が比例関係にあることを表す．
% $\bm \theta=\{\alpha,\beta\}$は2個の較正パラメータの集合であり，$\bm \theta$には，視軸とのオフセットが対応する．
% \alpha, \betaはそれぞれ視線の水平，垂直方向の回転角度である．
where $\bm \theta=[\alpha,\beta]$ is a set of two arbitrary calibration parameters.
% , which corresponds to the offset addressed in Section~1.
% , and $\bm \theta$ corresponds to the offset from the visual axis.
$\alpha$ and $\beta$ are the horizontal and vertical rotation angles of the gaze direction, respectively.

% 提案手法は，上記入力情報から較正パラメータ$\bm \theta$を大きく分けて次の2つのステップで推定する：固視検出，較正パラメータの最適化．
% 次節では，まず，本論文で使用する三つの固視検出アルゴリズムについて詳述する．
% それから，最適化の評価指標である再投影誤差ついて導入し，最適化法について述べる．
The proposed method estimates the calibration parameter $\bm\theta$ using the above input information in two major steps: fixation detection and calibration parameter optimization.
In the next section, we explain in detail the extended I-VDT algorithm for fixation detection.
Then, in \hyperref[sec:optimization]{\it Optimization of Reprojection Errors}, we introduce the reprojection errors and describe the optimization method using these errors.

\subsection{Fixation Detection}
\label{sec:detection-algorithm}
% 固視検出結果は，アルゴリズムと検出時の較正パラメータによって変わる．
% 提案手法は，視線自動較正であり，視軸と異なる較正パラメータで固視検出する必要がある．
% 較正関数として3D eye modelを用いた場合，光軸で固視検出することが考えられる．
% 本論文では，3種類の固視検出アルゴリズムを用いて，提案手法の性能を評価する．
% 使用する固視検出アルゴリズムは，既存のI-VTに加えて，提案する3次元シーン用に拡張したI-DT，I-VDTである．
Fixations are detected using the I-VDT algorithm extended to a 3D environment.
Our extended I-VDT (referred to as 3D I-VDT) combines I-VT with the extended I-DT (referred to as 3D I-DT) to 3D scenes; also, the original I-VDT is a combination of I-VT and I-DT.
Therefore, this section describes the I-VT, 3D I-DT, and 3D I-VDT in order.

% 固視検出は視線方向の集合$\{\bm{g}_i\}$に対して実行される．
% \label{sec:optimization}節で述べる最適化では，固視中の視線集合を使用する．
% 固視検出アルゴリズムは，一つの固視と判定するための最小のフレーム数をもつ．
% $m$は，最小の停留時間とサンプリングレートで決定され，次式で表される．
% The I-VT uses a minimum number of frames
% The 3D I-DT and 3D I-VDT algorithms commonly perform the fixation detection for a set of gaze directions in a time sliding window with a variable size.
% whose parameter is $\bm \theta = [0, 0]$ (i.e. a set of optical axes).
%On the other hand, the optimization described in Section~\ref{sec:optimization} uses gaze directions during fixations with an arbitrary parameter.
These algorithms generally adopt a moving window for successive data frames to check for a fixation candidate.
The window has minimum $m$ frames that must satisfy certain conditions.
$m$ can be determined by the minimum fixation time $T_{\rm{th}}$ and the sampling rate $S_{\rm{r}}$:
 \begin{equation}
 m = S_{\rm{r}} T_{\rm{th}}.
 \label{eq:initial_window}
\end{equation}
%
% I-DTは，連続したデータ点に対して移動窓を使用して，固視の可能性を確認する．
% 移動窓は，アルゴリズムの開始時に，最小数のフレーム数$m$を含む．
% ここで移動窓のインデックス集合をGとおく．
% アルゴリズムは，窓内の視線データから，速度値や分散値を計算して，閾値と比較する．
% つまり，特定の閾値の条件を満たすときアルゴリズムは窓内の視線を固視と判定する．
% もし満たすなら，移動窓は次のフレームが含まれるように拡張され，閾値との比較が行われる．
% 途中で満たさなくなったら，移動窓はジャンプし，$\bm{G} = {i+1,\cdots,i+m}$となる．
% もしその条件が満たされない場合，窓内の視線は固視と判定されない．
In any algorithm, the movement of the window is as follows.
Initially, the window contains $m$ frames.
Let ${\bm{G}}$ be an index set or the window, which is initially set as ${\bm{G}}= \{i-m+1,\cdots,i\}$, where $i$ is the current frame.
If some criterion computed from the gaze data in the window satisfies a certain condition, 
the window is expanded to include the next frame, i.e., $\{i-m+1,\cdots,i+1\}$.
The window continues to expand as many times as the condition is satisfied.
If the condition is no longer satisfied owing to the window expansion, the gaze data in the window right before the expansion is labeled as the same fixation.
The window is shrunk to the original size $m$ and jumped so that the index set is updated to ${\bm{G} = \{i+1,\cdots,i+m\}}$.
If the condition is not satisfied without the expansion, all the frames in the window are identified as not a fixation; the window slides by one frame.
Hereafter, $k$ is used as the local index representing one frame in the window.

\subsubsection{Algorithm of I-VT}
% I-VTは，各視線方向に対して，速度値を計算する．
% 具体的には，現離散時間$i$に取得された視線$\bm g_i$が，次式を$m$フレーム以上連続して満たすとき，$\bm g_i$を固視中の視線と判定する．
The I-VT algorithm~\parencite{Salvucci_ETRA2000} detects fixations based on the angular velocity for each gaze direction.
In implementation, $\{\bm{g}_k \}$ is categorized as a fixational gaze
if any gaze direction $\bm{g}_k$ satisfies the following equation.
\begin{eqnarray}
\frac{\bm{g}_{k-1}\cdot \bm g_{k}}{|\bm g_{k-1}||\bm g_{k}|} &>& \cos\phi_{\rm{th}}\ \ \ (k-1, k\in \bm{G}),
\label{eq:fixation_ivt}
\end{eqnarray}
% ただし，$\phi$は角度の閾値，$\cdot$は内積を表す．
where $\phi_{\rm{th}}$ is the angle threshold, and `$\cdot$' represents the inner product.
% その結果，計測された全ての視線データに対して，複数の固視が検出される．
% ここで，$j$番目の固視における視線集合を$\bm{G}_j = \{\bm g_{0},\cdots,\bm g_{i},\cdots,\bm g_{M_j}\}$とする．
% 以下，各集合$\bm g_{j}$を固視クラスタと呼ぶ．
%
% \textcolor{red}{
% As a result, multiple fixations are detected for all gaze directions $\{\bm{g}_i\}$.
% Here, let $\bm{G}_j$ be a set of $M_j$ gaze directions $\{\bm{g}_{1},\cdots,\bm{g}_{M_j}\}$ in the $j$-th fixation.
% Hereafter, each set $\bm{G}_j$ is called a fixation cluster.
% }

\subsubsection{Algorithm of 3D I-DT}
% まず，従来のI-DTのアルゴリズムについて説明する．

% I-DT法は次に，移動窓に含まれる視線データから何らかの注視点の分散を計算する．
% 最も離れた注視点間の距離や，平均注視点からの最大距離，標準偏差など複数の分散の計算法がある．
% 従来のI-DT法~\parencite{Salvucci_ETRA2000}は平面ディスプレイに限定した手法であり，注視点の分散は二次元平面内で計算される．
%First, we explain the conventional I-DT algorithm~\parencite{Salvucci_ETRA2000}.
The original I-DT algorithm~\parencite{Salvucci_ETRA2000} calculates a dispersion of the PoRs from the gaze data contained in the window $\bm{G}$.
Dispersion can be calculated using several methods, such as the distance between the most distant PoRs, the largest distance from the center of the PoRs to any PoR, and the standard deviation~\parencite{Blignaut_Springer_APAP2009}.
The conventional I-DT is limited to a flat monitor, and the dispersion of the PoRs is calculated on a 2D plane.

% 本論文では，3次元シーンかつ頭部移動中での視線データに対して固視を検出するために，I-DT法を拡張する．
% 3次元シーンでは，注視点の分布のサイズは，シーンカメラとオブジェクトの距離に依存する．
% また，単純な方法で計算された3D空間の分散では，固視にシーン中のオクルージョンが含まれる場合，極端に分散値が大きくなる．
% 拡張I-DTは，注視点の分布のサイズによる影響を受けないようにするために，一度シーンカメラに投影して分散を計算する．
To extend the I-DT to 3D scenes with head movements,
% In a 3D scene, the scale of the distribution of the PoRs depends on the depth from the scene camera to the object surface where the PoRs are distributed, as well as on the surface normal.
% In addition, a dispersion in a 3D space computed on the object surface by the original I-DT may have an extremely large dispersion~\parencite{Uramune_SIGMR202101} when the PoRs are distributed across occlusions in the scene.
the 3D~I-DT calculates a 2D dispersion after the PoRs are projected onto a scene camera.
% そこで提案手法では，移動窓内に含まれるシーンカメラから，$i_c$番目のカメラを中央部分に存在するセンターカメラとして選定する．
% 私たちは，PoRsが投影されるセンターカメラを移動窓に含まれるシーンカメラから選定する．
% $i_c$番目のカメラをセンターカメラとする選択基準は次のとおりである．
For the projection, we select a center camera from the multiple scene cameras corresponding to window $\bm{G}$.
The center camera represents an average head posture.
Because the window size is about a few hundred milliseconds and the change in head posture is small, a camera with an average gaze direction is selected as the center camera.
For gaze directions $\{\bm{g}_k | k\in \bm{G}\}$, the selection criterion to determine the $k_c$-th camera as the center camera is as follows:
\begin{eqnarray}
\label{eq:center-camera1}
\bm{g}^{\prime}_k  &\propto& \bm{\mathrm{R}}^{-1}_k \bm{g}_k, \\
\label{eq:center-camera2}
k_c  &=& \argmin_{k\in \bm{G}} |\bm{g}^{\prime}_k - \bar{\bm{g'}}|,
\end{eqnarray}
% where $\bar{\bm{g}}$ is the average gaze direction defined as:
% \begin{eqnarray}
% \bar{\bm{g}} = \frac{1}{m}\sum_{k\in \bm{G}} \bm{g}^{\prime}_k.
% \end{eqnarray}
where $\bar{\bm{g'}}$ is the sample mean of $\{\bm{g'}_k | k\in \bm{G}\}$. 
% $\bm{g}'_i$は，オブジェクト座標で表された視線であり，大きさが1に正規化されたものである．
% 図\ref{fig:reprojection}に示すように，3D I-DTの分散は，提案手法のコスト関数で使用する再投影誤差と同義である．
% 移動窓内のPoRsが続く式を満たすとき固視と判定する．
$\bm{g}_k$ and $\bm{g}^{\prime}_k$ are the gaze direction vectors expressed in the scene camera and object coordinates, respectively, and both are normalized so that their magnitude is 1.
$\bm{\mathrm{R}}_k$ is the rotation matrix from the scene camera coordinates to the object coordinates and is the rotational component of the inverse projection function $\pi_k^{-1}$.

\begin{figure}[t]
\begin{center}
\includegraphics[width=0.7\columnwidth]{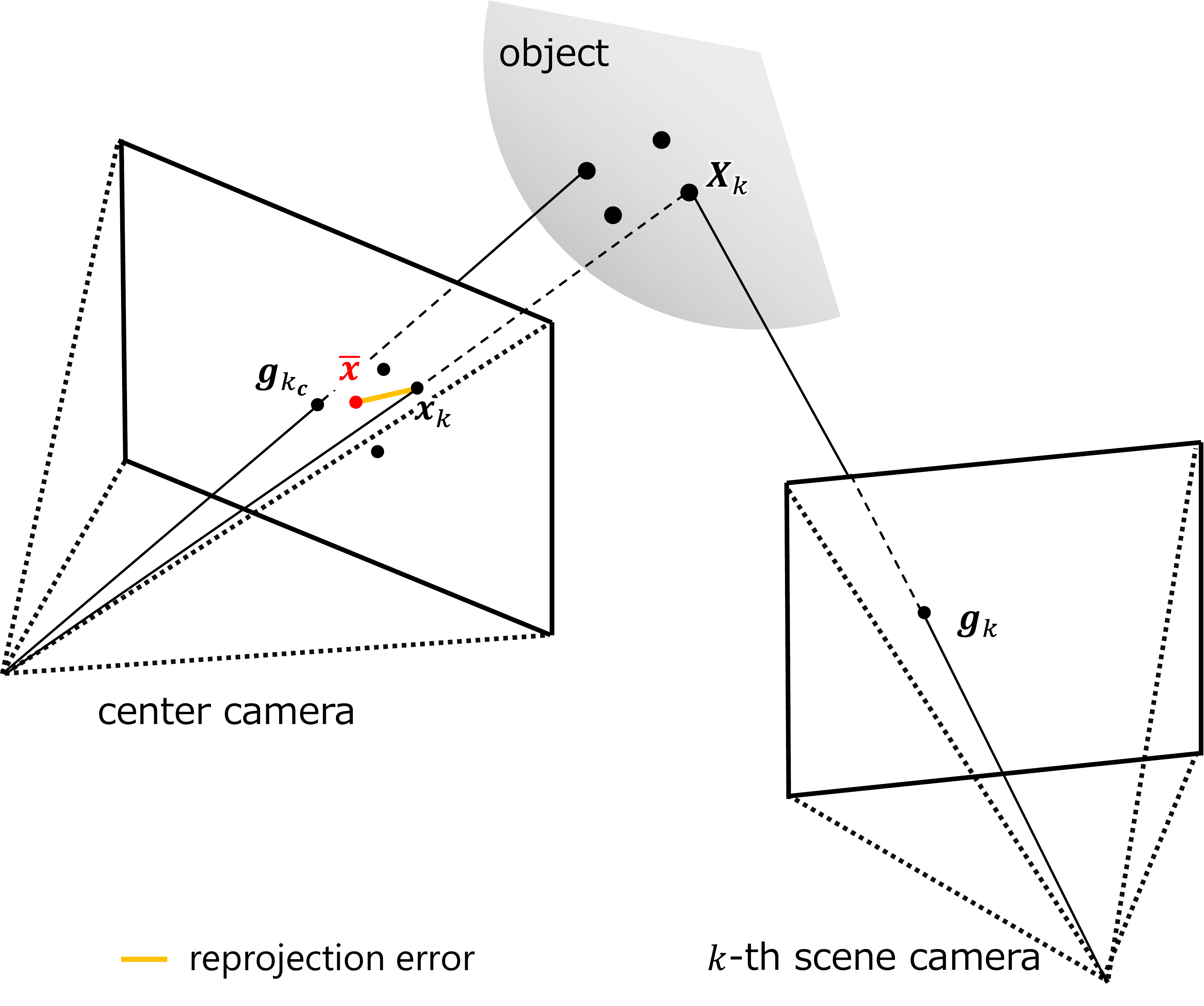}
\end{center}
\vspace{-4mm}
\caption{Reprojection errors.
In the proposed method, the calibration parameters are estimated by optimizing a cost function based on the reprojection errors.}
\label{fig:reprojection}
\end{figure}

%As shown in Fig.~\ref{fig:reprojection}, the dispersion of the 3D I-DT is equivalent to the reprojection error used in the cost function of the proposed method.
The dispersion is evaluated with reprojection errors on the image plane of the center camera in 3D I-DT. 
Fig.~\ref{fig:reprojection} shows a reprojection error as the 2D distance between a projected PoR and the average point of all the PoRs on the center camera image.
The 3D I-DT algorithm first calculates the 3D position of the PoR ${\bm{X}}_{k}$ as the inverse projection of the gaze direction $\bm{g}_k$:
\begin{eqnarray}
{\bm{X}}_{k}&=&\pi_k^{-1}(\bm{g}_k).
\end{eqnarray}
Next, the PoR ${\bm{X}}_{k}$ is projected to the center camera to obtain the reprojection point ${\bm x}_k$ using the following equation: 
\begin{eqnarray}
\bm x_k &=& \pi_{k_c}({\bm X}_{k}).
\label{eq:center-fixation-point1-1}
\end{eqnarray}
This projection prevents extremely large evaluation values even for PoRs scattered in the depth direction near the occlusion edge of the object.
% Then, the average position $\bar{\bm{x}}$ of ${\bm{X}}_{k}$ can be calculated by
% \begin{eqnarray}
% \bar{\bm{x}} &=& \frac{1}{m}\sum_{k\in\bm{G}} \bm x_k.
% \label{eq:center-fixation-point1-2}
% \end{eqnarray}
Finally, the gaze directions $\{\bm{g}_k | k\in\bm{G}\}$ are labeled as a single fixation if the reprojected PoRs satisfy the following equation:
\begin{eqnarray}
\argmax_{k\in \bm{G}} |\bm x_k - \bm{\bar x_{G}}|^2 < D_{\rm{th}},
\label{eq:fixation_idt}
\end{eqnarray}
%where:
%\begin{eqnarray}
%\bm x_k &=& \pi_{k_c}({\bm X}_{k}), \\
%\bar{\bm{x}} &=& \frac{1}{m}\sum_{k=i-m+1}^{i} \bm x_k.
%\label{eq:center-fixation-point1}
%\end{eqnarray}
% ただし，$D_{\rm{th}}$は分散閾値である．
% 上の式を満たさない場合，移動窓内の視線データは固視でないと判定され，移動窓は1離散時刻移動する．
% 満たす場合は，移動窓が次のフレームを含むように拡大され，再びセンターカメラと分散が計算される．
% 移動窓の分散が閾値を超えるまで拡大され続ける．
% 以上の手順により，移動窓をデータ点が無くなるまで移動させて，固視を検出する．
% 以下，誤解の恐れがない限り，拡張I-DT法を単にI-DT法と呼ぶ．
where $\bm{\bar x_{G}}$ is the sample mean of $\{\bm{x}_k | k\in\bm{G}\}$, and $D_{\rm{th}}$ is the dispersion threshold.
% where $D_{\rm{th}}$ is the dispersion threshold.
% If Equation~(\ref{eq:fixation_idt}) is not satisfied, the PoRs in the window are identified as not a fixation and the window slides by one frames.
% If satisfied, the moving window is expanded to include the next frame ($m$ is incremented), and the center camera and the dispersion are computed again.
% The moving window continues to expand until Equation~(\ref{eq:fixation_idt}) is no longer satisfied.
% \textcolor{red}{
% As a result, multiple fixation clusters are detected.
% As in the I-VT algorithm, let $\bm{G}_j$ denote the $j$-th fixation cluster.
% }

\subsubsection{Algorithm of 3D I-VDT}
% より正確に検出するために，I-VDT法は，I-VTとI-DTを組み合わせた固視検出アルゴリズムである．
% 3D I-VDTは，速度閾値と分散閾値の両方を用いる．
% I-DTと同様に，移動窓を用いる．
% 移動窓内のすべての視線が式(\ref{eq:fixation_threashold1})と，式(\ref{eq:fixation_threashold4})を満たすとき，I-VDTは視線を固視と判定する．
% I-DTと同様の手順で，移動窓をデータ点が無くなるまで移動させて，固視を検出する．
The I-VDT algorithm is a fixation detection algorithm that combines I-VT and I-DT.
% for more accurate detection.
3D I-VDT uses both a velocity threshold $\phi_{\rm{th}}$ and a dispersion threshold $D_{\rm{th}}$.
The same moving window as in 3D I-DT is used.
If all the gaze directions in the window simultaneously satisfy both  Equations~(\ref{eq:fixation_ivt}) and (\ref{eq:fixation_idt}), 
I-VDT categorizes the gaze directions $\{\bm{g}_k | k\in\bm{G}\}$ as a fixation.
% The window is continually expanded to include the next frame until these equations are no longer satisfied.
% Similar to the algorithms above, multiple fixation clusters are detected and the $j$-th fixation cluster is defined as $\bm{G}_j$.
% Using the same procedure as for 3D I-DT, the window is moved until there are no more data points.
% 以下，誤解の恐れがない限り，3D I-DT，3D I-VDTを単にI-DT，I-VDTと呼ぶ．

Both 3D I-DT and 3D I-VDT algorithms work equivalently to the original I-DT and I-VDT, respectively, under the fixed head condition.
For simplicity, we abbreviate below the 3D I-DT and 3D I-VDT algorithms as I-DT and I-VDT, respectively, unless there is a risk of misunderstanding. 
In addition, let $N$ fixations be obtained using the abovementioned I-VDT algorithm and $\bm{G}_j$ be the set of frame numbers corresponding to the $j$-th detected fixation that comprises of $M_j$ frames.
Hereafter, we call each set a fixation cluster.

\subsection{Optimization of Reprojection Errors}
\label{sec:optimization}
%----------------------------------------------------------------------
% 代表カメラ
%----------------------------------------------------------------------
% 本節の前半部では，最適化指標である再投影誤差を定義し，後半部では，最適化手法について説明する．
% 再投影誤差関数は，各固視クラスタに対して代表となるセンターカメラ上で定義される．
% まず，式(\ref{eq:center-camera1})，(\ref{eq:center-camera2})を用いて，$j$番目の固視クラスタに属する$M_j$個のシーンカメラから，$i_c$番目のカメラをセンターカメラとして選出する．
% 次に，固視クラスタ$\bm{G}_j$に対する再投影誤差は，次のように定義される．
In the first part of this section, we define the optimization metric, reprojection error; in the second part, we describe the optimization method.
As mentioned in \hyperref[sec:detection-algorithm]{\it Fixation Detection}, the reprojection error is defined on the center camera that is representative of each fixation cluster, as in Fig.~\ref{fig:reprojection}.
Although the definition is the same, and we refer to the same figure as in the description of fixation detection, the projection and inverse projection in the optimization are slightly different from those in I-DT: the parameters in the process of optimization are used for both projection and back projection.

The optimization metric is determined as follows.
First, we select the $k_c$-th camera as the center camera from the $M_j$ scene cameras belonging to the $j$-th fixation cluster using Equations~(\ref{eq:center-camera1}) and (\ref{eq:center-camera2}).
The reprojection error function $E_j$ of the $j$-th fixation cluster $\bm{G}_j$ is the mean of squares of the reprojection errors.
% , which can be computed using the same procedure as in Fig.~\ref{fig:reprojection}.
$E_j$ is defined by the following equation:
\begin{eqnarray}
E_j(\bm\theta)&=&\frac{1}{M_j}\sum_{k\in \bm{G}_j} |{\bm{x}}_{k} - \bm{\bar x}_{\bm{G}_j}|^2,
\label{eq:reprojection}
\end{eqnarray}
where $\bm{\bar x}_{\bm{G}_j}$ is the sample mean of $\{\bm{x}_k | k\in\bm{G}_j\}$.
% \begin{eqnarray}
% \label{eq:center-fixation-point2-1}
% {\bm{x}}_{k} &=& \pi_{k_c}({\bm{X}}_{k}), \\
% \label{eq:center-fixation-point2-2}
% \bar{\bm{x}} &=& \frac{1}{M_j}\sum_{k\in\bm{G}_j} {\bm{x}}_{k}.
% \end{eqnarray}
% PoRは次のように定義される．
That is to say, $E_j$ is equivalent to the 2D variance of the set of the PoRs $\{{\bm{X}}_{i}\}$ on the image plane.
The 3D point ${\bm{X}}_{k}$ is calculated as the inverse-projection point of the calibrated gaze direction $\bm{g}_k$:
\begin{eqnarray}
{\bm{X}}_{k}&=&\pi_k^{-1}(\,f({\bm g}_k\,;\,{\bm \theta})\,).
\label{eq:projection}
\end{eqnarray}
% \textcolor{red}{where $\bm{g}_{i}$ has the calibration parameters $\bm\theta$ as variables.}
% 以上のように，$N_j$個の視線$\bm g_i$からなる固視クラスタ$\bm{G}_j$の再投影誤差は，較正パラメータ$\bm \theta$に加えて$N_j$個の代表カメラ上の2次元位置$x_j$をパラメータとした最適化指標である．
Thus, we can find $E_j$ of cluster $\bm{G}_j$ is a function of $\bm\theta$.

% 最適化では，複数の固視クラスタによる再投影誤差を最小化する．
% $N$個の固視クラスタを用いて次の評価関数$E$によりパラメータ$\bm\theta$を最適化する．
Via optimization, the sum of the reprojection error functions for multiple fixation clusters is minimized.
The cost function simultaneously minimizes the reprojection error function for $N$ fixation clusters to estimate the optimal calibration parameters~$\hat{\bm\theta}$:
\begin{equation}
\hat{\bm\theta} = \argmin_{\bm \theta} \sum_{j=1}^{N} E_j(\bm{\theta}).
\label{eq:loss_func}
\end{equation}

%----------------------------------------------------------------------
% 最適化アルゴリズムの説明
%----------------------------------------------------------------------
% この評価関数には複数の局所解が存在するため大域最適化アルゴリズムを用いる必要がある．
% 良いパラメータを見つける方法として，differential evolutionが考えられる．
% この最適化手法は，進化計算アルゴリズムの一種であり，目的関数に微分可能性や凸性の仮定を必要とせず，大域最適解を探索できる．
% 提案手法では，differential evolutionを用いてパラメータの最適化を行う．
% 先にも述べたが，光軸と視軸は水平約4から5deg，垂直約1.5degのオフセットがあるため，$-5\leq\alpha\leq5, -5\leq\beta\leq5$の範囲を探索する．
% 探索性能を向上させるために，パラメータ空間を16分割してそれぞれの範囲でdifferential evolutionで最適化を行う．
% 最終的な結果には，最も評価値が最小であるパラメータを採用する．
This function has multiple local minima.
Therefore, a global optimization algorithm is required.
In our method, differential evolution~\parencite{Storn_GO1997} is applied as one possible approach for finding optimal parameters.
This optimization method is a type of evolutionary computation algorithm that does not require differentiability or convexity assumptions for the objective function and can search for the global optimal solution.
% In the proposed method, differential evolution is used to optimize the parameters.
%As mentioned in Section~1, the optical and visual axes are offset by approximately 4 to 5\Deg~horizontally and 1.5\Deg~vertically;
According to the offset range of approximately 4\Deg to 7\Deg, as mentioned in Section~1, the method searches the range $-5\leq\alpha\leq5, -5\leq\beta\leq5$ in degrees.
In this range, the maximum offset is 7.1\Deg.
To improve search performance, the parameter space is divided into 4~$\times$~4 square regions, and differential evolution is applied in each region.
For the final result, the parameters with minimum cost are adopted.
The preceding discourse in \hyperref[sec:detection-algorithm]{\it Fixation Detection} unequivocally demonstrates Assertion II.

\section{Experiments}
\label{sec:experiments}
%----------------------------------------------------------------------
% 実験で確認すること
%----------------------------------------------------------------------
% 本実験では，オクルージョンを含む一般的なシーンで提案手法の有効性を示すために，2種類のVRシーンで，提案手法を評価した．
% まず，提案手法の精度と，移動距離による提案手法の収束性能を確認する．
% 次に，提案手法の精度が，固視検出アルゴリズムと検出に必要なパラメータに依存することを示す．
This section demonstrates the performance of the proposed method by presenting three experiments using a common gaze dataset acquired in two VR environments.
The common experimental setup is described in detail in \hyperref[sec:experimental setup]{\it Experimental Setup}.
In \hyperref[sec:Accuracy Evaluation]{\it Accuracy Evaluation}, we show the accuracy of the proposed method. 
\hyperref[sec:Convergence Performance]{\it Convergence Performance} confirms the convergence performance with the walking distance.
Finally, we demonstrate the effect of the accuracy of the initial parameters on the estimation accuracy in \hyperref[sec:Dependence on Fixation Detection]{\it Dependence on Initial Parameters}.
All the experiments were approved by the ethics committee of the Graduate School of Engineering Science, Osaka University (R2-9) and, the participants gave informed consent.

%----------------------------------------------------------------------
% 実験環境
%----------------------------------------------------------------------
\subsection{Experimental Setup}
\label{sec:experimental setup}
% 本節では，実験に使用したデバイスと固視検出の閾値について，精度評価方法，被験者実験を詳細に説明する．
In this section, we explain the setup for acquiring gaze data used in the three experiments.

\subsubsection{HMD and Eye Tracker}
% HMDにはHTC Vive Pro Eyeを用いた．
% HMDの解像度は片目あたり1440 x 1600ピクセル（合計2880 x 1600ピクセル），
% リフレッシュレートは90~Hz，視野角110 degである．
% レンダリングはゲームエンジンのUnityで実装し，CPUはIntel Core i7-9700K，GPUはNVIDIA GeForce RTX 2080 SUPERを使用した．
HTC Vive Pro Eye was used as an HMD. 
The resolution of the HMD is 1440~$\times$~1600 pixels per eye (total 2880~$\times$~1600 pixels), the refresh rate is 90~Hz, and the viewing angle is 110\Deg.
Rendering was implemented in the game engine Unity on a PC 
(CPU: Intel Core i7-9700K, GPU: NVIDIA GeForce RTX 2080 SUPER).

% 視線データは，HMD内蔵の視線計測器 (120~Hz) を使用して取得し，実際の取得レートは約50~Hzであった．
% 頭部の位置姿勢は，HMD付属のLighthouseトラッキングシステムにより取得し，VRシーン提示中の取得レートは約50~Hzであった．
% 視線計測と頭部の位置姿勢の取得レートが仕様より低いのは，VRシーンの描画レートを優先させたためである．
% 上記のHMD内蔵の視線計測器を使用する場合，システム標準のアプリケーションソフトにより能動的な較正を行わなければ視線計測できない仕様となっている．
% 本実験では，視線計測器使用者と異なる人物でシステム内の能動的な較正を行い，被験者はシステムによって較正をしない．
Gaze data were acquired using the HMD's built-in eye tracker 
%(acquisition rate as per specifications: 120~Hz), 
with an actual acquisition rate of approximately 50~Hz.
The head position and posture were acquired by the Lighthouse tracking system attached to the HMD, and the acquisition rate during the VR scene presentation was approximately 50~Hz.
%The acquisition rates of the gaze measurement and the head position and posture are lower than those mentioned in the specifications because the VR scene rendering rate is prioritized.
% When using the HMD with a built-in gaze tracker, 
% gaze measurement is not possible without active user calibration using the standard application software. 
% In our experiment, the system was preliminary calibrated by a person different from our participants for the experiments.

% 本実験では，視線データ取得用のソフトウェアライブラリとして，SRanipal SDKを使用した．
% このライブラリにより得られる両眼の瞳孔中心の中点を起点として両眼の視線の平均を使用者の視線として用いた．
% また，同SDKでは，瞼の開閉度合いを0（完全に閉じている状態）から1（大きく開いている状態）の範囲で取得できる．
% 瞬きを含むデータを除くため0.5以下の視線データは排除した．
We used the SRanipal SDK as a software library for acquiring gaze data. 
The average of the gazes of both eyes starting from the midpoint of both pupil centers obtained from this library was used as the user's gaze. 
In addition, using the SRanipal SDK, we can obtain the degree of eyelid opening and closing (called openness) in the range from 0.0 (completely closed) to 1.0 (wide open).
To exclude blinks, we eliminated the gaze directions with the openness below 0.5.

\subsubsection{VR Environments} 
\begin{figure}[t]
    \centering
    \begin{minipage}{0.49\columnwidth}
    \centering
        \includegraphics[width=\textwidth]{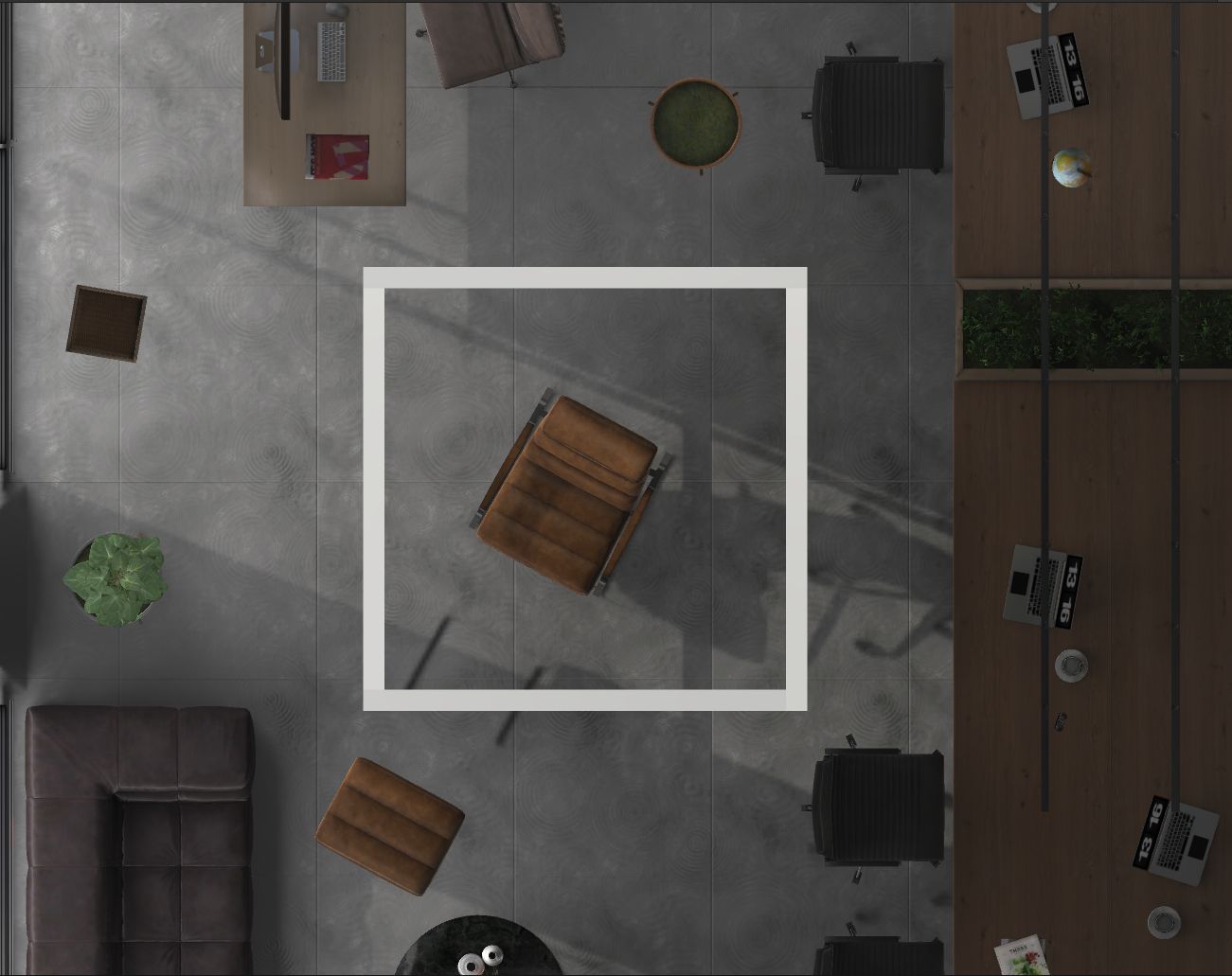}\\
        \vspace{1mm}
        \includegraphics[width=\textwidth]{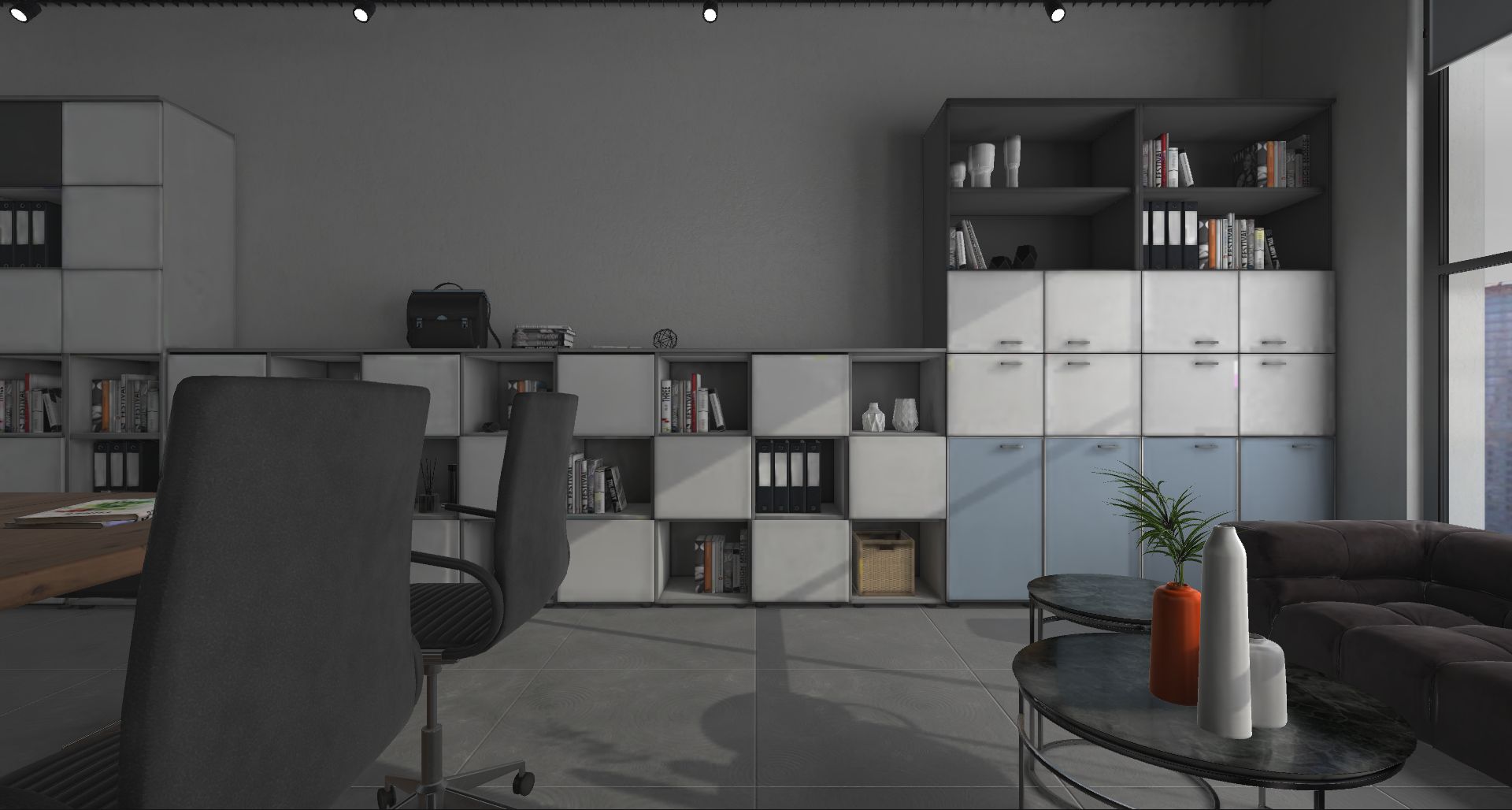}\\
        (a) Office

    \end{minipage}
    \begin{minipage}{0.49\columnwidth}
    \centering
        \includegraphics[width=\textwidth]{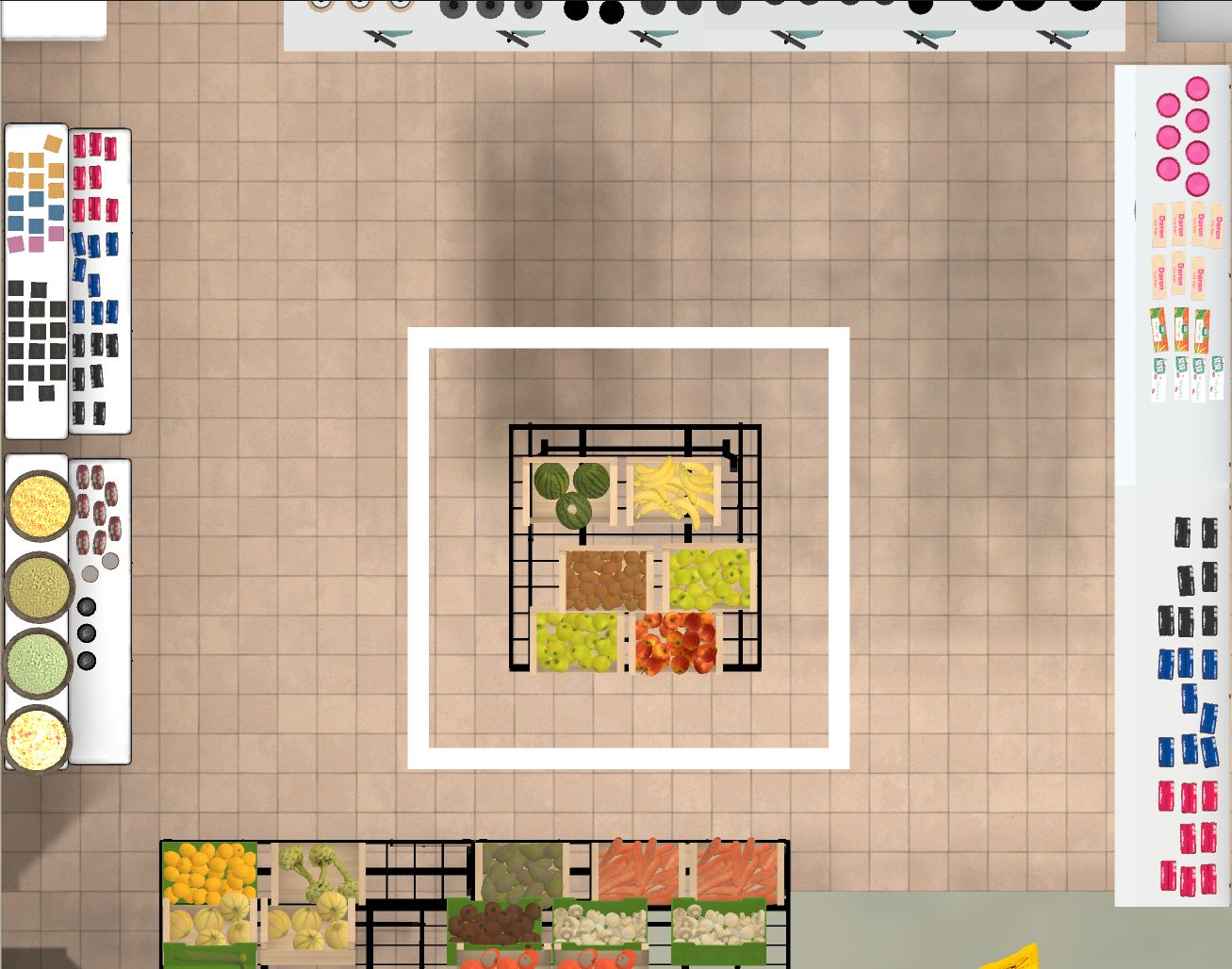}\\
        \vspace{1mm}
        \includegraphics[width=\textwidth]{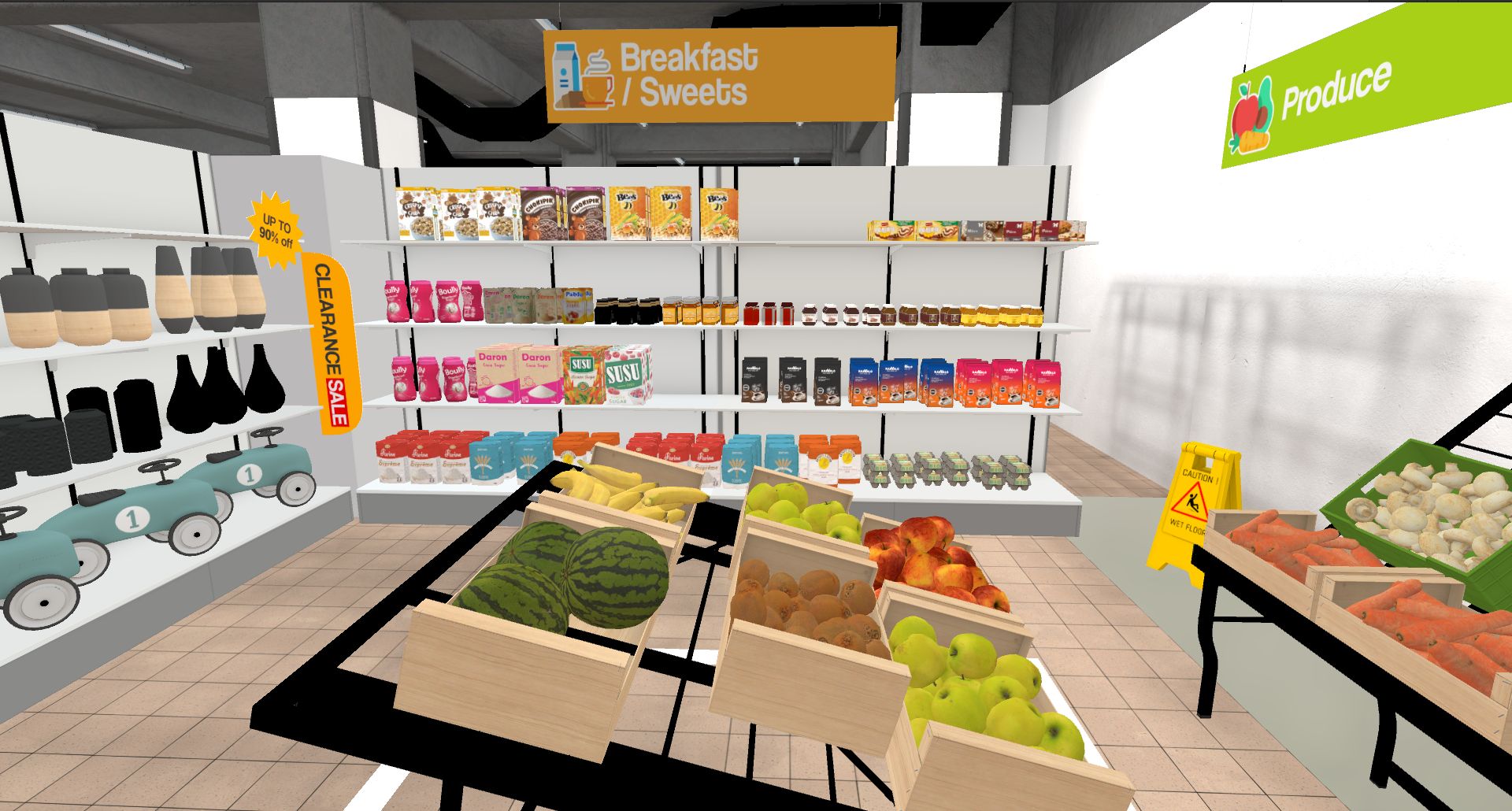}\\
        (b) Supermarket

    \end{minipage}

    \caption{VR environments.
Each participant was asked to walk along the white outline, counterclockwise and clockwise, twice each, for a total of four consecutive laps, starting from the upper right corner.}
    \label{fig:vr-scenes}
\end{figure}

%----------------------------------------------------------------------
% シーン
%----------------------------------------------------------------------
% VRシーンとしては，Unity Asset Store上で入手可能なModern Supermarket\footnote{\url{https://assetstore.unity.com/packages/3d/environments/modern-supermarket-186122}}と，Office\footnote{\url{https://assetstore.unity.com/packages/3d/environments/urban/office-interior-archviz-155701}}を一部加工して構築した．
% 図~\ref{fig:err1}に使用したVRシーンを示す．
% 安全性と歩くためのガイドのために，床面には2$\times$2mの正方形の白線が引かれており，白線上には仮想物を置おかなかった．
% 実環境は，白線の内側と周囲１mには何も物がない水平な床面だった．
``Office Interior Archviz\footnote{\url{https://assetstore.unity.com/packages/3d/environments/urban/office-interior-archviz-155701}}'' and ``Modern Supermarket\footnote{\url{https://assetstore.unity.com/packages/3d/environments/modern-supermarket-186122}},'' which are available on the Unity Asset Store, were partially modified for use as VR environments.
Here, they are simply named ``Office'' and ``Supermarket,'' respectively.
Fig.~\ref{fig:vr-scenes} shows these two VR environments.
For safety and guiding purposes, a 2~m white square outline was present on the floor, and no virtual objects were placed in the area around the white outline.
The real environment was a leveled floor with no objects inside or around 1~m of the square.

\subsubsection{Participants}
%----------------------------------------------------------------------
% 被験者
%----------------------------------------------------------------------
% 実験には，裸眼かコンタクトレンズでの視力が0.5以上（その他視覚障害なし）である22名の20代大学生・大学院生が参加した．
% 加えて，実験中もしくは実験後に，ベースライン推定用のデータを取得時に片方の眼のopennessが0.5以下であることが判明した場合は，その被験者のデータは使用しなかった．
% その結果，18名分の視線データが得られた．
Twenty-two undergraduate and graduate students with decimal visual acuity of 0.5 or better (no other visual impairment), either with naked eyes or contact lenses, participated in the experiment.
The mean age was 22.5 years (SD: 1.8 years).
They had experience wearing some HMDs but did not understand the details of the proposed method.
If, during or after the experiment, the average openness of one eye was found to be less than 0.5 or the average error of the control condition, described later, was found to be 1.0\Deg or greater, the data of that participant were not used.
As a result, gaze data were obtained for 18 participants.

%----------------------------------------------------------------------
% 実験手順
%----------------------------------------------------------------------
\subsubsection{Data Acquisition Procedure}
%----------------------------------------------------------------------
% 概要
%----------------------------------------------------------------------
% 提案手法は，VRシーン内で歩きながら自由に周囲を見回したときの固視クラスタのみから，較正パラメータを推定する．
% タスクの前に，実験手法の説明に加えて，被験者にシーンについてのクイズをすることも伝えた．

% 本実験では，被験者に，最初にクラシカルな較正をするためにマーカを注視させ，その後，二種類のVRシーンを歩いてもらった．
% 最後に，被験者にシーンに関する質問と，視力に関するアンケートを行った．
% 以下に，それぞれの詳細を述べる．
Gaze data were acquired while the participants were walking and freely looking around in VR environments.
Each participant first received instruction on the experiments and gave informed consent.
In addition, we told the participants that they would be given some simple quizzes about the objects in the VR environments after experiencing these environments.
Next, participants were asked to gaze at the markers for the traditional calibration and then walk in two different VR environments.
Finally, the participant was asked to complete short quizzes about the scenes and a questionnaire about their visual acuity.
These quizzes were dummy tasks.
We did not use their answer data.
% We recruited multiple participants for the experiment.
% Since the results of this experiment were likely to be affected by the user's intentions, such as the way of walking and the PoRs, 
The details of each are described below.

\subsubsection{Tasks}
%----------------------------------------------------------------------
% タスク
%----------------------------------------------------------------------
% 被験者が二種類のシーンを歩く順番を，被験者ごとに入れ替え，実験の順序依存性をなくした．
% 各参加者には，右上のコーナーから出発して白線上を歩き，反時計回り，時計回りをそれぞれ２回ずつ，計4回を合計時間が180秒になるよう歩いて回ってもらった．
% 実験前にこれらの実験方法等の説明に加えて，VRシーンの体験後にシーン内のオブジェクトに関する簡単なクイズをすると伝え，実際に体験後には視力等の情報の記入に加えた．
% このクイズは，ぼんやりと宙を眺めたりしないようにするための，ダミータスクであり，クイズの解答データは使用しなかった．
To counterbalance the order effect, the participants were divided into two groups that experienced the two environments in reverse order.
%The order in which participants walked in the two different scenes was swapped for each participant to eliminate the order dependence of the scenes in the experiment.
For each environment, starting from the upper right corner in Fig.~\ref{fig:vr-scenes}, each participant was asked to walk along the white outline in the VR environments, counterclockwise and clockwise, twice each, for a total of four consecutive laps.
The pace of the walk was sometimes instructed to the participant so that the total duration would be around 180~s.

%----------------------------------------------------------------------
% 固視検出の閾値
%----------------------------------------------------------------------
\subsubsection{Thresholds for Fixation Detection}
% 各固視検出アルゴリズムで用いた閾値の値について述べる．
% I-VDTでは，グリッドサーチによってI-VDTのベストな閾値を調査した従来研究に従い，速度閾値を80\Deg/s，分散閾値を0.7\Degに設定した．
% I-VT，I-DTでも同様に，それぞれの閾値を80\Deg/s，0.7\Degに設定した．
% 固視と判断する最小時間は，一般には100~200 msが設定される\parencite{Salvucci_ETRA2000}ため，3アルゴリズムともに，$T_{\rm th}$を160~msで統一した．
The threshold values for each fixation detection algorithm were as follows.
In I-VDT, the velocity threshold was set to 80~\Deg/s (${\phi_{\rm {th}}=0.16}$\Deg, ${S_{\rm{r}}=50}$~Hz) and the dispersion threshold to 0.7\Deg~(${D_{\rm{th}}=1.5\times10^{-4}}$), following previous studies~\parencite{Startsev_BRM2019} that investigated the best threshold of I-VDT by grid search.
Similarly, for I-VT and I-DT, the thresholds were set to 80\Deg/s and 0.7\Deg, respectively.
Because the minimum fixation time is generally set to a value from 100 to 200~ms~\parencite{Salvucci_ETRA2000}, we set $T_{\rm{th}}$ to 160~ms for all three algorithms.

\subsubsection{Control Condition and Accuracy Evaluation}
\label{sec:Control Condition and Accuracy Evaluation}
% HMD内蔵の視線計測器では，ユーザごとの光軸を得ることができず，HMDのシステム較正後の視線しか取得できない．
% また，システム較正の精度は，1\Degを超える場合もあり，十分ではない．
% 本実験では，真値の代わりに，古典的な較正方法によって得られた視線を使用した．
% 以後，この視線をベースラインと呼ぶことにする．
% この古典的な較正は，デバイスに依存せずにベースラインを生成するためのものであり，式(\ref{eq:calib_func})による較正とは異なる．
We regarded the parameters estimated by traditional marker-based calibration with a 2D regression model as the control condition. 
The gaze direction obtained after a built-in calibration of the Vive Pro Eye for each participant was not treated as the ground truth because of its insufficient accuracy, sometimes far exceeding 1\Deg.
With the Vive Pro Eye, built-in calibration should be performed at least once to acquire gaze data.
Therefore, the built-in calibration was performed only once for one of the authors, who is not included in the participants, and we maintained this state while acquiring the gaze data of all the participants.
% We regard the parameters estimated by a traditional marker-based calibration as the ground truth.
% Note that this calibration is different from our self-calibration based on Equation~(\ref{eq:calib_func}).

% ベースラインは，25個（=水平5$\times$垂直5）のクロスマーカを注視することによって得た視線データから推定した．
% マーカは，前方1mのシーンカメラ座標系に固定された平面上にランダムな順で表示された．
The parameters of the control condition was estimated using 5~$\times$~5~black cross markers.
The markers were displayed in pseudo random order on a gray-colored plane fixed to the scene camera coordinate system and 1~m in front of the user, with a FOV angle of 20\Deg.
% 各マーカを注視しながら，HTC Viveのコントローラのトリガを引いた後3秒間被験者に注視してもらい，取得した3秒のうち中央の1秒分を用いた．
% これにより25セットの視線集合を得た．
% 視線データ25セットのうち16個を推定用に，9セットを評価用に用いた．
We asked each participant to gaze at each marker for 3~s.
This resulted in 25~sets of gaze data for each participant.
We used 16 of the these sets for the estimation of the control condition parameters; the other 9 sets were used for confirming the accuracy of 
the control condition and the proposed methods.
%the ground parameters $\hat{\bm \theta}$.
% マーカの表示方法や視軸の推定方法の詳細は付録に記載する．
Details on the methods for displaying the markers and estimating the control condition are provided in the supplemental material.

The Vive Pro Eye does not provide the optical axis, but it provides the gaze direction corresponding to the visual axis estimated by the built-in calibration process.
Therefore, we generated arbitrary parameters from the control condition, i.e., the estimated visual axis.
The optical axis $\bm{g}_{\rm{opt}}$ can be calculated from the visual axis $\bm g_{\rm{vis}}$ by the inverse transformation of Equation~(\ref{eq:opt_to_vis}), as shown in the following equation:
\begin{eqnarray}
\label{eq:vis_to_opt}
\bm{g}_{\rm{opt}} &\propto& f^{-1}(\bm g_{\rm{vis}}\,;\,{\bm \theta}_{\rm{offset}}).
\end{eqnarray}
We determined the gaze direction $\bm{g}$ calibrated by arbitrary parameters ${\bm \theta}$ by the following redefined equation:
\begin{eqnarray}
\label{eq:vis_to_calib}
\bm{g} &\propto& f^{-1}(\bm g_{\rm{vis}}\,;\,{\bm \theta}).
\end{eqnarray}
In this case, the control condition parameters can be represented as $\bm \theta=[0,0]$.
% In our implementation, the initial parameters for the fixation detection and any parameters in the optimization process were generated from the visual axis because our eye tracker does not provide the optical axis directly.
% This is a simulation-like method, which differs slightly from the actual usage, where parameters in the optimization are generated by rotating the optical axis obtained by an eye tracker.
% However, for the purpose of confirming the accuracy, it is equivalent to the actual self-calibration.
% また，視線較正精度の評価は，9セットの視線をEquation~(\ref{eq:calib_func})で較正し，マーカと較正後の視線との誤差を度単位で評価した．
% To evaluate the accuracy of the results, the nine sets of gaze directions were corrected with Equation~(\ref{eq:calib_func}), and the average error between the marker and the calibrated gaze was computed in degrees.
For the evaluation of the accuracy of both the proposed method and the control condition, the estimated parameters were applied to the same nine sets, and the average absolute error from the markers was evaluated in degrees.

%----------------------------------------------------------------------
% 精度評価
%----------------------------------------------------------------------
\subsection{Accuracy Evaluation}
\label{sec:Accuracy Evaluation}
% 提案手法の自動較正性能を確認するため，提案手法の精度を光軸の精度と比較した．
% 提案手法は自動較正手法であるため，固視検出と最適化の初期パラメータに光軸を用いる必要がある．
% この条件で推定された較正パラメータの精度が，提案手法の精度そのものになる．
% この初期条件に加えて，固視検出の初期パラメータに視軸を使用し，最適化の初期値パラメータに光軸を使用したときに推定された較正パラメータを評価する．
% これは高精度の固視検出結果を用いた場合の提案手法を評価することを意味する．
% 従って，光軸または視軸に対応する較正パラメータで固視検出し，提案手法の精度を評価した．
\subsubsection{Purpose}
In this study, accuracy evaluation has two purposes.
The first is to confirm the feasibility of our self-calibration method. 
One straightforward criterion for feasibility is that the accuracy of the visual axis estimated by the proposed method is higher than that of the optical axis as the initial parameters obtained by the 3D eye model.
The second is to determine the impact of the choice of fixation detection algorithm and the use of uncalibrated gaze data for fixation detection on accuracy.

% \textcolor{red}{
% It is necessary to use the optical axis as initial parameters for the fixation detection and optimization of the proposed method, because the proposed method is a self-calibration method.
% The accuracy of the calibration parameters estimated under this initial condition is the accuracy of the proposed method itself.
% In addition to this initial condition, we evaluate the calibration parameters estimated when the visual axis is used as the initial parameter for the fixation detection and the optical axis is used as the initial parameter for the optimization.
% This means that the proposed method using highly accurate fixations is evaluated.
% Therefore, we evaluated the accuracy of the proposed method by detecting fixations with calibration parameters corresponding to the optical axis and visual axis.
% }

\subsubsection{Method}

\begin{table}[bp]
    \centering
    \caption{Combination of Fixation Detection Algorithms and Initial Parameters}
    \scriptsize
    %\begin{tabular}{p{\len}cccccc}
    \begin{tabular}{lcccccc}
    \toprule
    %\begin{tabular}{l}
    Methods
    %\end{tabular} 
    & I-VT (opt) & I-VT (vis) & I-DT (opt) & I-DT (vis) & {\bf I-VDT (opt)} & I-VDT (vis)\\
    \hline
    \grayrow
    %\begin{tabular}{l}
    Fixation detection algorithms
    %\end{tabular} 
    & I-VT & I-VT & I-DT & I-DT & {\bf I-VDT} & I-VDT\\
    %\hline
    %\begin{tabular}{l}
    Initial parameters for fixation detection
    %\end{tabular} 
    & optical axis & visual axis & optical axis & visual axis & {\bf optical axis} & visual axis\\
    %\hline
    \grayrow
    %\begin{tabular}{l}
    Initial parameters for optimization
    %\end{tabular} 
    & optical axis & optical axis & optical axis & optical axis & {\bf optical axis} & optical axis\\
    \bottomrule
    \end{tabular}
    \label{tb:evaluation-method}
\end{table}

We evaluated the accuracy of the variations of the proposed method and compared the accuracy with that of the optical axis and the visual axis (control condition). 
% それぞれのシーンで検出された固視を用いた場合（Office, Supermarket）と，二種類のシーンで検出された固視を合算した固視を用いる場合（Office+Supermarket）の三条件で各手法を評価した．
% また，光軸と視軸の平均精度も確認した．
As variations of the proposed method, we prepared six methods that are all possible combinations of three fixation detection algorithms (I-VT, 3D I-DT, and 3D I-VDT) and two initial values (opt: optical axis, vis: visual axis) given to each fixation detection algorithm.
%Therefore, we confirmed the average absolute error for 18 participants of I-VT~(opt), I-VT~(vis), I-DT~(opt), I-DT~(vis), I-VDT~(opt), and I-VDT~(vis), according to the fixation detection algorithms and the calibration parameters.
%Note that "opt" and "vis" refer to the use of the optical and visual axes for the fixation detection, respectively.
For all the methods, the initial parameter for optimization is the optical axis.
Table~\ref{tb:evaluation-method} summarizes all the methods.
% 光軸または視軸に対応する較正パラメータで固視検出し，提案手法の精度を評価した．
% 光軸には，平均的な視軸とのオフセットを用いた．
% Abassら~\parencite{Abass_CER2018}は，ブラジル人，中国人，イタリア人計1020人の左右の眼それぞれの光軸と視軸のオフセット角を調査した．
% 本実験では，先行研究の全被験者の左右の平均オフセットをさらに平均化して，単眼のオフセットを$\{\alpha,\beta\}=\{-1.02,-3.30\}$と近似した．
For the optical axis, average offsets reported in a previous study were used.
The literature~\parencite{Abass_CER2018} reported the average offsets of both eyes in a total of 1020 Brazilian, Chinese, and Italian participants.
In our experiment, the offset of the cyclopean eye was approximated $\bm{\theta}_{\rm{offset}}=[1.02,3.30]$ by averaging each component of both eyes' offsets.

We applied each method to three sets of data containing fixations detected in Office, Supermarket, and their combination (named Office+Supermarket).
For reference, the average absolute errors of the optical and visual axes (control condition) of all the participants were also computed.
In total, the average absolute error of 20 conditions (3 detection algorithms $\times$ 2 initial values $\times$ 3 environments + 2 references) was evaluated.

% シーン条件ごとに，提案手法と光軸と視軸を含む8手法の結果を多重比較検定した．
% 事前のShapiro-Wilk法\parencite{Shapiro_Wilk_1965}によると，それぞれの絶対誤差は正規分布していなかったので，差の検定にはSteel--Dwass法を用いた\parencite{Steel_1960,Dwass_1960}．
% それは，一般の確率分布を仮定したノンパラメトリック検定である．
For each environment condition, we performed a multiple test on the seven results of all the participants: the results of the six proposed methods and the optical axis.
Because the prior Shapiro-Wilk test~\parencite{Shapiro_Wilk_1965} confirmed that the absolute errors of each method were not normally distributed, we applied Steel--Dwass's method~\parencite{Steel_1960,Dwass_1960}, which is a non-parametric multiple comparison test that assumes arbitrary probability distributions.

%----------------------------------------------------------------------
% 精度評価（実験結果）
%---------------------------------------------------------------------
\subsubsection{Results}
\begin{figure}[tp]
\begin{center}
\includegraphics[width=0.9\columnwidth]{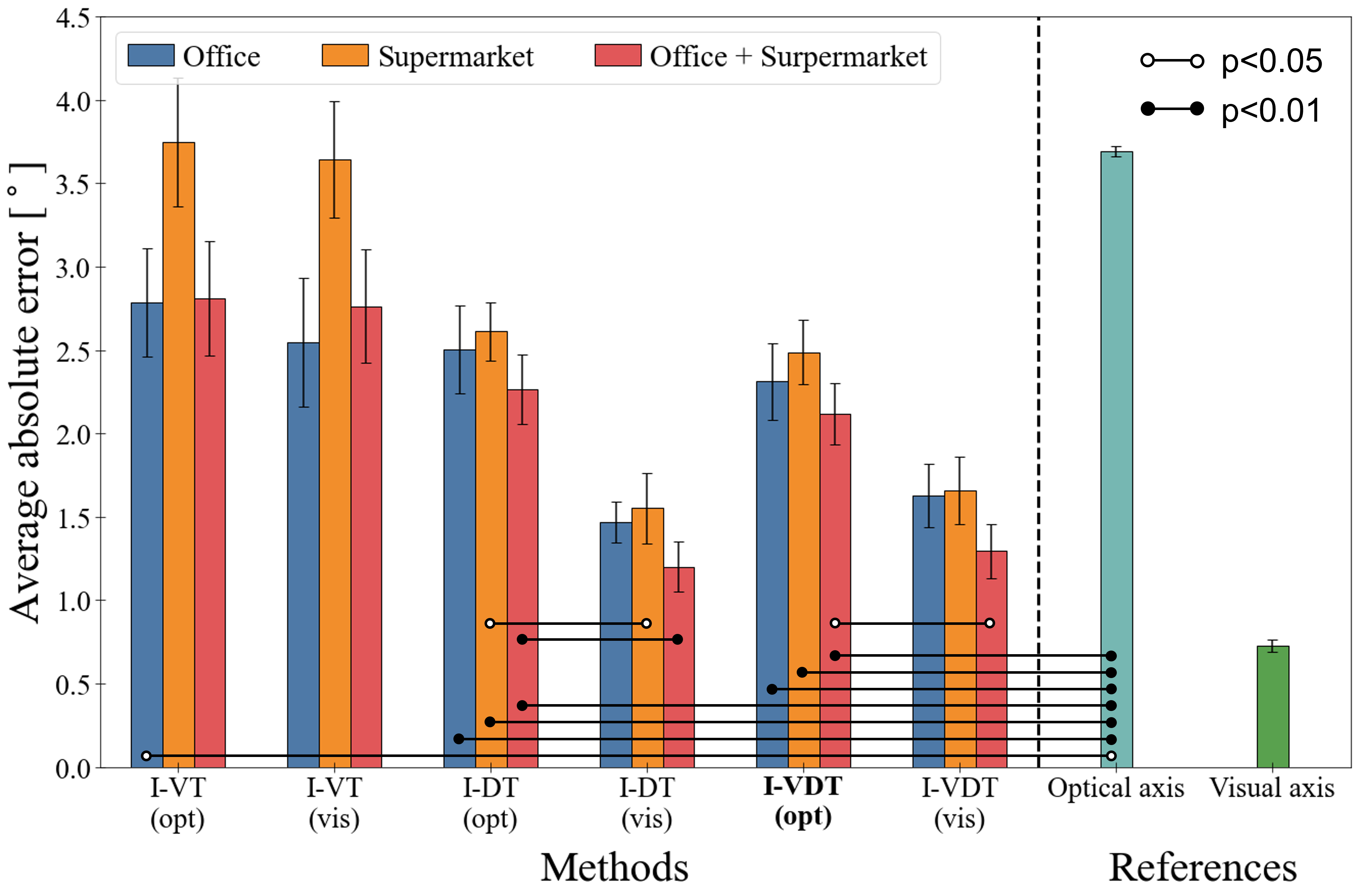}
\end{center}
\vspace{-4mm}
\caption{Absolute errors of the proposed method. ``opt" indicates that the initial calibration parameters corresponding to the optical axis were used in the fixation detection. Similarly, ``vis" refers to fixation detection using the parameters of the visual axis.
``Visual axis'' represents the accuracy of the control condition with 16 of the 25 markers.
``Optical axis'' shows the accuracy of the optical axis generated by adding the average offset to the control condition.
All the errors were evaluated with the same gaze data of the other nine markers.
The error bars indicate standard errors for all participants.
}
\label{fig:abs-result}
\end{figure}

% 図\ref{fig:abs-result}は，18人の被験者に対する提案手法の平均精度を示す．
% 図中の，エラーバーは標準誤差を示す．
% I-DT (opt) とI-VDT (opt) の絶対誤差は，どの条件においても，Optical axisの絶対誤差がよりも統計的有意 (p$<$0.05) に低かった．
% Office+SupermarketにおけるI-VDT (opt) が，光軸で固視検出した手法のうち，最も高精度であり，2.12\Degだった．
% この精度は，任意のシーンに適用可能な従来の視線自動較正手法と比べても，最も高い．
% 一方，I-VT (opt) は，Officeを除き，Optical axisに対して，有意差は確認されなかった．
Fig.~\ref{fig:abs-result} shows the average absolute errors under all the conditions.
First, we focus on the case ``opt" and present the results of using the optical axis for fixation detection.
The absolute errors of I-DT (opt) and I-VDT (opt) were significantly ($p<0.01$) lower than those of the optical axis in all environmental conditions.
I-VDT (opt) in the Office + Supermarket had the highest accuracy with 2.12\Deg among the ``opt" methods.
Meanwhile, I-VT (opt) was not significantly different from the optical axis, except in the Office.

% I-DT (vis) とI-VDT (vis) は，どの条件においても，I-DT (opt) とI-VDT (opt) それぞれに対して精度向上が確認された．
% Office+Supermarketにおいて，I-DT (vis) とI-VDT (vis) は，すべての光軸を用いた手法に対して，有意差 (p$<$0.05) が確認された．
% Office+SupermarketにおけるI-DT (vis) は，視軸で固視検出した手法のうち，最も高精度であり，1.20\Degだった．
% Office+SupermarketにおけるI-DT (vis) が，提案手法のうち，最も高精度であり，1.20\Degだった．
% この結果は，フラットモニタで確認された従来の視線自動較正手法と比較しても，同等以上の精度である．
Next, we focus on the comparison of ``opt" with ``vis", i.e., the difference between the initial parameters.
I-DT (vis) and I-VDT (vis) showed improved accuracy of I-DT (opt) and I-VDT (opt), respectively, under all conditions.
In Office + Supermarket, I-DT (vis) and I-VDT (vis) were significantly better ($p < 0.05$) for all ``opt" methods.
I-DT (vis) in the Office + Supermarket was the most accurate with 1.20\Deg.

\subsubsection{Discussion}
\label{sec:exp1-discussion}
% This accuracy was also the highest compared to all the conventional self-calibration methods~\parencite{Liu_IEEE_Access2020,Shi_Elsevier_CAG2020,Murauer_iWOAR2018} for arbitrary scenes.
% This result is comparable to or more accurate than the conventional scene model-based self-calibration methods~\parencite{Nagamatsu_ETRA2010a,Model_ETRA2010,Model_IEEE_TBE2010,Maio_FG2011,Wang_Elsevier_PR2018} for a flat monitor.
%----------------------------------------------------------------------
% 精度評価（考察）
%----------------------------------------------------------------------
% 提案手法は，scene model based self-calibrationであるため，予想通り，任意シーンに適用可能な従来手法よりも精度が高かった．
% 提案手法は，固視検出アルゴリズムによって結果が異なる．
% I-DT，I-VDTの絶対誤差に比べ，I-VTの絶対誤差が悪いことが確認された．
% 従来研究は~\parencite{Salvucci_ETRA2000}，I-VTの固視検出精度はI-DTに比べて低いことを報告した．
% 従って，提案手法は固視検出精度に依存すると考えられる．
Some expected results were obtained.
First, the proposed method with I-VDT (opt) was more accurate than the optical axis.
This means that the proposed method is applicable to gaze estimation based on the 3D eye model.
None of the self-calibration methods applicable to arbitrary scenes shown in \hyperref[sec:relatedwork]{\it Related Work} have been statistically shown to be more accurate in estimating calibration parameters than the optical axis.
Thus, the proposed method is the first whose feasibility has been confirmed.

Second, the average absolute errors of I-VDT (opt) in any scenes were also smaller than those of any conventional self-calibration methods for arbitrary scenes~\parencite{Liu_IEEE_Access2020,Shi_Elsevier_CAG2020,Murauer_iWOAR2018}, i.e., about 3\Deg at best.
%これらの手法はsaliency basedとsmooth pursuit basedであるのに対して，提案法がモデルベースであることを考えると，
This result is consistent with the fact that scene model-based methods~\parencite{Nagamatsu_ETRA2010a,Model_ETRA2010,Model_IEEE_TBE2010,Maio_FG2011,Miki_ETRA2016,Wang_Elsevier_CVIU2005} tend to be more accurate than saliency-based methods~\parencite{Sugano_IEEE_PAMI2013,Chen_IEEE_TIP2015} or smooth pursuit-based methods~\parencite{Murauer_iWOAR2018}, as shown in Table~\ref{table:auto-calib}.
In previous studies, participants were asked to search for a specific object~\parencite{Shi_Elsevier_CAG2020} or to see a scene in with one salient object~\parencite{Tamura_ETRA2020,Murauer_iWOAR2018} in their experiments.
In our experiments, we have not employed the tasks, which could potentially increase the accuracy of the calibration.
However, the statistical superiority of the proposed method cannot be determined without the unification of experimental conditions.
% For example, our method adopts a cyclopean eye, whereas in previous studies, the parameters were independently estimated for both eyes.
% In terms of the stability of parameter estimation, this clearly works in favor of the proposed method.
% Comparative experiments under standardized conditions have not been performed in previous studies. This is our future work.

% Third, the absolute error of I-VT was larger than that of I-DT and I-VDT because the accuracy of I-VT was reported to be lower than that of I-DT.
% which suggests the effectiveness of using a scene model not only for the optimization but also for the fixation detection in the proposed method.

% 固視検出に視軸を用いた場合，提案手法の精度が1.2\Degまで大きく改善された．
% 視軸における絶対誤差を考慮すると．かなり高精度である．
% また，フラットモニタで確認された従来の視線自動較正手法にも匹敵する精度である．
% この結果は，固視検出結果の精度が検出に必要な較正パラメータに依存することを示唆する．
% 通常，固視検出は，較正後の視線方向を用いるのが一般的であり，これまでの研究で，較正精度が固視検出結果に与える影響は議論されていない．
% 節\ref{sec:Dependence on Fixation Detection}で，網羅的に，検出に必要な較正パラメータと提案手法の精度について調査する．
% 較正パラメータに依存しない固視検出手法が開発された場合の提案手法が更に向上することを示している．
% 本実験により，提案手法が1\Deg程度まで向上する可能性を示した．
% 提案手法は，固視検出とパラメータ最適化の二部構成である．
% 固視検出の精度が十分高い場合，提案手法の精度が高い．
% それゆえ，再投影誤差を用いた評価関数の有用性は十分に示された．
% 今後は，固視検出の開発アルゴリズムの開発に取り組む必要がある．
In contrast, unexpected results were also obtained.
First, the absolute error of ``vis" was smaller than that of ``opt" in I-DT and I-VDT in Office + Supermarket environments. 
In particular, the accuracy of ``vis" in I-DT was surprisingly improved to 1.2\Deg, which is fairly close to the accuracy of the visual axis (green bar in Fig.~\ref{fig:abs-result}).
Note that we used the visual axis parameters only for fixation detection and not for optimization, and the effect of the different parameters on the fixation detection is only a slight increase or decrease in the number of frames judged to be fixations.
This result indicates the little to no cause for inaccuracy in the optimization, and that mainly fixation detection has room for improvement.
In general, calibrated gaze data must be used for fixation detection. 
To our knowledge, previous studies have not discussed the effect of calibration accuracy on the results of fixation detection.
Therefore, in \hyperref[sec:Dependence on Fixation Detection]{\it Dependence on Initial Parameters}, we comprehensively investigate the relationship between the calibration parameters for fixation detection and the accuracy of the proposed method.
The above trend did not exist in the case with I-VT.
We expected this to some extent because changes in the offset of the gaze direction have little effect on the angular velocity of the gaze direction used by I-VT as a criterion.
In the future, development of a fixation detection method that is less dependent on initial parameters and the optimization of parameters for the entire method including fixation detection are necessary.

% どのようなVRシーンであっても局所的もしくは区分的には平面のポリゴンにより表現されているため，PoRsが分布する面の傾きが平均化されることが重要である．
% 偏った面の方向ばかり注視した場合，提案手法の精度は悪化する．
% Office + Supermarketの結果の方が高精度であったのは，PoRsが分布する面の傾きが平均化されたためと考えられる．
% また，単純に固視数が増えた影響も考えられる．
% 以上のことから，最適化に使用する固視を，PoRsが分布する面の傾きが平均化するように選択することで，推定精度の向上させることが可能になる．
Second, it was found that the accuracy of Supermarket tended to be lower and the that of Office + Supermarket tended to be higher than that of the other scenes.
Although statistical significance was not confirmed, the Office scene was more accurate than the Supermarket scene under any condition, and the same trend was observed in \hyperref[sec:Convergence Performance]{\it Convergence Performance} and \hyperref[sec:Dependence on Fixation Detection]{\it Dependence on Initial Parameters}.
The cause of this scene dependence could not be clarified in our experiments.
A possible cause is the variation of surface normals.
The sensitivity of the reprojection error depends on the scene distance and the surface inclination. 
The proposed method was expected to average out the bias owing to such sensitivity using multiple fixations because a realistic scene had surfaces with various normal directions.
The abovementioned scene dependence is consistent with the results if interpreted as follows: Supermarket had a biased distribution of surface normals and was less accurate than Office; Office + Supermarket had the best accuracy owing to the enhanced averaging by the increased variations of surface normals.
Another possible cause could simply be an increase in the number of fixations.
This is contradicted by the experiments in the next section.

%----------------------------------------------------------------------
% 距離による収束性能
%----------------------------------------------------------------------
\subsection{Convergence Performance by Walking Distance}
\label{sec:Convergence Performance}
\subsubsection{Purpose}
% 参加者の歩行距離に応じた提案手法の精度を確認した．
This experiment aimed to confirm the relationship between the accuracy of the estimated parameters and the cumulative translational distance of users.
This relationship indicates how far the user must move to complete the calibration.
The convergence speed of self-calibration is often based on time or the number of frames~\parencite{Sugano_IEEE_PAMI2013}. 
Because the proposed method requires fixations to occur along with the translational movement of the user's head, we confirmed the accuracy and the number of fixations with respect to the translational movement distance.

\subsubsection{Method}
% 計測開始時から特定の歩行距離の間に検出された固視を使用して提案手法を適用し，18人の平均精度を得た．
% 二つのVRシーンそれぞれにおいて，3~mから34~mの範囲を1~m刻みで，平均精度を取得する．
% 移動距離の最小値は3~mで，これより短い移動距離の場合，固視数が0の被験者が存在する．
% 移動距離の最大値は34~mで，これは，18人のうち，合算の移動距離が最も短い被験者の移動距離である．
% 固視検出には，節\ref{sec:Accuracy Evaluation}で定義した光軸を使用し，I-VT，I-DT，I-VDTの3つのアルゴリズムを用いた．
For the two VR scenes, we obtained the average absolute errors in distance intervals from 3 to 34~m in 1~m increments.
A minimum distance of 3~m was considered because in some cases, fixation is not observed at distances less than 3~m.
A maximum distance of 34~m was set as the shortest cumulative distance among the 18 participants.
Three algorithms, I-VT, I-DT, and I-VDT, were used for fixation detection, with the optical axis as defined in \hyperref[sec:Accuracy Evaluation]{\it Accuracy Evaluation}.
% 各手法による収束性能は，移動距離が3~m，18~m，34~mであるときの3条件をそれぞれ比較した．
% 各精度の集合は，事前のShapiro-Wilk法で正規分布していなかったので，差の検定にはSteel--Dwass法を用いた．
%The convergence performance of each method was compared at 3~m, 18~m, and 34~m.
The average absolute errors at 3, 18, and 34~m were compared for the three fixation detection algorithm.
The Steel--Dwass method was used to test for differences.

%----------------------------------------------------------------------
% 距離による収束性能（実験結果）
%----------------------------------------------------------------------
\subsubsection{Results}
\begin{figure}[t]
    \centering
    \begin{minipage}{0.49\columnwidth}
    \centering
        \includegraphics[width=\textwidth]{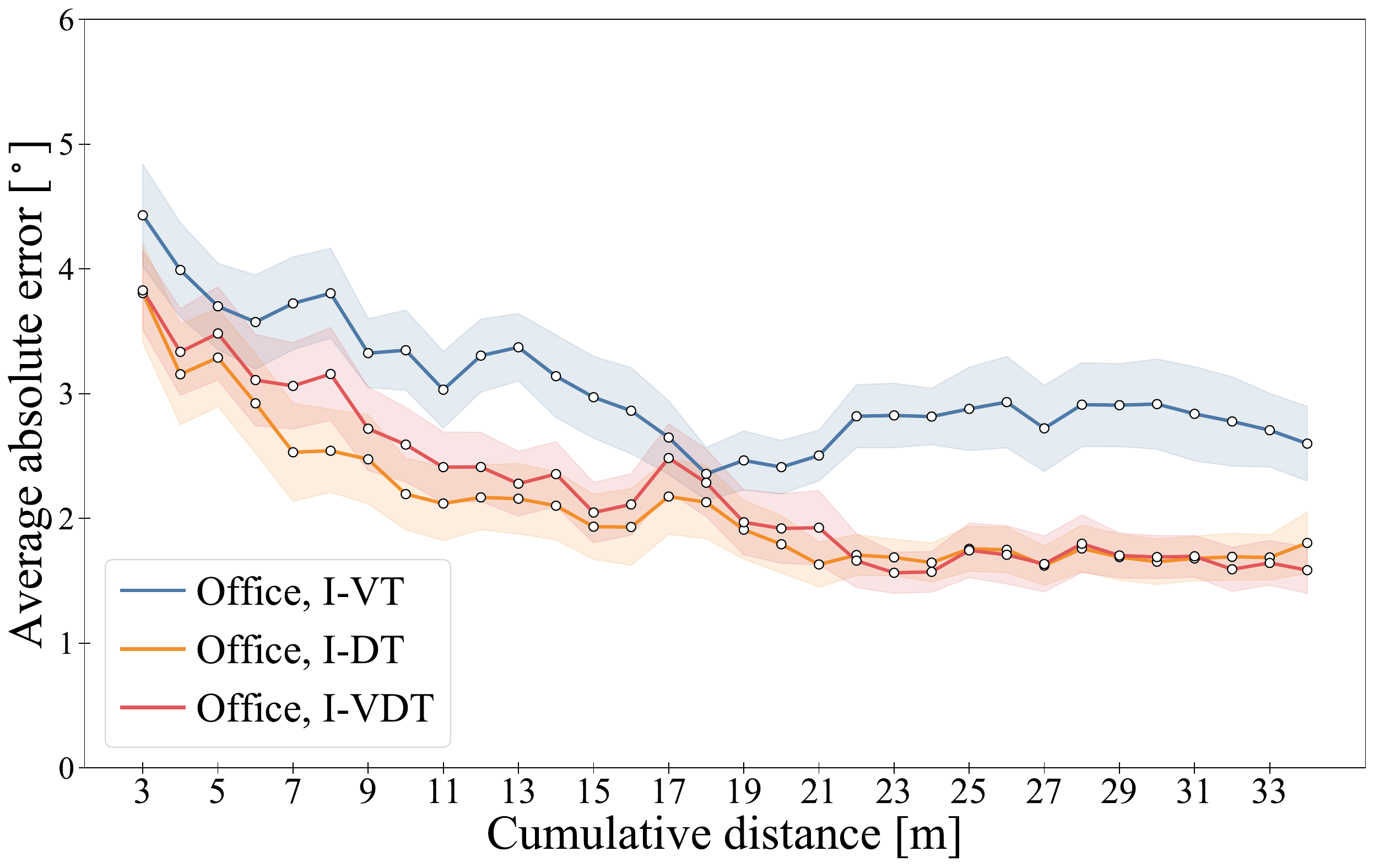}\\
        (a)\\
        \vspace{3mm}
        \includegraphics[width=\textwidth]{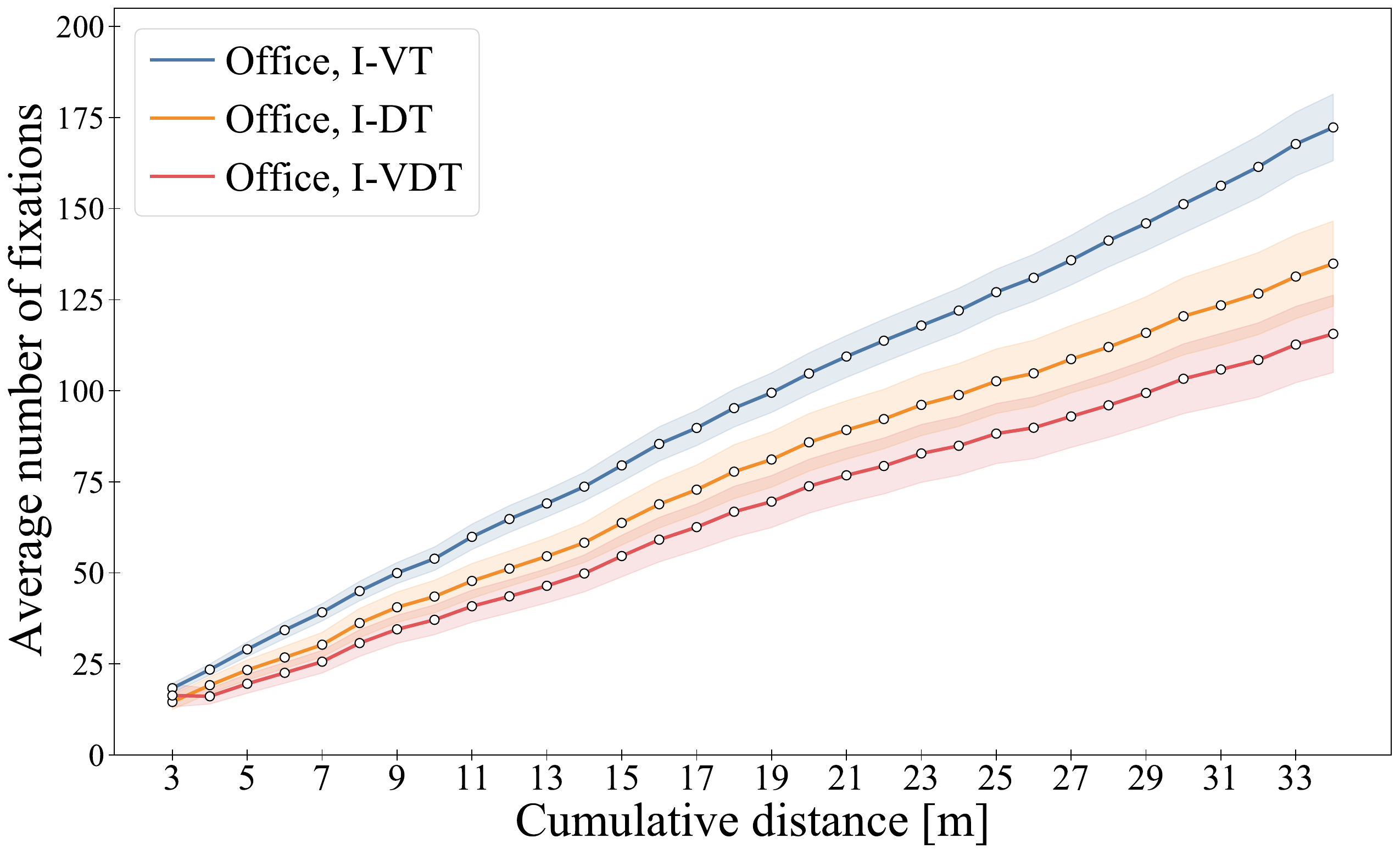}\\
        (c)
    \end{minipage}
    \begin{minipage}{0.49\columnwidth}
    \centering
        \includegraphics[width=\textwidth]{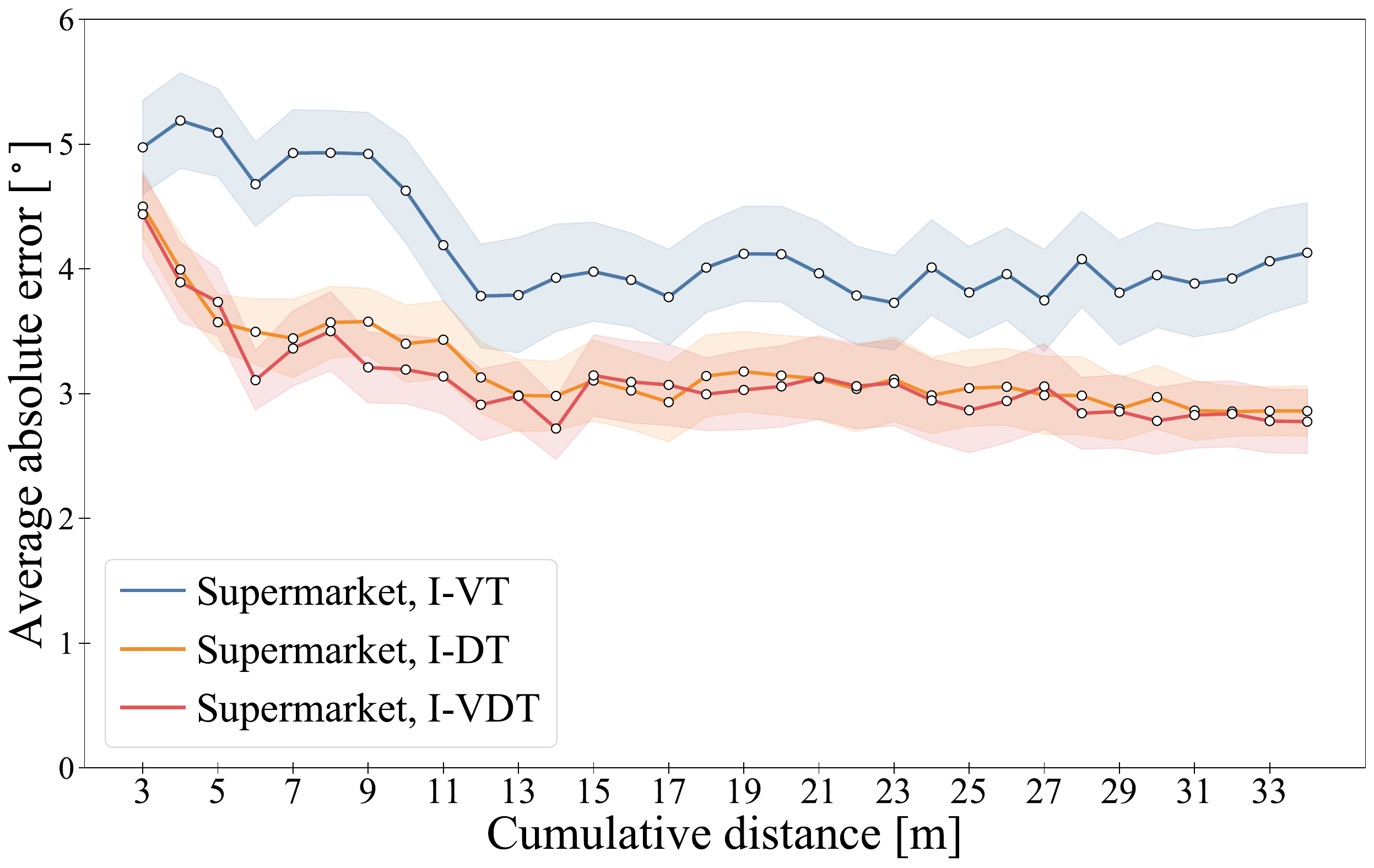}\\
        (b)\\
        \vspace{3mm}
        \includegraphics[width=\textwidth]{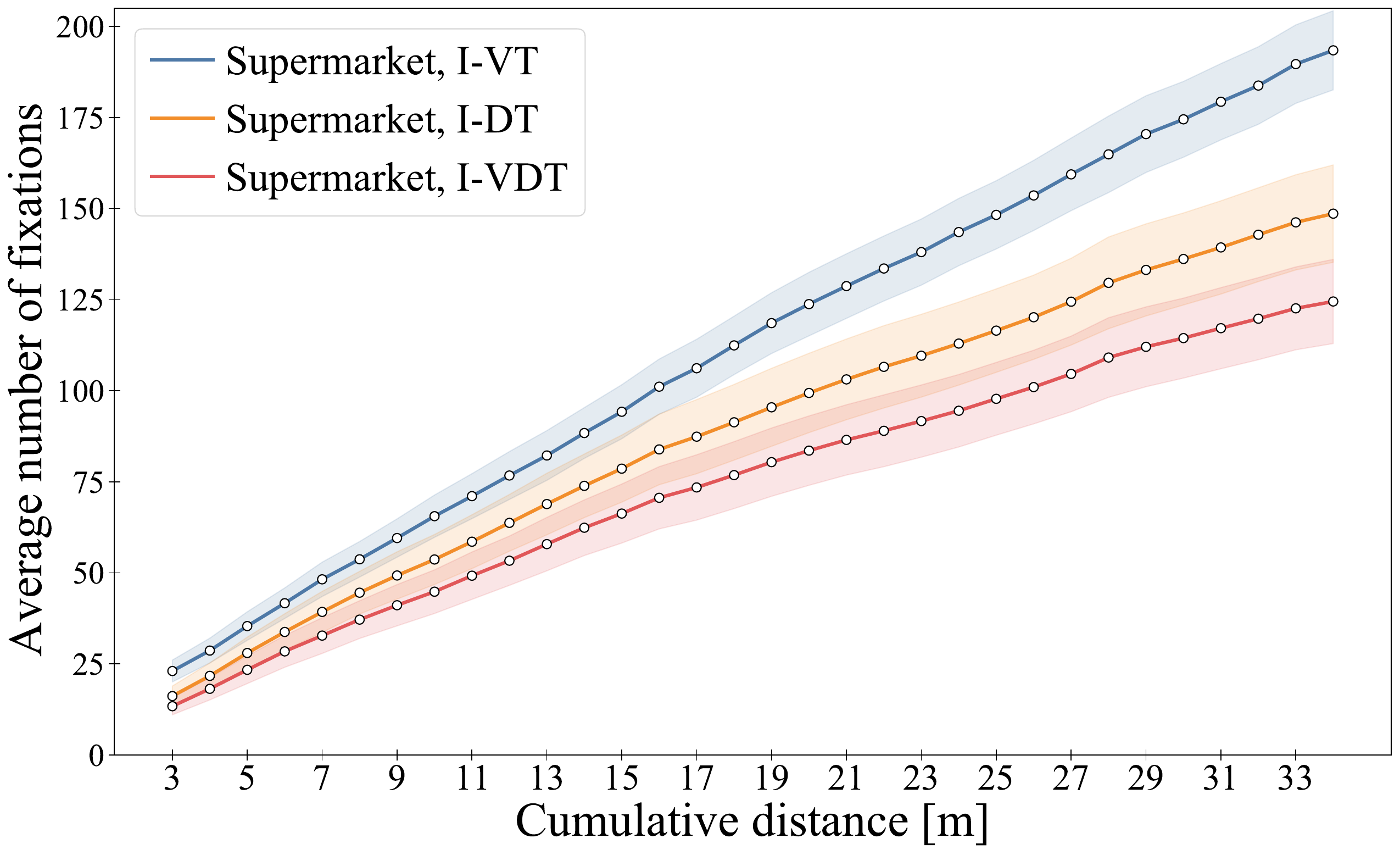}\\
        (d)
    \end{minipage}
    \caption{Absolute error and number of fixations.
    The cumulative distance is the sum of the translational distances of the scene cameras between adjacent frames.
    }
    \label{fig:dis-result}
\end{figure}

% 図~\ref{fig:dis-result}(a)，図~\ref{fig:dis-result}(b)は，Office，Supermarketシーンにおける精度である．
% 図~\ref{fig:dis-result}(c)，図~\ref{fig:dis-result}(d)は，Office，Supermarketシーンにおける固視数である．
% 図中のエラーバンドは，標準誤差を表す．
Fig.~\ref{fig:dis-result} (a) and (b) show the average absolute error versus cumulative distance in Office and Supermarket, respectively.
Fig.~\ref{fig:dis-result} (c) and (d) show the average number of fixations versus cumulative distance in Office and Supermarket, respectively.
The error bands with light colors along the lines indicate standard errors.
% SupermarketでのI-VTの結果を除いて，移動距離18~mと34~mは，移動距離3~mよりも高い精度を示し，その差は統計的に有意 (p$<$0.05) であった．
% 移動距離18~mと34~mには，有意差 (p$<$0.05) は確認されなかった．
% SupermarketでのI-VTの結果では，移動距離が3~m，18~m，34~mともに有意差 (p$<$0.05) は確認されなかった．
% 固視数は，I-VDT，I-DT，I-VTの順に少なかった．
% 二つのシーンでの平均したI-VDTのパラメータ最適化時間は，3~m，18~m，34~mでそれぞれ，XX~s，XX~s，XX~sであった．
% 固視数が多くなるほど最適化時間が長くなる．
Except for I-VT in Supermarket, the average errors at 18 and 34~m showed significantly higher accuracy than that at 3~m ($p < 0.05$).
No significant difference was found between the average errors at 18 and 34~m.
I-VT in Supermarket showed no significant difference between any two of the three distances.

The number of fixations decreased in the order of I-VT, I-DT, and I-VDT at any distance.
Significant differences between I-VT and I-VDT were confirmed at 3~($p < 0.05$), 18~($ p< 0.05$), and 34~m ($p < 0.01$).
For reference, the average computation times of the two scenes were 10, 28, and 56~s at 3, 18, and 34~m, respectively, in the case of I-VDT.
% It was also found that the more the number of fixations, the longer the optimization time.

\subsubsection{Discussion}
%----------------------------------------------------------------------
% 距離による収束性能（考察）
%----------------------------------------------------------------------
% 提案手法の精度は，移動距離が18~mの時点で，歩行開始時に対し向上し，34~m時点の精度と同程度になる．
% このことから，歩行距離が18~m程度で，自動較正が有効であると言える．
The accuracy of the proposed method in I-DT and I-VDT improved significantly at the cumulative distance of 18~m compared with that at the start of walking, and this value is not significantly different from the accuracy at 34~m.
% Therefore, our self-calibration is effective when the cumulative distance is about 18~m.
Therefore, the parameters almost converge after a walking distance of about 20~m.
The speed to reach 34~m in 180~s is approximately 0.7~km/h, 
which is considerably slower than the typical walking speed of 4~km/h.
This is because we could only prepare a small environment, and we instructed them to walk at such a speed.
Considering that the proposed method utilizes the translation of the head position, even faster convergence can possibly be achieved if the user walks at normal speed.

% cumulative distanceの増加に伴い，全体的には誤差は減少した．
% しかしながら，誤差が増大した箇所も確認された．
% 提案手法の最適化で用いる固視に，誤った固視が含まれる場合，パラメータの推定精度は悪化する．
% また，提案手法は，頭部移動が前提のため，立ち止まっているときに検出された固視が多く含まれる場合，推定精度が悪化する．
% 立ち止まって多くの固視が検出された場合，固視数が極端に上昇するcumulative distanceが存在する．
% 図~\ref{fig:dis-result}(c)，図~\ref{fig:dis-result}(d)から，本実験では，極端に上昇するcumulative distanceはないため，そのような固視は多くは含まれていない．
% 偏った面の方向ばかり注視した場合，提案手法の精度は悪化する．
Although the trend was for the error to decrease with distance, at some distances, the error locally increased in both scenes even with I-DT and I-VDT.
We present four possible hypotheses regarding the cause of this increase: 
(i) the head position was almost stationary, 
(ii) few fixations were detected, 
(iii) the dispersion of the detected fixations was large, and 
(iv) the inclination of the surface where the PoRs of fixations are distributed was biased.
% 被験者の頭部移動なしに検出された固視は，パラメター推定に悪影響を与える．
% しかし，その場合は，累積距離が増えないためその距離の固視数が不連続に増加するはずである．
% Fig.~\ref{fig:dis-result}(c)と(b)からはそのような増加は見られないのでこの仮説はつじつまが合わない．
% 2番目の仮説についても同じである．
According to Hypothesis (i), fixations detected without head movement adversely affect parameter estimation.
However, in that case, the number of fixations at that distance should rapidly increase  because the cumulative distance does not increase.
As shown in Fig.~\ref{fig:dis-result} (c) and (b), no such increase is found; thus, Hypothesis (i) does not hold. 
In contrast, if no fixation is detected as the cumulative distance increases, the absolute error should remain constant.
Therefore, Hypothesis (ii) is also rejected because the same is true.
Hypothesis (iii) cannot be ruled out because the dispersion of the fixation depends on psychological states~\parencite{Hafed_Elsevier_VisRes2002}.
We asked each participant to pace themselves or reverse their walking direction, which may have led to a loss of concentration.
In our method, such psychological factors are not taken into account.
Regarding Hypothesis (iv), the possibility of a bias in the inclination of the surface at some distances cannot be ruled out.

% If fixations used in the optimization includes incorrect fixations, the accuracy of parameter estimation deteriorates.
% In addition, since the proposed method assumes head movement, the estimation accuracy deteriorates when many fixations detected while the participant is standing still are included.
% When many fixations are detected by standing still, there exists a cumulative distance at which the number of fixations extremely increases.
% From Fig.~\ref{fig:dis-result}(c) and (d), there is no extremely increasing cumulative distance in this experiment, and thus many such fixations are not included.
% The accuracy of the proposed method deteriorates when only the biased inclination of the surface is gazed at.

\subsection{Dependence on Initial Parameters}
\label{sec:Dependence on Fixation Detection}
\subsubsection{Purpose}
% 固視検出で使用する較正パラメータと提案手法の精度の依存性を解析をした．
% 2節でも述べたように，固視検出結果は，初期パラメータに依存する可能性がある．
% 本節では，固視検出時の初期パラメータが，提案手法の精度に与える影響について確認した．
This experiment was performed to analyze the dependence of the accuracy of the proposed method on the initial calibration parameters used for fixation detection.
As mentioned in \hyperref[sec:relatedwork]{\it Related Work}, in general, fixation detection results depend on the algorithm and the initial calibration parameters.
To the best of our knowledge, there is no study on fixation detection using uncalibrated parameters.
In this section, we confirm the effect of the initial parameters of fixation detection on the accuracy of the proposed method.

\subsubsection{Method}
% パラメータ空間を五つの領域に分割し，それぞれの領域をパラメータ範囲1～5とした．
% それぞれのパラメータ範囲で較正パラメータ$\{\alpha,\beta\}$を50点ずつランダムに生成し，各パラメータで固視検出した．
% 各被験者の較正精度を固視検出結果を用いて取得し，各パラメータ範囲の平均精度を計算した．
% つまり，各パラメータ範囲の精度は，900個の結果(=50パラメータ点$\times$18人)の平均である．
The parameter space was divided into five regions, and each region was referred to as Range~1--5.
We define Range~$l$ as ${l-1 \leq|\alpha|\leq l}$ or ${l-1 \leq|\beta|\leq l}$ ($l=1,2,\cdots,5$ in degrees).
% We defined ${0\leq|\alpha|\leq1}$ or ${0\leq|\beta|\leq1}$ as range1, 
% ${1<|\alpha|<2}$ or ${1<|\beta|\leq2}$ as range2,
% ${2<|\alpha|\leq3}$ or ${2<|\beta|\leq3}$ as range3,
% ${3<|\alpha|\leq4}$ or ${3<|\beta|\leq4}$ as range4,
% and ${4<|\alpha|\leq5}$ or ${4<|\beta|\leq5}$ as range5.
Fifty sets of random calibration parameters $[\alpha,\beta]$ were generated with the uniform distribution for each parameter range, and fixation detection was performed using each set of parameters.
We obtained the calibration accuracy for each participant using the fixation detection results and computed the average absolute error for each parameter range.
Consequently, we acquired 900~results ($=50$~parameters $\times$ 18~participants).
% 2シーン（オフィス，スーパー），3アルゴリズム（I-VT，I-DT，I-VDT）の計6条件の精度結果を得た．
% 前述の実験と同様に，事前のShapiro-Wilk法によると，各パラメータ範囲の精度は正規分布していなかったので，各条件ごとにパラメータ範囲の集合に対してSteel-Dwass法を用いて検定した．
The same accuracy results were obtained for six conditions: two scenes (Office, Supermarket) and three algorithms (I-VT, I-DT, I-VDT).
% As in the experiments described above, according to the prior Shapiro-Wilk method, the absolute error for each parameter range was not normally distributed.
As in the experiments described above, the Steel--Dwass method was applied for testing against a set of parameter ranges for each condition. 

%----------------------------------------------------------------------
% 固視検出時のパラメータによる影響（結果）
%----------------------------------------------------------------------
\subsubsection{Results}
\begin{figure}[t]
\begin{center}
\includegraphics[width=0.9\columnwidth]{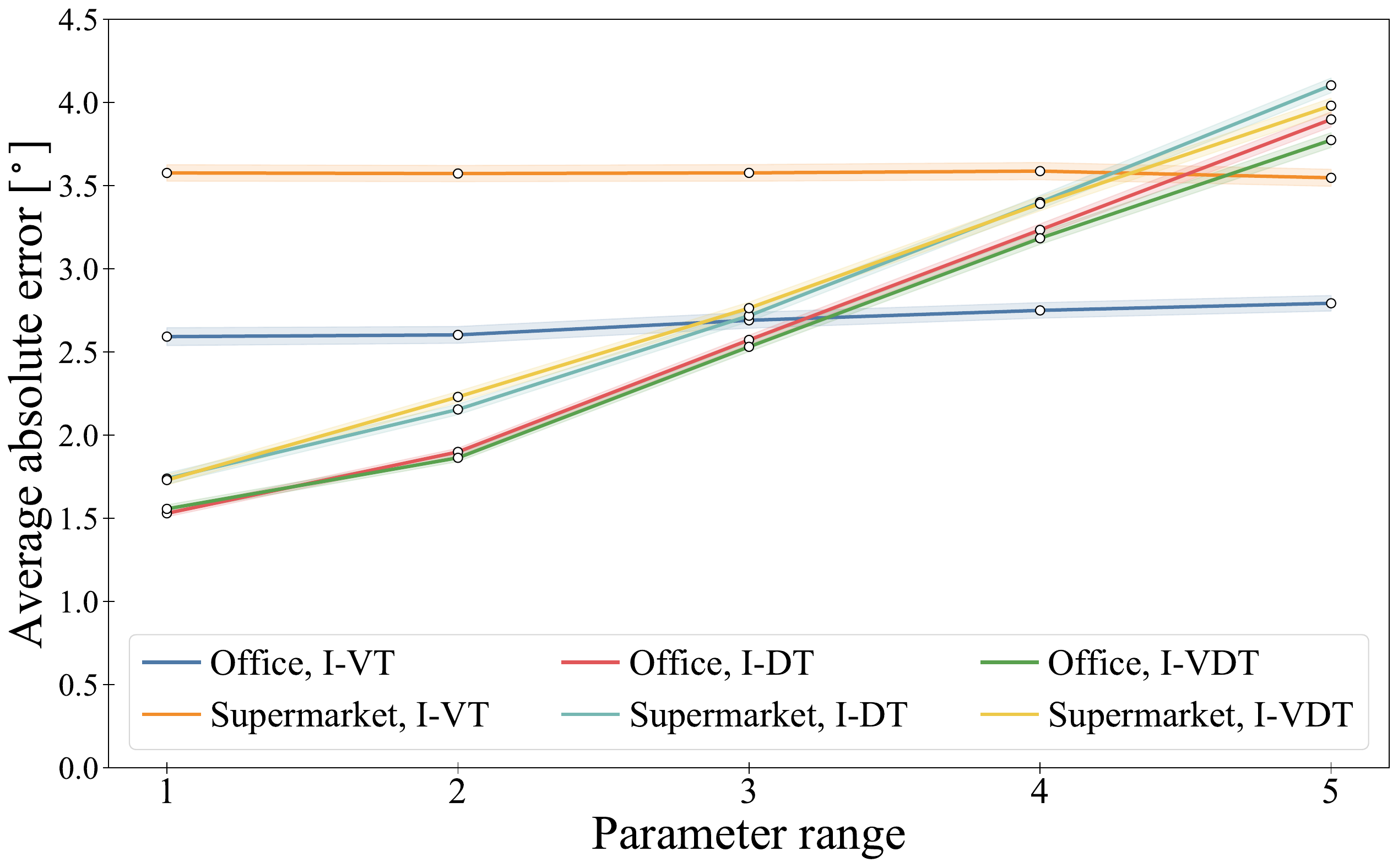}
\end{center}
\vspace{-4mm}
\caption{Dependence on parameters for fixation detection.
Range~$l$ indicates ${l-1 \leq|\alpha|\leq l}$ or ${l-1 \leq|\beta|\leq l}$ ($l=1,2,\cdots,5$ in degrees). The bands with light colors indicate standard errors.
}
\label{fig:exp-param}
\end{figure}

% 図\ref{fig:exp-param}は，提案手法の精度と固視検出に必要な較正パラメータの関係を示す．
% グラフのバンドで表されるエラーバーは，標準誤差を示す．
Fig.~\ref{fig:exp-param} shows the relationship between calibration accuracy and parameter range.
The error bands with light colors along the lines indicate standard errors.
%----------------------------------------------------------------------
% パラメータ範囲間の比較
%----------------------------------------------------------------------
% 二つのシーンにおいて，I-VDTとI-DTはともに，全ての各パラメータ範囲間が統計的に有意(p$<$0.01)であった．
% オフィスシーンにおけるI-VTは，パラメータ範囲1と3，1と4，1と5，2と4，2と5の間では有意差(p$<$0.01)が確認されたが，それ以外では，有意差は確認されなかった．
% スーパーシーンでのI-VTは，どのパラメータ範囲間も統計的に有意ではなかった．
In the cases of I-DT and I-VDT, any two of the parameter ranges were significantly different ($p < 0.01$) from each other in both scenes. 
% For the I-VT in the Office, significant differences (p$<$0.01) were confirmed between Ranges 1 and 3, 1 and 4, 1 and 5, 2 and 4, and 2 and 5, although no significant differences were confirmed otherwise.
For I-VT in Office, significant differences ($p < 0.01$) were observed between Range~1 and any of Range~3--5 and between Range~2 and any of Range~4 and 5.
For I-VT in Supermarket, no significant difference was observed between any two parameter ranges.

%----------------------------------------------------------------------
% アルゴリズム間の比較，シーン間の比較
%----------------------------------------------------------------------
% オフィスシーンでは，パラメータ範囲3でI-DT，I-VT，I-VDTの精度が同程度で，スーパーシーンでは，パラメータ範囲4でI-DT，I-VT，I-VDTの精度が同程度である．
% そのため，I-DTとI-VDTはパラメータ範囲1,2で有効であり，I-VTはパラメータ範囲5で有効である．
% どの固視検出アルゴリズムのどのパラメータ範囲においても，オフィスよりもスーパーマーケットでの精度が低かった．
% これは，I-VTで顕著に確認された．
In Office, the absolute error of the three methods was comparable in Range~3.
In Supermarket, the absolute error of I-VT, I-DT, and I-VDT were comparable in Range~4.
Therefore, I-DT and I-VDT are superior for Range~1 and 2, whereas I-VT is superior for Range~5.
In additon to the results of the accuracy evaluation shown in Fig.~\ref{fig:abs-result}, the absolute error was lower in Supermarket than in Office for all parameter ranges of any of the fixation detection algorithms.

\subsubsection{Discussion}
%----------------------------------------------------------------------
% 固視検出時のパラメータによる影響（考察）
%----------------------------------------------------------------------
% 図~\ref{tb:param-result}は，特定の固視検出アルゴリズムと較正パラメータを選択したときの提案手法の平均的な精度を示している．
% 固視検出アルゴリズムによって，パラメータ依存性が大きく異なることが分かった．
%Fig.~\ref{fig:exp-param} shows the average accuracy of the proposed method when a specific fixation detection algorithm and calibration parameters for the fixation detection are selected.
%We found that the parameter dependence varied greatly depending on the fixation detection algorithm.
Fig.~\ref{fig:exp-param} clearly shows that some fixation detection algorithms such as I-VT have almost no parameter dependence on accuracy and that the initial calibration parameters for fixation detection with I-DT and I-VDT have a strong influence on the overall processing.
The fact that accurate initial parameters are needed to estimate accurate calibration parameters may seem to be a ``chicken-or-egg" situation; nevertheless, the proposed method was able to estimate parameters with higher accuracy than the optical axis.
To further improve accuracy, two issues need to be addressed: providing accurate initial parameters and developing an accurate fixation detection method with lower parameter dependence.

The output of the conventional methods can be used to obtain accurate initial parameters.
For accurate fixation detection methods, fortunately, our experiments allow a quantitative evaluation of the accuracy of a fixation detection method itself. 
Studies on the objective and quantitative evaluation of fixation detection methods are limited and do not deal with 3D scenes where the head moves~\parencite{Blignaut_Springer_APAP2009,Komogortsev_BRM2013}.

In summary, based on the preceding results,
it becomes apparent that Assertion III has been elucidated.

\section{Limitations}
\label{sec:limitations}
% 本節では，複数の視点から提案法および実験により確認されたことの限界を明らかにする．

It has been demonstrated that by showing Assertions I to III, the proposed method can automatically calibrate an eye-tracker with an accuracy approximately 2\Deg in a situation where the head moves within arbitrary 3D environment without using images.
In this section, we clarify the limitations of the proposed method.
%and what has been confirmed by experiments from several perspectives.

\subsection{Statistical Comparison with Previous Studies}
% 本論文は，従来手法との統計的な精度比較はしていない．
% この弱点をカバーするために，定性的な比較と各論文で報告された精度との定量比較を示している．
% 定性的比較では，提案手法がシーンを限定しないことから\parencite{Liu_IEEE_Access2020,Shi_Elsevier_CAG2020,Murauer_iWOAR2018,Tamura_ETRA2020}のみが比較対象となることをSec2で示した．
% 定量比較では，それら4手法で報告された精度が，提案手法よりも明らかに悪いことについて述べている．
% この原因は，入力データの違いと手法の違いの二つが考えられるが，本実験条件に特段提案手法に有利な条件は見当たらない．
% この事から，提案手法は，一定程度の可能性を示していると言える．
The qualitative comparisons in Table~\ref{table:auto-calib} show that the conventional methods available for 3D environments~\parencite{Liu_IEEE_Access2020,Shi_Elsevier_CAG2020,Murauer_iWOAR2018,Tamura_ETRA2020} are directly comparable to the proposed method.
However, we did not perform a quantitative comparison with these conventional methods.
The same is true of most of studies on these conventional methods.
The reason is probably the same as ours: the source code is not open, and implementation details are not sufficiently clear.
Therefore, the standardization of evaluation methods and quantitative comparison under the same conditions will be our focus in future work.

\subsection{Moving \& Deforming Objects}
% 今回は実装が複雑化することを避けるために，シーンに動物体を含めなかった．
% 理論上は，動物体を注視するsmooth pursuitが生じれば，オブジェクト座標上では頭部移動が生じていることと同等である．
% これは提案法にとって有利に働くはずである．
% これを実現するためには，固視クラスタごとにオブジェクトIDを保持し，各オブジェクト座標系でのシーンカメラの位置姿勢を記述し，投影関数，逆投影関数を表す．
% 変形する物体に対しても同様で，固視クラスタが存在するポリゴンに固定された座標系でシーンカメラを記述すれば良い．
To avoid complications in implementation, we did not include moving objects in the VR scenes.
A smooth pursuit to gaze at a moving object is equivalent to a moving head in the object coordinate system.
Theoretically, the presence of moving objects should work to the advantage of our method.

Moving or deforming objects complicate the calculation of the collision position in the gaze direction.
In particular, the collision point depends on the time as well as the position and orientation of the scene camera.
Therefore, during optimization, the collision points must be calculated with various parameters at various times.
The implementation can be simplified using only the line of sight to static objects.
% In order to implement this, we need to keep an object ID for each fixation cluster, and describe the scene camera position (or the projection and back projection functions) in each object coordinate system.
% The same is true for deforming objects. 
% We can describe the scene camera in a coordinate system that is fixed to the polygon where the gaze cluster exists.

\subsection{Physiological \& Psychological Factors}
The proposed method does not take optokinetic response (OKR) and vestibulo-ocular reflex (VOR) into account, although free head movement and rotation are assumed.
The accuracy of the proposed method could be improved using a mathematical model of OKR or VOR gain, which causes gaze shift.
In addition, the proposed method does not take psychological effects into account.
The introduction of oculomotor and psychological models is a future challenge.

\subsection{Cyclopean Eye}
% 多くの従来のシーンモデルベース手法においては，両眼を独立に扱っているが，本論文では，両眼中央にあるCyclopian eyeのシーンカメラを配置して較正パラメータを算出した．
% この理由は，本論文で扱うシーンは，従来法で対象とされていた平面モニタとくらべて遠い物体が多く，両眼独立に扱っても効果が殆どないことが予想されたためである．
% 左右の各眼に１つずつシーンカメラを配置すれば両眼独立に扱うこと自体は容易である．
In many conventional scene model-based methods, both eyes are treated independently.
In contrast, in this study, we applied the scene camera at the cyclopean eye, as mentioned in \hyperref[sec:exp1-discussion]{\it Discussion} of \hyperref[sec:Accuracy Evaluation]{\it Accuracy Evaluation}, because independently treating both eyes was expected to have little effect.
The scene depth in our experimental environments was dominated by the region over 5~m, and we expected that the change in convergence angle would not be stable compared with the case of a planar monitor placed at about 1~m.
% It was expected that treating them independently would have little effect or cause instability.
Because many gaze analyses use some method of averaging the gaze directions of both eyes~\parencite{Nakamura_VR2003,Wang_ETRA2018}, we conclude that the proposed method is at least effective for such applications.

\subsection{Optimization Algorithm \& Computation Time}
% 提案手法の最適化には，多数の局所解が存在する．
% differential evolutionは全探索な手法に比べ，推定精度は落ちるが，計算時間は早い．
% 今回は，実装の都合上，並列化は行わなかった．
% 最適化では，固視ごとに複数の視線を3Dオブジェクトにレイキャストして再投影誤差を計算する．
% そのため，提案手法の最適化は並列計算が可能であり，最適化時間の向上する余地がある．
Many local solutions to the optimization of the proposed method are available.
Differential evolution is less accurate than the brute force method.
The calculation of collision points and the projection of PoRs require most of the computation time during optimization, which can be parallelized.
Therefore, there is room for improving the computation time.
% 一方で，初期値や勾配を用いたアルゴリズムは，解の不安定性や局所解という問題は残るものの，計算時間の大幅な改善が見込める．
% そのため，より適した最適化アルゴリズムの選定により，平均精度と計算時間が向上する余地があると言える．
% On the other hand, optimization methods using initial values and gradients are expected to significantly improve the computation time, although the problems of solution instability and local solutions remain.
% Therefore, there is room to improve the average accuracy and computation time by selecting a more suitable optimization algorithm.
% また，視軸で固視検出した場合の最適値よりも，光軸で固視検出した場合の最適値の方が大きい例が複数確認された．
% そのため，最適化に固視検出を含めなかった．
% In addition, there were several cases where the optimal value when the fixation was detected in the optical axis was larger than the optimal value when the fixation was detected in the visual axis.
% For this reason, we did not include fixation detection in the optimization.

\subsection{AR Scenario}
% 提案手法は，シーンモデルや頭部姿勢を前提条件としており，VRシーンに適した手法であるが，理論上ARシーンにも拡張可能である．
% シーンモデルや頭部姿勢の精度が較正精度にどの程度影響するのかは，非常に重要で関心がある問題である．
The proposed method is suitable for VR environments because of its assumptions about scene models and head postures; theoretically, it can be also extended to AR environments.
The difference between AR and VR is the accuracy of the scene model.
The effects of the scene model and head pose accuracy on calibration accuracy is a topic of great importance and interest.

\section{Conclusion}
\label{sec:conclusion}
% 本論文では，固視検出を前提とした視線計測器の自動較正手法を提案した．
% 提案手法は，シーンモデルを用いる手法の一種であり，平面ディスプレイを仮定する多くの従来法の拡張版である．
% 提案手法は，大きな頭部移動や，多数のオクルージョンを含む非平面のシーンに対応しており，バーチャルリアリティ用のHMDに搭載された視線検出器の較正に向いていいる．
In this study, we propose a self-calibration method for an eye-tracking VR HMD based on fixation detection.
Although the proposed method can be categorized as a scene model-based method, it is applicable to arbitrary scenes with large head movements and a large number of occlusions, unlike conventional methods.
% VRシーンを用いた実験では，提案手法は固視検出アルゴリズムと，検出に必要なパラメータに依存することを明らかにした．
% 一般的な3D eye modelを想定した提案手法の較正パラメータの推定精度が約2\Degで，先行研究と同程度以上の精度であることを確認した．
% さらに，固視検出アルゴリズムでは約1\Degまで向上する可能性があることを確認した．
The proposed method can automatically calibrate the eye tracker with an accuracy of about 2\Deg in about 30~s if the user walks about 20~m.
The 3D I-VDT algorithm is the best for fixation detection in terms of accuracy and computational cost compared with the I-VT and 3D I-DT algorithm.
% There is little room for improvement in the evaluation function for the optimization.
The evaluation function for optimization has little room for improvement.
Further improvement in accuracy can be achieved by developing a fixation detection algorithm that is less dependent on the initial calibration accuracy or by developing a method for the simultaneous optimization of both the calibration parameters and the fixation detection results.
Quantitative comparison with conventional methods under uniform conditions is the main future research avenue.

\section{Acknowledgments}
This work was supported in part by the JSPS Grant-in-Aid for Scientific Research No. 21K03964 and 22H03632.

\printbibliography

\appendix
\section{Calibration Process}
\subsection{Implementation Details}
\label{sec:calib-process}
\begin{figure}[t]
\begin{center}
\includegraphics[width=0.6\columnwidth]{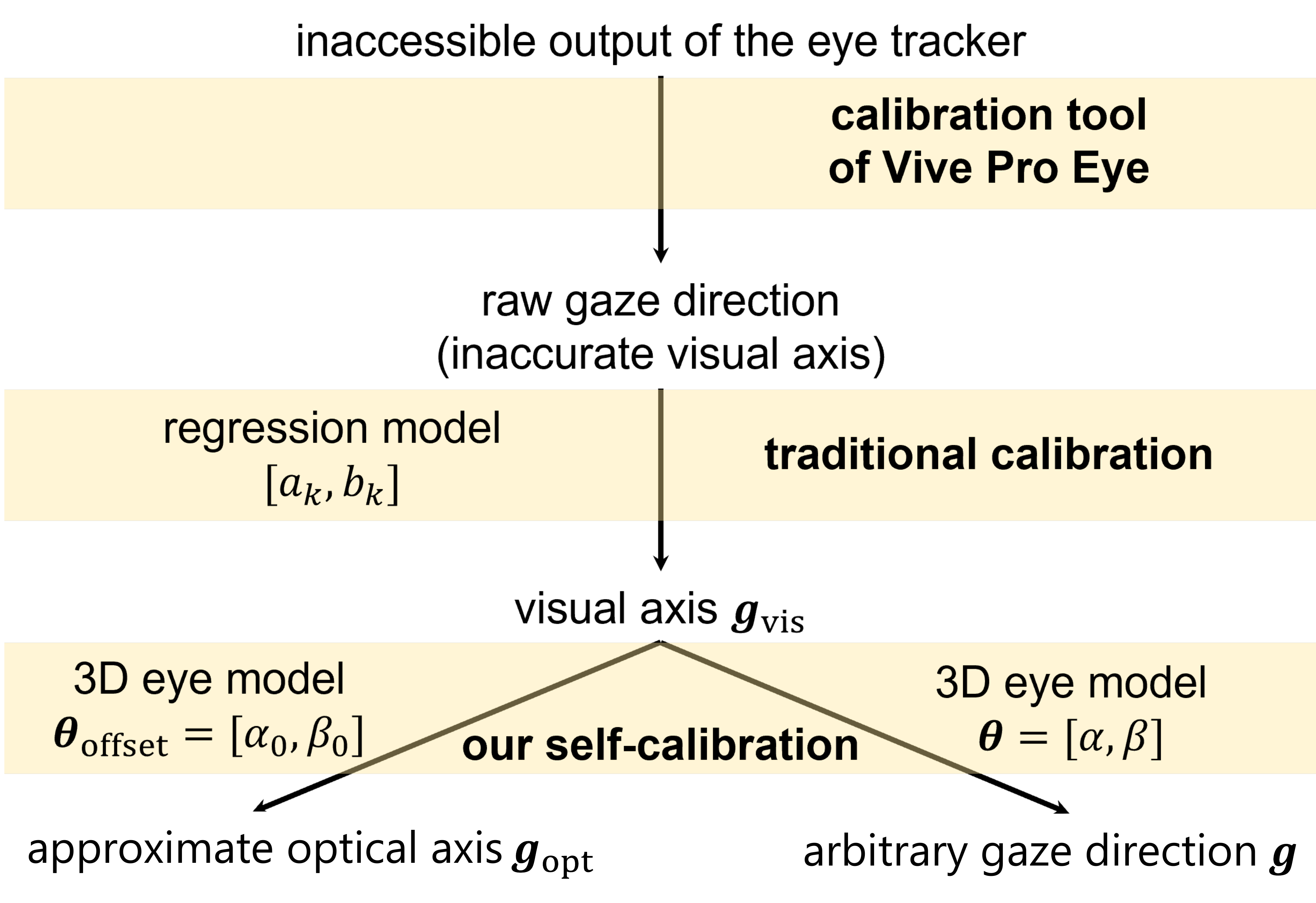}
\end{center}
% \vspace{-4mm}
\caption{Schematic of the calibration process in our experiment.
% このプロセスには，3種類の「較正」が使用される．
% この文書では，traditional calibrationの方法について述べる．
% 本論文では，一番下のプロセスであるour self-calibrationを提案した．
Three types of calibration method are used in this process.
This document describes the traditional calibration method for estimating the visual axis.
}
\label{fig:calib_process}
\end{figure}

% 我々の実験では，図\ref{fig:calib_process}のschematic illustrationに示すように，3種類の異なる「較正」という単語が登場し，それら読者を混乱させるかもしれない．
% 本節では，まずこれらの違いと必要性についてour paperとの重複を恐れに説明したい．
In our manuscript, three different calibration methods are referred to, which may confuse readers.
We have summarized the relationship between these calibration methods as a schematic in Fig~\ref{fig:calib_process}.
In this section, we explain the differences between and the necessity of these calibration methods.

% 1つ目のキャリブレーションはHTC Vive Pro Eyeの「built-in calibration tool」である．
% この較正は，その精度がどうであれ，実行しなければ仕様上，視線方向など様々なeye trackerの出力が得られない．
% そのため我々の実験では，実験参加者とは異なる著者の1名が較正を実行した．
% もう一つの重要な点は，この較正後の視軸に相当するデータしか得られず，光軸は得られないことである．
% したがって，光軸が直接得られるeye trackerを使う場合は，1つ目の較正と次段落で述べる2つ目の較正は必要ない．
% しかも，得られる視軸データは残念ながら我々の予備実験では，精度が低く1度を超えることが多かった．
% そのため，次に述べる2番目の較正が必要となる．
The first involves the use of the built-in calibration tool of the HTC Vive Pro Eye.
The important point is that, whatever its accuracy, any data cannot be obtained without performing this calibration process.
To overcome this limitation, one of the authors performed this calibration before data acquisition from the participants of our experiments.
Another point is that this eye tracker provides only the visual axis calibrated by this process and not the optical axis. 
Unfortunately, the accuracy of this visual axis was low, often exceeding 1\Deg.
Therefore, the second calibration method described below is required.

% 2番目の較正は，上記視軸の精度を補う目的で行う較正である．
% 本文中では，「traditional calibration」もしくは「traditional marker-based calibration」と呼んでいる処理がこれに相当する．
% この処理が，本documentで詳説する較正である．
% この較正では，精度を重視して，良く知られたregression modelと十分な数のマーカと計測時間を設けて，精度の高い視軸を推定する．
% この処理について次節および\ref{sec:ground-truth}節で詳述する．
The second calibration method is applied to compensate for the lack of accuracy of the calibration tool.
This method is referred to as traditional calibration or traditional marker-based calibration in our manuscript and is mainly described in this document.
In this method, a well-known regression model and a sufficient number of markers are used to accurately estimate the visual axis.
We detail this method in the next section and in \hyperref[sec:ground-truth]{\it Calibration Error}.

% 3番目の較正が，我々の論文で提案するself-calibrationである．
% これは3D eye modelを採用しており，2番目の較正により得られた信頼性の高いvisual axisを基に，最適化のための任意のパラメータに対応する視線方向$\bm g$や，典型的なオフセット角から算出したapproximated optical axis $\bm g_{\rm{opt}}$を算出している．
% これにてついては本文4.1節の「Ground Truth and Accuracy Evaluation」で詳しく述べているので本ドキュメントでは省略する．
The third calibration method is the self-calibration method proposed in our study.
It employs a 3D eye model and computes the approximate optical axis $\bm g_{\rm{opt}}$ at the typical offset angles and the gaze direction $\bm g$ corresponding to the arbitrary parameters for optimization based on the reliable visual axis estimated in the second calibration.
This is discussed in detail in \hyperref[sec:Control Condition and Accuracy Evaluation]{\it Control Condition and Accuracy Evaluation} in our manuscript, and is therefore omitted from this document.

% 本実験では，HMDにHTC Vive Pro Eyeを使用し，視線データはHMD内蔵の視線計測器を使用して取得した．
% HMD内蔵の視線計測器を使用する場合，システム標準のアプリケーションによる能動的な較正を行わなければならない．
% In our experiment, HTC Vive Pro Eye was used as the head-mounted display (HMD), and gaze data was acquired using the HMD's built-in eye tracker.
% User calibration with the system-standard application must be performed before using the eye tracker.
% そのため，HMD内蔵の視線計測器では，ユーザごとの光軸を得ることができず，HMDのシステム較正後の視線しか取得できない．
% また，システム較正の精度は，1\Degを超える場合もあり，十分ではない．
% Therfore, the eye tracker cannot obtain the optical axis of each user and can only obtain the gaze data after the HMD's system calibration.
% In addition, the accuracy of the system calibration is not sufficient, sometimes exceeding 1\Deg.
% 本実験では，古典的な較正方法によって推定された視軸をベースラインとして使用した．
% この古典的な較正は，デバイスに依存せずにベースラインを生成するためのものであり，式(\ref{eq:calib_func})による較正とは異なる．
% We used the visual axis obtained by the traditional calibration method as a ground truth.
% Note that this calibration is different from our self-calibration based on Equation~(3) in our paper.

\subsection{Setup of Calibration Markers}
\label{sec:calib-markers}
%----------------------------------------------------------------------
% シーンカメラに対するマーカの位置の関係
% シーンカメラと眼の位置の関係
% 二種類のマーカ
%----------------------------------------------------------------------
% 本節では，traditional calibrationのための，実験セットアップについて詳述する．
% 具体的には，推定用と評価用の二種類のマーカの配置を説明し，使用者の視野を表すシーンカメラ，マーカ位置，眼の位置関係について述べる．
This section details the experimental setup for the traditional calibration.
In particular, we explain the relationship between the scene camera that represents the user's field of view (FOV), marker positions, and eye positions.
We also describe the arrangement of two types of markers: one for estimation and the other for evaluation.

\begin{figure}[t]
\begin{center}
\includegraphics[width=0.7\columnwidth]{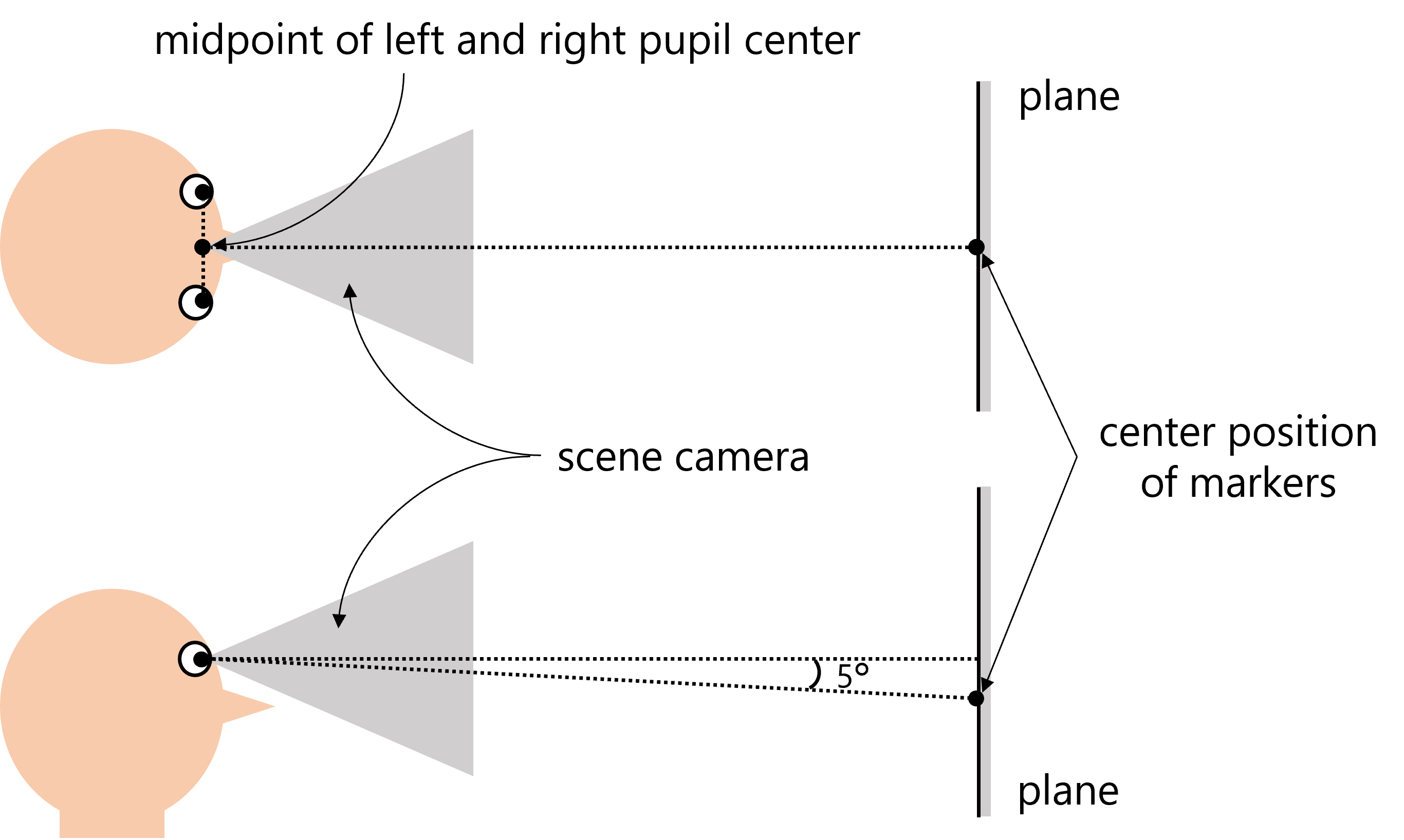}
\end{center}
% \vspace{-4mm}
\caption{Relationship between the scene camera, marker positions, and eye positions.
% シーンカメラは両眼の中央に配置した．
% マーカの中心位置は，シーンカメラの光軸から垂直方向に約5\Deg下向きに配置した．
The scene camera was placed at the center of both eyes.
The center of markers was positioned vertically downward from the z-axis of the scene camera by approximately 5\Deg.
}
\label{fig:calib_setup}
\end{figure}

% 視軸は，25個（=水平5$\times$垂直5）のクロスマーカを注視することによって得た視線データから推定した．
% マーカはシーンカメラ座標系に固定された前方1mの平面上にランダムな順に一つずつ表示された．
We used 5~$\times$~5~black cross markers.
The markers were displayed one by one in pseudo random order on a gray-colored plane fixed in the scene camera coordinate system 1~m in front of the user, with a FOV angle of 20\Deg both horizontally and vertically.
% Fig.~\ref{fig:calib_setup}に，シーンカメラと眼，マーカの位置関係を示す．
% 提案手法では，HMD上では両眼に異なる画像が提示されていても，視線の計算では，cyclopean eyeを表す1台の仮想のシーンカメラ（ピンホールカメラ）を用いる．
% シーンカメラの原点は，視線計測で取得された両瞳孔中心の中央に配置した．
% シーンカメラの姿勢は，HMD付属のLighthouseトラッキングシステムにより取得した．
Fig.~\ref{fig:calib_setup} shows the positional relationship between the scene camera, markers, and both eyes.
Our method requires two kinds of pinhole scene cameras: a single camera for gaze analysis and stereoscopic cameras for rendering.
We created the scene camera for analysis by translating one of the cameras for rendering.
The origin of the scene camera was placed at the midpoint of both pupil centers obtained by the eye tracker.
The pose was set to the same as those of the cameras for rendering, 
which were at the default setting obtained by the Lighthouse tracking system attached to the head-mounted device.
% 人間の垂直方向の視野角は，上側よりも下側の方が10から20\Deg程度広いため，マーカの中心は被験者の眼の高さから5\Deg下側に表示した．
Because the vertical FOV of humans is approximately 10\Deg -- 20\Deg wider at the lower end than at the upper end~\parencite{Su_Access2020}, the center position of the plane was placed 5\Deg below the central axis of the scene camera or, more precisely, 5\Deg below the optical axis of the pinhole camera for rendering.

% In the experiment, we asked each participant to gaze at each marker for 3~s after the user pulls the trigger on the HTC Vive controller.
% To ensure the reliability of gazing at the markers, we used the middle 1~s of the 3~s.
% This resulted in 25~sets of gaze data for each participant.

\begin{figure}[t]
\begin{center}
\includegraphics[width=0.55\columnwidth]{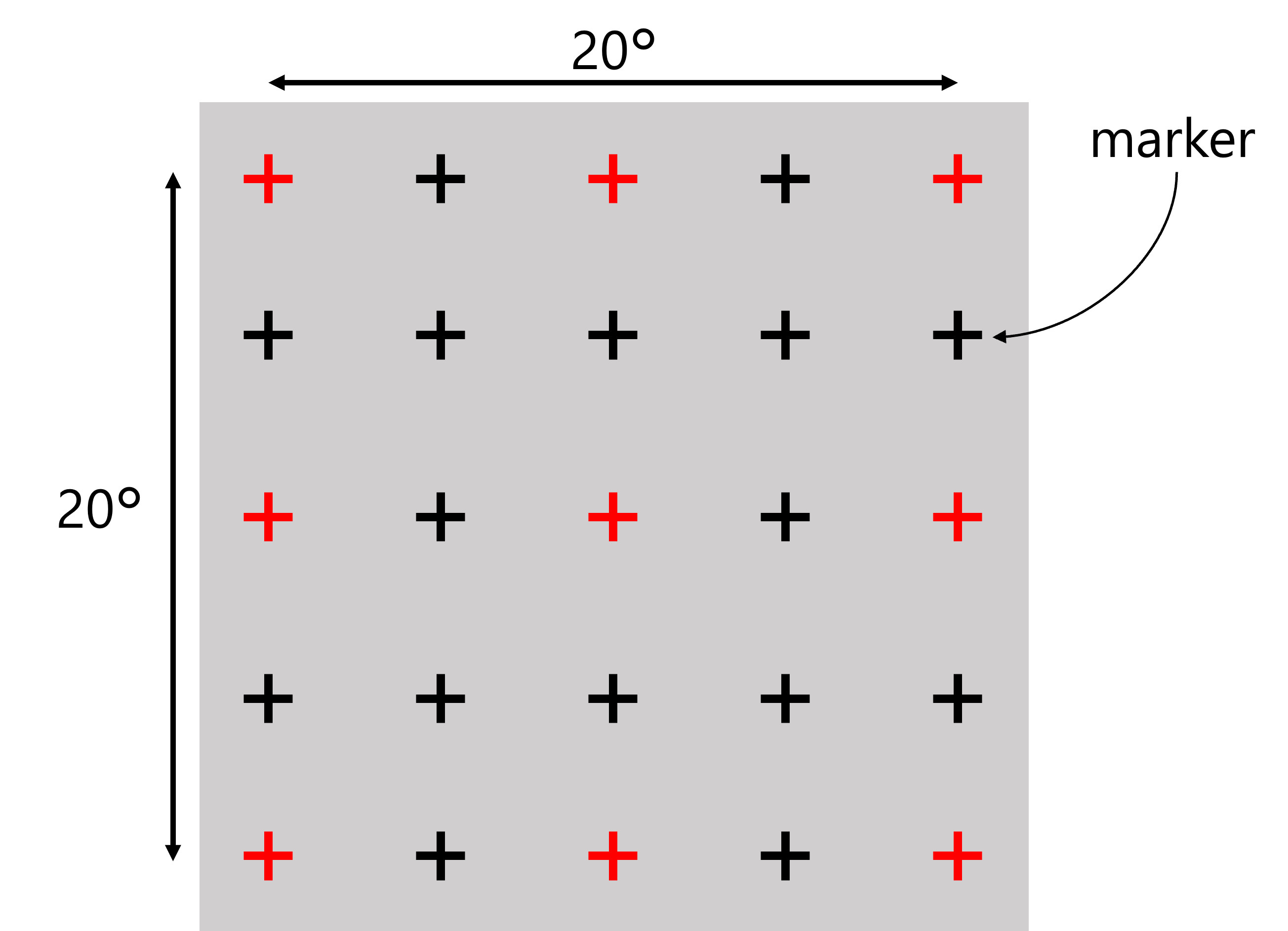}
\end{center}
% \vspace{-4mm}
\caption{Marker positions for the traditional calibration.
% 赤のマーカは，視軸の推定に使用した．
% 黒のマーカは，較正パラメータの評価に使用した．
% 各マーカは，視野角20\Degの範囲内に表示された．
Black crosses represent the markers used to estimate the visual axis.
Red ones were the markers used to evaluate calibration parameters.
Actually, the markers were presented one by one in black color.
All the markers were displayed within a viewing angle of 20\Deg.
}
\label{fig:marker}
\end{figure}

% データ計測のため，被験者に各マーカを3s間注視してもらった．
% 計測時には始め，平面上に一つのマーカが表示されており，被験者がHTC Viveコントローラのトリガーを引くと，ビープ音により3s間カウントダウンされる．
% 被験者は，ビープ音が鳴っている間，マーカを注視した．
% ビープ音が終了すると，マーカは擬似ランダムな位置にジャンプし，新たにトリガーを引くとビープ音が鳴り始める．
% 被験者はこれを25回繰り返し，結果として，25セットの視線集合を得た．
% 確実にマーカを注視しているデータで推定と評価をするために，我々は3sのうち1sを使った．
To measure the gaze data, we asked each participant to gaze at each marker for 3~s.
Throughout the measurement, only one black marker appeared on the plane.
When the participant pulls the trigger on the HTC Vive controller, a beep sound counts down for 3~s.
The participants were asked to gaze at the marker during the beep.
At the end of the beep, the marker jumped to a pseudo random position, and a new trigger pull started the beep.
Each participant repeated this 25 times.
As a result, we acquired 25 sets of gaze data for each participant.
To estimate and evaluate the reliability of the marker gazing data, we used the middle 1~s out of the 3~s.
% 図\ref{fig:marker}に，表示したマーカ位置を示す．
% 視線データ25セットのうち16個を推定用に，9セットを評価用に用いた．
Fig.~\ref{fig:marker} shows the positions of the displayed markers.
Sixteen of the twenty-five~sets (black crosses in Fig.~\ref{fig:marker}) of gaze data were used for the estimation of the visual axis, and the other nine sets (red crosses in Fig.~\ref{fig:marker}) were used for the evaluation of the calibration parameters.

\subsection{Calibration Error}
\label{sec:ground-truth}
% 本節の前半では，16個のマーカを用いて，regression modelにより視軸を推定する方法を説明し，後半では，9個のマーカを用いて，パラメータの精度評価法について説明する．
This section explains the regression model for estimating the visual axis using 16 markers and the method for evaluating the accuracy of the calibration parameters using nine markers.
% 視軸は，次の1次多項式\cite{White_IEEE_SMC1993}を使用して，平面上における各PoRとマーカの距離の二乗和が最小となるパラメータを求めた．
The visual axis was estimated to minimize the sum of squared distances between each marker and its points of regard (PoRs) on the plane using the following first-order polynomials~\parencite{White_IEEE_SMC1993}:
\begin{eqnarray}
u_{\rm{vis}} &=& a_1 + a_2 \hspace{0.5mm} u_{\rm{raw}} + a_3 \hspace{0.5mm} v_{\rm{raw}} \label{eq:model1},\\
v_{\rm{vis}} &=& b_1 + b_2 \hspace{0.5mm} u_{\rm{raw}} + b_3 \hspace{0.5mm} v_{\rm{raw}} \label{eq:model2},
\end{eqnarray}
% ただし，$(u',v',1)$は，SRanipal SDKを使ったキャリブレーション後の視線である．
% $(u,v,1)$は，節\ref{sec:preliminary}で定義した視軸$\bm g$の成分である．
% 従って，ベースラインパラメータは，$\{\alpha,\beta\}=\{0,0\}$である．
where $[u_{\rm{raw}},v_{\rm{raw}},1]^T$ is the raw gaze direction after calibration using the tool of the head-mounted display.
$[u_{\rm{vis}},v_{\rm{vis}},1]^T$ is the visual axis as the control condition.
% The optical axis and any calibrated gaze direction are generated from the visual axis by Equation~(19) and (20) in our paper, respectively.
% In this case, the ground truth parameters can be represented as $(\alpha,\beta)=(0,0)$.
% In our implementation, the initial parameters for the fixation detection and any parameters in the optimization process were generated from the visual axis because our eye tracker does not provide the optical axis directly.
% This is a simulation-like method, which differs slightly from the actual usage, where parameters in the optimization are generated by rotating the optical axis obtained by an eye tracker.
% However, for the purpose of confirming the accuracy, it is equivalent to the actual self-calibration.

\begin{figure}[t]
\begin{center}
\includegraphics[width=0.6\columnwidth]{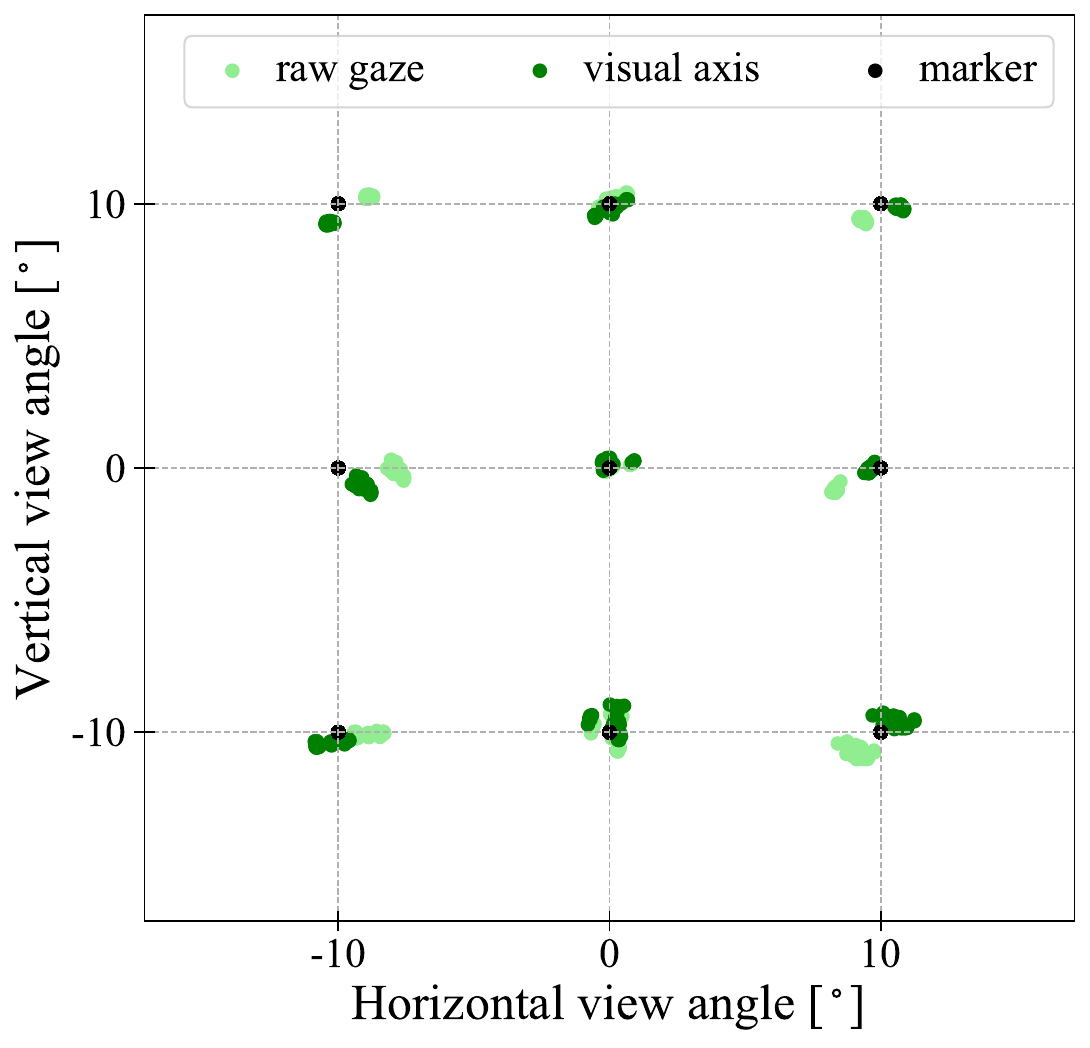}
\end{center}
% \vspace{-4mm}
\caption{PoRs of raw gaze direction and visual axis for markers.
% 薄緑と緑の点は，マーカが表示されている平面上のraw gaze directionと視軸の注視点を示す．
% 黒い点は，マーカ位置を表す．
The light green and green points indicate the PoRs of the raw gaze direction and visual axis on the plane where markers are displayed, respectively.
The black points represent marker positions.
}
\label{fig:eval_data}
\end{figure}

% Fig~\ref{fig:eval_data}に，マーカ注視時の最初の較正後の視線と，二回目の較正後の視線を示す．
% 人の視線は揺らぐため，Fig~\ref{fig:eval_data}中において，真値であるvisual axisにも誤差が存在する．
% これが，機械学習において訓練データと評価データを分けるように，我々は精度を評価するため，regression modelのパラメータ推定に必要なマーカと評価用のマーカを分けた理由である．
Fig.~\ref{fig:eval_data} shows the gaze directions after the first calibration (raw gaze direction in Fig.~\ref{fig:calib_process}) and the gaze directions after the second calibration (visual axis in Fig~\ref{fig:calib_process}) when gazing at the nine markers for evaluation.
Because the human gaze fluctuates, even with the visual axis, which is the control condition, has errors.
This is why we separated the markers necessary for estimating the regression model parameters from those required for evaluating the calibration parameters, similar to machine learning where training data and evaluation data are separated.
% regression modelのパラメータや，提案手法で使用される3D eye modelのパラメータを用いて9セットの視線を較正し，マーカと較正後の視線との誤差を度単位で評価した．
% Fig~\ref{fig:eval_data}の例では，raw gaze directionの誤差は1.02\Degで，visual axisの誤差は0.65\Degであった．
For evaluation, nine sets of gaze directions were calibrated using the parameters of the regression model or the 3D eye model used in the proposed method, and the average absolute error between the markers and the visual axis or the calibrated gaze directions was computed in degrees.
In Fig.~\ref{fig:eval_data}, for example, the error with the raw gaze direction was 1.02\Deg and that with the visual axis was 0.65\Deg.

\end{document}